\pdfoutput = 1
\documentclass[twocolumn]{article}
\usepackage[numbers]{natbib}
\usepackage{balance}
\usepackage{style}

\usepackage{gensymb}
\usepackage{mhchem}

\usepackage{xurl}

\usepackage{times}

\usepackage[english]{babel}
\usepackage{blindtext}
\usepackage{graphicx}
\usepackage{amsmath, amsthm, amssymb, bbm, bm}
\usepackage{enumerate}
\usepackage{float}      
\usepackage{subcaption}  
\usepackage{wrapfig}
\usepackage[margin=0cm]{caption}
\usepackage{tabularx}

\usepackage{multirow}
\usepackage{hhline}
\usepackage{makecell}
\usepackage{placeins}  

\usepackage[x11names, usenames, dvipsnames, svgnames, table]{xcolor}
\definecolor{firebrick}{rgb}{.698,.133,.133}
\definecolor{mybluelight}{rgb}{0.9, 0.9, 1.}
\definecolor{myorangelight}{rgb}{1., 0.9, 0.9}

\usepackage[utf8]{inputenc} 
\usepackage[T1]{fontenc}    
\usepackage{url}            
\usepackage{booktabs, colortbl}       
\usepackage{amsfonts}       
\usepackage{nicefrac}       
\usepackage{microtype}      

\usepackage{csquotes}
\usepackage{latexsym}

\usepackage{pifont}
\usepackage[boxruled, vlined, linesnumbered]{algorithm2e}
\SetAlFnt{\small}
\SetAlCapFnt{\small}
\SetAlCapNameFnt{\small}
\usepackage{algorithmic}
\algsetup{linenosize=\tiny}

\let\oldnl\nl
\newcommand{\nonl}{\renewcommand{\nl}{\let\nl\oldnl}}

\usepackage{paralist}

\usepackage{xspace}
\usepackage{soul}
\usepackage{dsfont}
\usepackage{stmaryrd}
\usepackage[textwidth=15mm]{todonotes}
\usepackage{dirtytalk}
\usepackage{pbox}
\usepackage{cprotect}
\usepackage{bm}

\usepackage{verbatim}
\usepackage{textcomp}
\usepackage[normalem]{ulem}

\usepackage{mathtools}
\usepackage{etextools}
\usepackage[inline]{enumitem}

\usepackage[colorlinks=true,allcolors=firebrick,bookmarks=false]{hyperref}

\usepackage[toc,page,header]{appendix}
\usepackage{minitoc}

\renewcommand \thepart{}

\DeclareMathOperator{\tr}{tr}

\definecolor{darkergreen}{RGB}{21, 152, 56}
\definecolor{red2}{RGB}{252, 54, 65}
\definecolor{Gray}{gray}{0.85}
\newcolumntype{g}{>{\columncolor{Gray}}c}

\let\OLDthebibliography\thebibliography
\renewcommand\thebibliography[1]{
  \OLDthebibliography{#1}
  \setlength{\parskip}{0pt}
  \setlength{\itemsep}{0pt plus 0.3ex}
}

\newcommand{\reals}{\mathbb{R}}

\newcommand\pxtp{\texttt{TP*}\xspace}
\newcommand\pxtn{\texttt{TN*}\xspace}
\newcommand\pxfp{\texttt{FP*}\xspace}
\newcommand\pxfn{\texttt{FN*}\xspace}

\newcommand\glas{\texttt{GlaS}\xspace}
\newcommand\camsixteen{\texttt{CAMELYON16}\xspace}

\newcommand\beloc{\texttt{B-LOC}\xspace}
\newcommand\becl{\texttt{B-CL}\xspace}

\newcommand\pxap{\texttt{PxAP}\xspace}

\newcommand\cl{\texttt{CL}\xspace}

\newcommand\pixelcam{\texttt{PixelCAM}\xspace}

\theoremstyle{definition}

\DeclarePairedDelimiterX{\divx}[2]{(}{)}{%
  #1\;\delimsize\|\;#2%
}

\makeatletter
\@namedef{ver@everyshi.sty}{}
\newcommand{\removelatexerror}{\let\@latex@error\@gobble}

\newlength{\NFwidth}
\NewDocumentCommand{\NFline}{O{l}m}{\footnotesize\makebox[\NFwidth][#1]{#2}}

\NewDocumentCommand{\NFentry}{sm}{%
  \makebox[.5\NFwidth][l]{\normalsize
    \IfBooleanT{#1}{\makebox[0pt][r]{\textbullet\ }}%
    #2}\ignorespaces}

\title{PixelCAM: Pixel Class Activation Mapping for Histology
Image Classification and ROI Localization}

\renewcommand\footnotemark{}

\author{Alexis~Guichemerre${^{1}}$, \quad 
\textbf{Soufiane~Belharbi}${^1}$, \quad
\textbf{Mohammadhadi Shateri}${^1}$, \quad
\textbf{Luke McCaffrey}${^2}$, \quad
\textbf{Eric~Granger}${^1}$\vspace{0.05in}\\
$^1$LIVIA, Dept. of Systems Engineering, ETS Montreal, Canada \vspace{0.02in}\\
$^2$Goodman Cancer Research Centre, Dept. of Oncology, McGill University, Montreal, Canada \vspace{0.02in}\\
{\tt\scriptsize\{soufiane.belharbi, mohammadhadi.shateri, eric.granger\}@etsmtl.ca
}
\\
{\tt\scriptsize \{alexis.guichemerre.1\}@ens.etsmtl.ca
}
\\
}

\newcommand{\ignore}[1]{}

\begin{document}
\doparttoc 
\faketableofcontents 

\thepart{} 
\parttoc 

\maketitle\thispagestyle{fancy}
\maketitle
\rhead{\color{gray} \small Guichemerre et al. \;  [MIDL 2025]}

\begin{abstract}
Weakly supervised object localization (WSOL) methods allow training models to classify images and localize ROIs. WSOL only requires low-cost image-class annotations yet provides a visually interpretable classifier, which is important in histology image analysis. Standard WSOL methods rely on class activation mapping (CAM) methods to produce spatial localization maps according to a single- or two-step strategy. While both strategies have made significant progress, they still face several limitations with histology images. Single-step methods can easily result in under- or over-activation due to the limited visual ROI saliency in histology images and scarce  localization cues. They also face the well-known issue of asynchronous convergence between classification and localization tasks. The two-step approach is sub-optimal because it is constrained to a frozen classifier, limiting the capacity for localization. Moreover, these methods also struggle when applied to out-of-distribution (OOD) datasets. 
In this paper, a multi-task approach for WSOL is introduced for simultaneous training of both tasks to address the asynchronous convergence problem.  In particular, localization is performed in the pixel-feature space of an image encoder that is shared with classification. This allows learning discriminant features and accurate delineation of foreground/background regions to support ROI localization and image classification. 
We propose \pixelcam, a cost-effective foreground/background pixel-wise classifier in the pixel-feature space that allows for spatial object localization. Using partial-cross entropy, \pixelcam is trained using pixel pseudo-labels collected from a pretrained WSOL model. Both image and pixel-wise classifiers are trained simultaneously using standard gradient descent. In addition, our pixel classifier can easily be integrated into CNN- and transformer-based architectures without any modifications.  
Our extensive experiments\footnote{\href{https://github.com/AlexisGuichemerreCode/PixelCAM}{https://github.com/AlexisGuichemerreCode/PixelCAM}} on \glas and \camsixteen cancer datasets show that \pixelcam can improve classification and localization performance when integrated with different WSOL methods.  Most importantly, it provides robustness on both tasks for OOD data linked to different cancer types, with large domain shifts between training and testing image data. 
\end{abstract}



\section{Introduction}
\label{sec:introduction}
\dosecttoc
\secttoc

Histology image analysis remains the gold standard for diagnosing cancers of the brain~\cite{khalsa2020automated}, breast~\cite{Veta2014}, and colon~\cite{xu2020colorectal}. However, training deep learning (DL) models for accurate localization of cancerous regions of interest (ROIs) requires pixel-wise annotation by pathologists which is a complex and time-consuming task, especially over Whole Slide Images (WSI). WSOL has emerged as a low-cost training approach~\cite{zhou2017brief}. Using only image class supervision, a DL model can be trained to classify an image, but also to provide spatial localization of ROIs associated with that class~\cite{negev,choe2020evaluating,rpwsol,rony23}. This significantly reduces the large cost of annotation and the need for dense pixel supervision. In addition, it builds interpretable classifiers~\cite{belharbi2022minmaxuncer,neto24} which is critical in medical domains such as in histology image analysis.

Recent progress in WSOL is dominated by CAM methods~\cite{rony23}. 
Several methods extract spatial activation maps in a single step~\cite{deepmil,oquab2015object,SAT}, and require a module on top of a feature extractor to construct these maps. Then, a spatial pooling module extracts per-class scores used for supervised training using image-class labels. However, without any explicit guidance at the pixel level, these CAM methods can lead to poor maps due to the challenging nature of histology images where objects are often not salient~\cite{rony23}. This can lead to under- or over-activation which creates a high false negative/positive rate at the pixel level. Moreover, training a single model may face the issue of asynchronous convergence of classification and localization tasks~\cite{choe2020evaluating,rpwsol,rony23}.

To bypass this convergence issue and the lack of pixel guidance, a recent direction has emerged in WSOL where it aims at providing explicit localization cues as pseudo-labels with dual models~\cite{negev,fcam,Murtaza2023dips,wei2021shallowspol,zhang2020rethinking,murtaza2025ted,murtaza2022dipssypo,zhao2023generative,murtaza2022dips}. In particular, a per-task model is considered where the localizer is trained using pseudo-labels~\cite{zhao2023generative}. This leads to more parameters and training cycles. In addition, both tasks are disconnected leading to unrelated decisions in both models.
A different approach~\cite{negev} uses a single model where a decoder is combined with a frozen classifier.
While this yields good results, this architecture is limited as it is tied to frozen features at many layers using skip connections. This prevents the localizer from a better adaptation and hinders its performance.

In addition, recent work has revealed a major limitation to WSOL methods when dealing with domain shift in histology analysis~\cite{sfdawsol}. The performance of WSOL methods is shown to decline on both tasks with out-of-distribution (OOD), limiting their real-world application. Further inspection suggests that features at the pixel level can be the root cause. Since they lack direct localization supervision, a feature encoder may provide poorly separated features concerning classes, making them vulnerable to domain shift, and class confusion.

To alleviate the aforementioned issues, we propose a novel multi-task WSOL method for histology image analysis. It is based on called \pixelcam -- a cost-effective foreground / background (FG/BG) pixel-wise classifier working in the pixel-feature space, allowing for spatial object localization. 
It aims to explicitly learn discriminant pixel-feature representations and accurate delineation of FG and BG regions. This improves the ROI localization and image classification accuracy. Training is achieved through localization pseudo-labels extracted from a pretrained WSOL model. 
In addition, such pixel-wise classification provides the model with robustness to OOD data. Our single-step multi-task framework allows to simultaneously perform classification and localization tasks. 
%
Multi-task training is therefore leveraged to learn rich features for both tasks, compared to the constrained learning of features in a two-step approach. Our approach cooperates to converge to a satisfying solution for both tasks. The multi-task optimization is performed using standard gradient descent by using image-class labels and pixel-wise pseudo-labels extracted from a pretrained WSOL model.
%



Our main contributions are summarized as follows. 
\textbf{(1)} A novel multi-task WSOL method called \pixelcam is proposed for histology image analysis. Multi-task optimization of both classification and localization tasks is achieved in a single step by integrating a pixel classifier at the pixel-feature level while leveraging localization pseudo-labels. \pixelcam alleviates the asynchronous convergence issue in WSOL, improves the performance of both tasks, and importantly, provides robustness to OOD data. Our pixel classifier is versatile in that it can easily be integrated into any CNN- or transform-based classifier architecture without modification.
\textbf{(2)} We conduct extensive experiments on two public datasets for colon (\glas), and breast cancer (\camsixteen). Our method outperforms WSOL baseline methods on both tasks. Additionally, when dealing with large domain shifts across cancer types, our \pixelcam method can maintain a high level of accuracy, making it an ideal choice for OOD scenarios. We provide several ablations to further analyze our method.

\section{Proposed Method} 
\label{sec:proposed}

\begin{figure*}[!ht]
    \centering
    \includegraphics[width=0.99\linewidth]{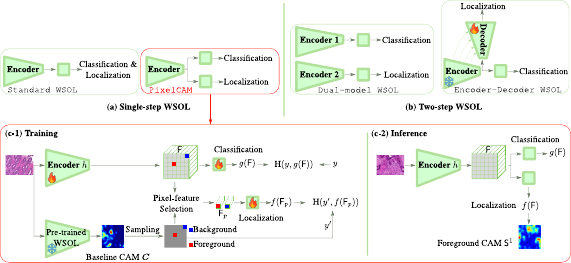}
    \caption{
    \textbf{Top row: WSOL learning strategies}. \textit{(a) Single-step WSOL methods} perform classification and localization in one step. While standard WSOL uses a sequential approach using only image-class and lacks leveraging localization cues, \pixelcam relies on a multi-task training using both image-class and localization cues. \textit{(b) Two-step WSOL methods} use a model per task or an encoder-decoder which are trained sequentially. 
    \textbf{Bottom row: \pixelcam training (c-1) and inference (c-2)}. The training relies on pseudo-labels collected from a WSOL CAM classifier pretrained on the same dataset ${\mathbb{D}}$. The classification head aims to classify the extracted features ${\mathsf{F}}$. The pixel-wise classifier predicts the randomly selected locations as foreground or background. At test time, all locations are classified using the localization head ${f}$ producing a localization map, while classifying the whole image using the classification head ${g}$.
    }
    \label{fig:proposal}
     \vspace{-1em} 
\end{figure*}

Let us denote by ${\mathbb{D} = \{(\bm{X}, y)_i\}_{i=1}^N}$ a training set with $N$ samples, where ${\bm{X}: \Omega \subset \reals^2}$ is a 2d  input image of dimension ${h^{\prime}\times w^{\prime}}$ and ${y \in \{1, \cdots, K\}}$ is its global label, with $K$ being the number of classes. Our model is composed of three parts (Fig.\ref{fig:proposal}): (a) feature extractor, (b) image classifier, and (c) pixel-wise classifier.
(a) The feature extractor backbone $h$ with parameter ${\bm{\theta}_1}$ produces a tensor spatial features ${h(\bm{X}) = \mathsf{F} \in \reals^{h^{\prime}\times w^{\prime} \times d}}$, with depth $d$. For simplicity, we consider that $\mathsf{F}$ has the same dimensions as the input image ${\bm{X}}$ through interpolation. 
(b) The global image classifier head $g$ classifies the image content, with parameters ${\bm{\theta}_2}$, where  ${g(\mathsf{F})}$ is the per-class probability.
(c) The pixel-wise\footnote{For simplicity, we refer to a location in a spatial tensor as a pixel. However, in DL models, such location typically covers more than one pixel in the image space due to the receptive field. Interpolation can be used to upscale the features to the full image size.} classifier $f$ which classifies the embedding ${\mathsf{F}_{i, j, :} = \mathsf{F}_p \in \reals^d}$ of a pixel at location ${(i, j, :)}$ or simply $p$ in ${\mathsf{F}}$ with parameters ${\bm{\theta}_3}$ into either foreground (1) or background (0) classes. This is typically a linear classifier.
We refer by ${f(\mathsf{F}_p)_0}$, ${f(\mathsf{F}_p)_1}$ as the probability for the pixel at location $p$ to be background and foreground, respectively. For simplicity, we use ${f(\mathsf{F}_p)}$ for pixel-class probability.
Classifying all locations creates the two localization maps ${\mathsf{S} = f(\mathsf{F}) \in [0, 1]^{h^{\prime}\times w^{\prime} \times 2}}$, where ${\mathsf{S}_{:, :, 0}, \mathsf{S}_{:, :, 1}}$ or simply ${\mathsf{S}^0, \mathsf{S}^1}$ refer to the background and foreground maps, respectively. For simplicity, we refer by ${y^{\prime} \in \{0, 1\}}$ as the pseudo-label of a pixel location ${p}$.
%
Let us denote the standard cross-entropy by $\mbox{H}(\cdot, \cdot)$. The total parameters of our model is referred to as ${\bm{\theta}}$ which is composed of ${ \{\bm{\theta}_1, \bm{\theta}_2, \bm{\theta}_3\}}$. When ${y, y^{\prime}}$ are used with ${\mbox{H}}$, it is the one-hot encoding version that is being considered.

To train the model for full image classification, we use standard cross-entropy over $g$:

\begin{equation}
\label{eq:image_cl}
\begin{aligned}
\min_{\bm{\theta}_1, \bm{\theta}_2} \quad &  \mbox{H}(y, g(\mathsf{F})) \;.
\end{aligned}
\end{equation}

\noindent\textbf{Pixel-wise classifier}. Training the model using only Eq.\ref{eq:image_cl} builds a spatial feature ${\mathsf{F}}$ that is guided only by the image classification loss. It therefore lacks awareness of localization as it cannot directly access localization supervision cues. This may lead to poor pixel-wise features and confuse classes at the feature level. This is particularly true when dealing with less salient and strongly similar objects, as found in histology images. Subsequently, localization on top of these features is less reliable. Most importantly, introducing domain shift over the input image will lead to further feature confusion and failure by subsequent modules to localize and classify. To mitigate this issue, our pixel-classifier $f$ is used to create well-separated features between foreground/background in the pixel-feature space.

Training $f$ requires localization cues. Since the WSOL setup does not provide them, we resort to using a pretrained WSOL CAM classifier trained on the same train dataset ${\mathbb{D}}$, based on CNN or transformer, to acquire pseudo-labels. This approach is typically employed in WSOL to train a new DL model or a decoder for localization~\cite{rony23}. However, this is cumbersome and requires another training phase with many additional parameters.  
To prevent this issue, our method trains a linear model to classify pixel embeddings while training the image classifier and sharing a full backbone. The number of parameters and training cycles is thereby reduced considerably.

To produce pixel-wise pseudo labels, standard WSOL methods~\citep{rony23} are adopted. We use the CAM ${\bm{C}}$ of the true image class $y$ from a WSOL CAM classifier pretrained on the same train dataset ${\mathbb{D}}$. Strong activations typically indicate foreground regions, while low activates point to background regions~\cite{durand2017wildcat}. Instead of directly fitting these regions, recent works showed that it is better to stochastically sample regions to avoid overfitting~\cite{Murtaza2023dips,fcam,murtaza2022dips}. To sample $n$ random foreground pixel locations, a multinomial distribution is used with ${\bm{C}}$ as its pixel-sampling probability. Typically, this allows for sampling more frequently from high-activation locations. For background sampling, ${1 - \bm{C}}$ is instead used as a sampling probability. Sampled pixels are collected in the sets ${\mathbb{C}^+, \mathbb{C}^-}$ for foreground and background pixels, respectively. Note that sampled locations change for each image at every training step. This allows to explore different regions, and it reduces overfitting to specific regions. The sampled pseudo-labels can be directly used to train our pixel classifier, using partial cross-entropy since only part of the image space ${\Omega}$ is considered at once,

\begin{equation}
\label{eq:pixel_cl}
\begin{aligned}
\min_{\bm{\theta}_1, \bm{\theta}_3} \quad &  \sum_{p \in \{\mathbb{C}^+ \; \cup\; \mathbb{C}^- \}} \mbox{H}(y^{\prime}, f(\mathsf{F}_p)) \;.
\end{aligned}
\end{equation}

\noindent The pixel pseudo-label ${y^{\prime}}$ is collected based on the CAM of the true image class ${y}$. Therefore, ${y^{\prime}}$ can be perceived as an instance of the object in that image. Our pixel-classifier learns to directly recognize the class $y$, allowing it to perform object localization.

\noindent\textbf{Total training loss}. Unlike previous works that use a two-step approach and train different models for classification and localization, \pixelcam trains a single model with two heads simultaneously. A multi-task optimization setup is considered to train for both tasks, with most parameters being shared. To this end, we use the following composite loss,

\begin{equation}
\label{eq:totla_loss}
\begin{aligned}
\min_{\bm{\theta}} \quad & \mbox{H}(y, g(\mathsf{F})) + \lambda  \sum_{p \in \{\mathbb{C}^+ \; \cup\; \mathbb{C}^- \}} \mbox{H}(y^{\prime}, f(\mathsf{F}_p)) \;,
\end{aligned}
\end{equation}

\noindent where ${\lambda}$ is a balancing coefficient. Eq.\ref{eq:totla_loss} is minimized using standard gradient descent to perform both tasks. This mitigates the convergence issue, and allows a single training cycle. It also ensures a cooperation between localization and classification to reach a better solution for both tasks at the same time.
\pixelcam does not involve a considerable computation overhead. Given our lightweight pixel classifier, the number of additional parameters is negligible compared to standard DL models. Moreover, this pixel classifier is generic and can be integrated into CNN- or transformer-based models without architectural changes. 
Our pixel-classifier creates a boundary in the pixel feature space to well separate ROIs from noise and increase the feature's discriminant power. This helps localization, but also the subsequent classification module as they use the same spatial features for decision. Consequently, it makes features robust to OOD data, allowing better performance of both tasks in such scenario.
In Section \ref{sec:results}, we show that including this linear classifier into a standard deep WSOL model, allows us to improve its localization and classification accuracy. Moreover, it provides robustness to domain shift, a common issue in WSOL models for histology image analysis~\cite{sfdawsol}.

\begin{figure}[!t]

  \includegraphics[width=\linewidth, trim=0 145 0 0, clip]{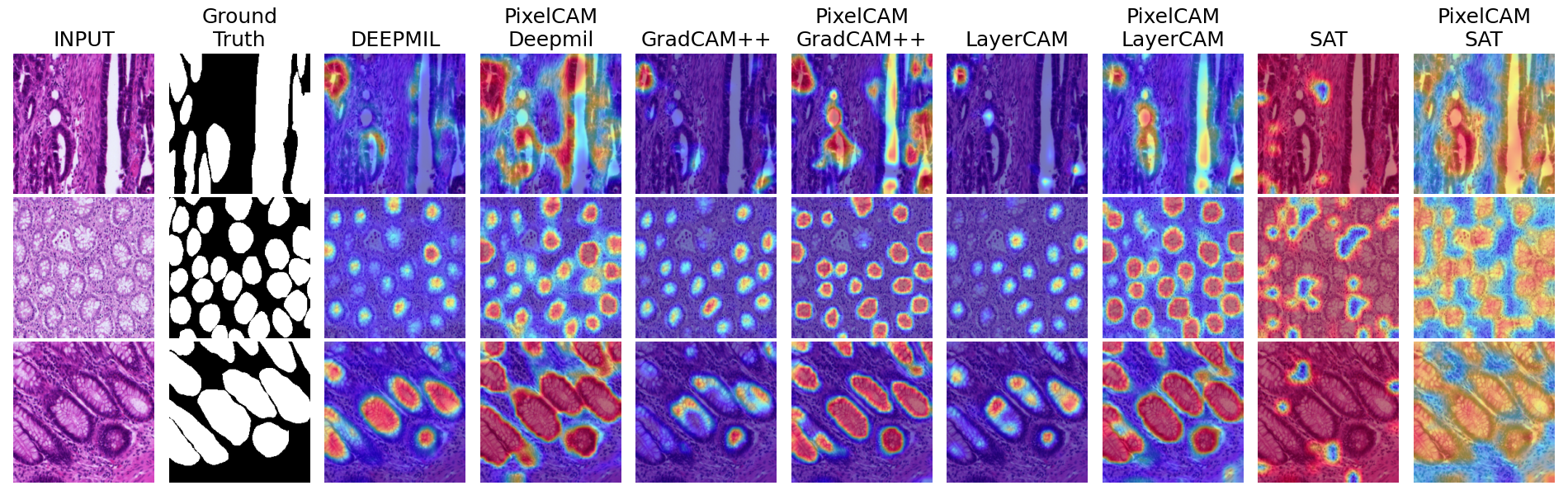}
  \caption{
  Standard WSOL setup. First column: input images from \glas. Second column: Ground truth. Next columns: We display the visual CAM results for WSOL baseline without and with \pixelcam, respectively.
  }
  \label{fig:visual-example-glas}
\end{figure}

\section{Results and Discussion} 
\label{sec:results}

\subsection[Experimental Methodology]{Experimental Methodology\footnote{A more detailed description of datasets and implementation are in Appendices ~\ref{sec:datasets-details} and~\ref{sec:training-details} , respectively.}}

\noindent\textit{(a) Datasets.} Our experiments are performed on two standard public datasets for WSOL in histology images. In particular, we use \glas dataset~\cite{sirinukunwattana2015stochastic} for colon cancer, and the patch version of \camsixteen~\cite{camelyon2016paper,rony23} for breast cancer. In both cases, we follow the same WSOL experimental protocol used in~\cite{negev,sfdawsol,rony23}.


\noindent\textit{(b) Implementation details.}
To assess the impact of our method, we first train a baseline WSOL method alone, and then contrast its performance when combined with our method. The pseudo-labels are extracted from the baseline model. 
Different recent WSOL baseline methods are used from single- and tow-step families, including DeepMIL~\cite{
deepmil}, GradCAM++~\cite{gradcampp}, LayerCAM~\cite{layercam}, SAT~\cite{SAT}, and NEGEV~\cite{negev}. For CNN-based models, we used ResNet-50 as the backbone, and for SAT, the transformer-based model DeiT-Tiny is used.
In terms of performance measures, standard WSOL metrics are used, including image classification accuracy, and pixel-localization accuracy \pxap~\cite{choe2020evaluating, fcam, negev, sfdawsol}. In addition, pixel-wise true/false positive/negative rates are used as well. For the localization task, we compare all WSOL methods to the full-supervised case using the U-Net model.


{
\setlength{\tabcolsep}{3pt}
\renewcommand{\arraystretch}{1.0}
\begin{table*}[!t]
\vspace{-4pt}
\centering
\resizebox{.7\textwidth}{!}{
\begin{tabular}{ | l  l |l | c c c | c  c c c|}
\hline
& & &  & \multicolumn{2}{c}{\glas}  & & \multicolumn{2}{c}{\camsixteen} &\\
& \textbf{WSOL models} & Venue &  & \pxap  \(\uparrow\)& \cl \(\uparrow\) &  & \pxap \(\uparrow\) & \cl \(\uparrow\)& \\
\hline \hline
&  DeepMIL~\cite{deepmil} $\dagger$ & \textit{icml}   &  & 79.9 &\textbf{100.0}& &71.3 & 85.0& \\
&  DeepMIL w/ NEGEV \cite{negev} $\star$ & \textit{midl}  &  &\textbf{85.9} & \textbf{100.0}& &\underline{75.6}&\textbf{89.9}& \\
&  DeepMIL w/ \pixelcam  $\dagger$ & - &  & \underline{85.5} &\textbf{100.0}& &\textbf{75.7}& \underline{88.2}& \\
\hline
&  GradCAM{\textit{++}}~\cite{gradcampp} $\dagger$ & \textit{wacv} &  & 76.8 &\textbf{100.0}& &49.1 & 63.4& \\
&  GradCAM{\textit{++}} w/ NEGEV \cite{negev} $\star$& \textit{midl} &  &\underline{77.1}& \textbf{100.0}& &\textbf{68.6}& \textbf{89.4}& \\
& GradCAM{\textit{++}} w/ \pixelcam  $\dagger$ & - & & \textbf{86.6} &\textbf{100.0}& &\underline{64.1} & \underline{85.1}& \\
\hline
&  LayerCAM~\cite{layercam} $\dagger$ & \textit{ieee tip} &  & \underline{75.1}&\textbf{100.0}& &33.2& 84.8& \\
&  LayerCAM w/ NEGEV\cite{negev} $\star$ & \textit{midl} &  &73.8& \textbf{100.0}& &\textbf{66.8}&\textbf{89.1}& \\
&  LayerCAM w/ \pixelcam $\dagger$ & - & & \textbf{83.6}&\textbf{100.0}& &\underline{66.2}&\textbf{89.1}& \\
\hline
&  SAT~\cite{SAT} $\dagger$ & \textit{iccv} &  &\underline{65.9} &\underline{98.8}& &\underline{32.8} & \underline{83.2}& \\
&  SAT w/ \pixelcam $\dagger$ & - &  & \textbf{79.1}&\textbf{100.0}& &\textbf{51.2}&\textbf{87.2}& \\
\hline
& U-Net~\citep{unet} & \textit{miccai} &  &95.8&n/a & &81.6 & n/a& \\
\hline
\end{tabular}
}
\caption{Localization (\pxap) and classification (\cl) accuracy on \glas and \camsixteen test sets. $\dagger$ refers to model with a one-step approach while $\star$ refers to model from two-step family.}
\label{tab:accuracy-bloc-source}
\end{table*}
}

\setlength{\tabcolsep}{3pt}
\renewcommand{\arraystretch}{1.0}
\begin{table*}[!t]
\centering
\resizebox{0.7\textwidth}{!}{
\small
\begin{tabular}{ | l  l | l | c c | c c |}
\hline
& & & \multicolumn{2}{c|}{\glas} & \multicolumn{2}{c|}{\camsixteen} \\
& \textbf{WSOL models} & \textbf{Venue} & \textbf{T-stat} $\uparrow$ & \textbf{p-value} $\downarrow$ & \textbf{T-stat} $\uparrow$ & \textbf{p-value} $\downarrow$ \\
\hline \hline
& DeepMIL~\cite{deepmil} $\dagger$ & \textit{icml} & 10.8 & $1.5 \times 10^{-11}$ & 4.8 & $4.1 \times 10^{-5}$ \\
& GradCAM{\textit{++}}~\cite{gradcampp} $\dagger$ & \textit{wacv} & 29.1 & $1.8 \times 10^{-22}$ & 17.9 & $7.4\times 10^{-17}$ \\
& LayerCAM~\cite{layercam} $\dagger$ & \textit{ieee tip} & 12.7 & $4.0 \times 10^{-13}$ & 26.5 & $2.2 \times 10^{-21}$ \\
& SAT~\cite{SAT} $\dagger$ & \textit{iccv} & 58.3 & $8.9 \times 10^{-31}$ &99.7 & $2.8 \times 10^{-37}$ \\
\hline
\end{tabular}
}
\caption{T-test statistics between baselines (DeepMIL,GradCAM{\textit{++}},LayerCAM,SAT) and our method (\pixelcam) for localization performance.}
\label{tab:tstat-results}
\end{table*}

\subsection{WSOL Results}

We first evaluate \pixelcam using the standard WSOL protocol over  \glas and \camsixteen dataset~\cite{rony23}. 

Results are reported in Tab.~\ref{tab:accuracy-bloc-source}.
Combining our method with a WSOL baseline often provides an improvement in localization accuracy. 

\begin{figure}[!htb]
\centering
\includegraphics[width=0.99\linewidth]{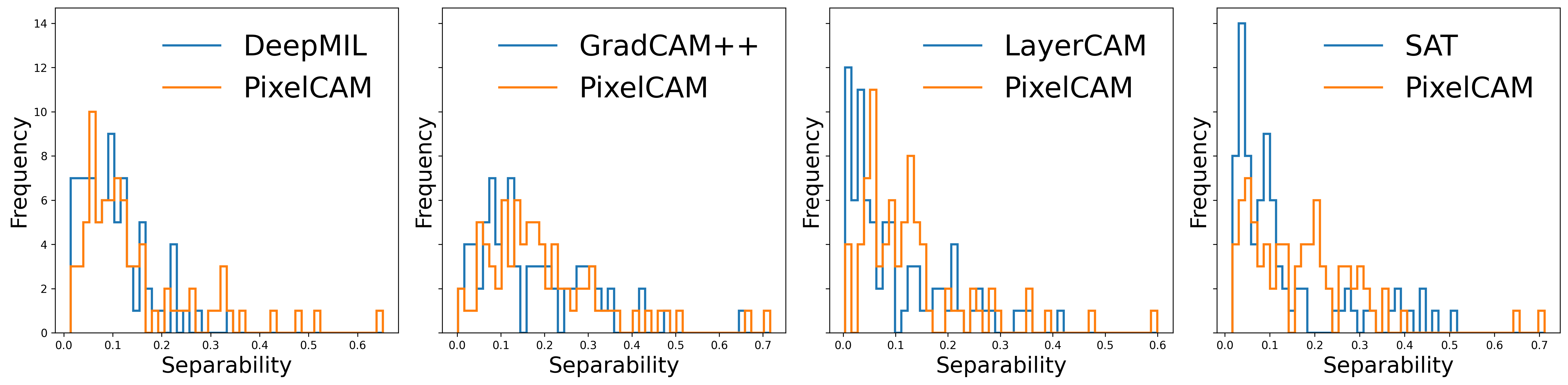}
\caption{Histogram of foreground/background separability index for WSOL baseline methods alone vs when combined with \pixelcam on \glas test set. The higher the value, the better the separability is.
The definition of class separability is in Appendix~\ref{sec:class_separability}.
}
  \label{fig:histogram_glas_separability}
\end{figure}

Over \glas dataset, our method improves the \pxap performance by (+5.6\%,+9.8\%,+8.5\%) compared to the WSOL baseline methods (DeepMIL, GradCAM++, LayerCAM) respectively, while on \camsixteen dataset, our method gains +4.4\%,+15.0\%,+33.0\%, respectively. Fig.~\ref{fig:visual-example-glas} shows visual prediction examples. The localization performance gains are supported by statistical significance, as confirmed by t-tests, in Tab.~\ref{tab:tstat-results}, yielding very low p-values (below 0.05).
Additionally, our method improves classification performance as well. This indicates that well-separated classes at pixel-feature level in our model help improve localization but also facilitate global image classification. We further provide measures of how well are separated FG and BG pixel-features in Fig~\ref{fig:histogram_glas_separability} on the \glas test set. These results show that \pixelcam histogram of class separability index is shifted to the right compared to the baseline. This indicates a better class separability.

\subsection{Out-of-Distribution Results}
{
\setlength{\tabcolsep}{3pt}
\renewcommand{\arraystretch}{1.0}
\begin{table*}[!t]
\centering
\resizebox{0.7\textwidth}{!}{
\small
\begin{tabular}{ | l  l | l | c c c | c  c c c|}
\hline
& & &  & \multicolumn{2}{c}{\camsixteen $\rightarrow$ \glas}  & & \multicolumn{2}{c}{\glas  $\rightarrow$ \camsixteen} &\\
& \textbf{WSOL models} & \textbf{Venue} &  & \pxap \(\uparrow\) & \cl \(\uparrow\) &  & \pxap \(\uparrow\)& \cl \(\uparrow\)& \\
\hline \hline
&  DeepMIL~\cite{deepmil} $\dagger$& \textit{icml} &  &\underline{64.5} &\underline{81.2}& &\underline{29.0} &\textbf{55.2}& \\
&  DeepMIL w/ NEGEV\cite{negev} $\star$& \textit{midl} &  &62.7 &\underline{81.2}& &27.8 &\underline{52.9}& \\
&  DeepMIL w/ \pixelcam $\dagger$& - &  &\textbf{69.1} &\textbf{83.8}& &\textbf{30.2} &52.5& \\
\hline
&  GradCAM{\textit{++}}~\cite{gradcampp} $\dagger$& \textit{wacv} &  & 52.9 &53.7& &\textbf{39.1} &52.4& \\
&  GradCAM{\textit{++}} w/ NEGEV\cite{negev} $\star$& \textit{midl} &  &\textbf{64.5} &\textbf{75.0}& &24.2 &\underline{56.7}& \\
&   GradCAM{\textit{++}} w/ \pixelcam $\dagger$& - &  & \underline{56.2} &\underline{71.2}& &\underline{36.9} & \textbf{63.3}& \\
\hline
&  LayerCAM~\cite{layercam} $\dagger$& \textit{ieee tip} &  & 59.5 &77.5& &30.4 &51.4& \\
&   LayerCAM w/ NEGEV\cite{negev} $\star$& \textit{midl} &  & \underline{63.9} &\textbf{83.7}& &\underline{27.9}& \underline{52.3}& \\
&   LayerCAM w/ \pixelcam $\dagger$& - &  & \textbf{67.2} &\underline{78.8}& &\textbf{34.8}& \textbf{55.8}& \\
\hline
&  SAT~\cite{SAT} $\dagger$& \textit{iccv} &  & 50.7 &67.5& &\textbf{24.3} &50.2& \\
&   SAT w/ \pixelcam $\dagger$& - &  & \textbf{55.1} &\textbf{78.8}& &22.5&\textbf{50.4}& \\
\hline
\end{tabular}
}
\caption{
OOD results: Localization (\pxap) and classification (\cl) accuracy on target test dataset: \glas and \camsixteen. The symnbol "$\dagger$" refers to model with a one-step approach, while "$\star$" refers to model from two-step family.}
\label{tab:accuracy-bloc-target}
\end{table*}
}

Additional experiments were conducted to assess the ability of \pixelcam to train robustness models on OOD data. To this end, consider the domain shift between \glas (colon cancer) and \camsixteen (breast cancer).
This simulates a domain adaptation problem in the source-only case, where a model is pre-trained on source domain data, and then evaluated on target data. Results are reported in Tab.~\ref{tab:accuracy-bloc-target}. As shown in~\cite{sfdawsol}, WSOL methods typically decline when facing OOD in histology data. Overall, our \pixelcam method yields better performance over both tasks and across all baselines. This is mainly due to better separated pixel-features learned by our model which make them more robust to OOD data. This is demonstrated in the appendix~\ref{sec:class_separability} by inspecting the class separability of pixel-features of the target data. As presented in Tab.~\ref{tab:accuracy-bloc-target}, using \pixelcam outperforms in general WSOL baselines alone on the case \mbox{\camsixteen $\rightarrow$ \glas}. This improves \pxap scores by 4.6\%, 3.3\%, 7.7\%, and 4.4\% compared to DeepMIL, GradCAM++, LayerCAM, and SAT, respectively. Moreover, \pixelcam also improves image classification performance. The other case, \mbox{\glas $\rightarrow$ \camsixteen}, is much more challenging since the train source dataset is very small, compared to the very large and difficult target set. While the overall performance is low compared to when \glas is the target, our method still yields better localization performance in general. Moreover, classification performance is improved overall across both scenarios.
More results and interpretations are included in Appendices ~\ref{sec:ablations} and ~\ref{sec:appendix_visualization}.

We further evaluate the robustness of \pixelcam by altering the stainings in the test sets of \glas  using stain styles from the other dataset. The stainings were progressively modified by selecting stain variations ordered by their distance to the original stain. Stain 1 introduces a minor alteration, while stain 10 represents a substantial shift. This setup simulates an increasing staining shift while keeping the organ type unchanged. Additional results for \camsixteen are presented in Appendices~\ref{sec:staining}.

\begin{figure}[!htbp]
  \centering
  \includegraphics[width=0.99\linewidth]{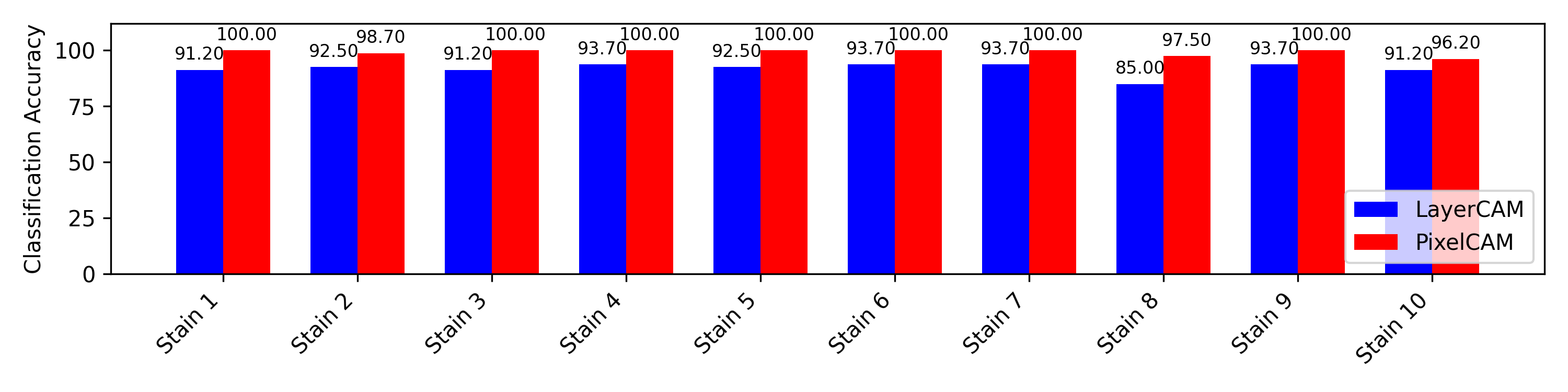}
  \caption{Classification (\cl) accuracy on \glas test sets with LayerCAM and \pixelcam with different stainings.
  }
  \label{fig:stain-cl-glas}
\end{figure}
\begin{figure}[!htbp]
  \centering
  \includegraphics[width=0.99\linewidth]{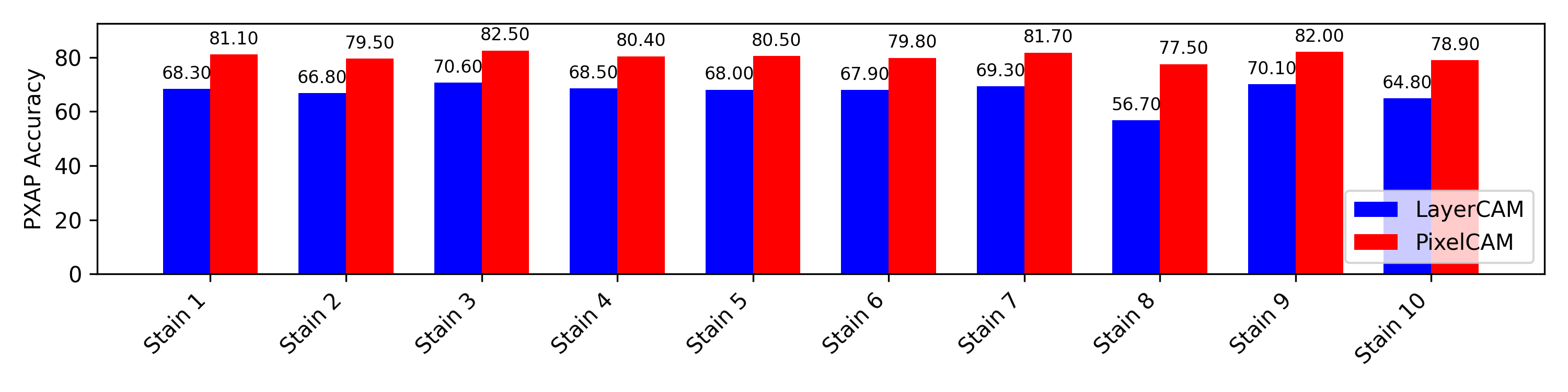}
  \caption{Localization (\pxap) accuracy on \glas test sets with LayerCAM and \pixelcam with different stainings.
  }
  \label{fig:stain-loc-glas}
\end{figure}
\section{Conclusion} 
\label{sec:conclusion}
In this paper, \pixelcam is introduced for multi-task single-step WSOL method. It is based on a cost-effective FG/BG pixel-wise classifier in the pixel-feature space, allowing for spatial object localization. It leverages pixel-pseudo labels cues for localization. It aims at explicitly learning discriminant pixel-features and accurately separate FG/BG regions, promoting better ROI localization and image classification. Both tasks are optimized in parallel without incurring notable computational cost nor additional parameters. 
%
Our pixel classifier is versatile. It can easily be integrated into any CNN- or transform-based classifier architecture without modification.
%
Results on two histology datasets showed that \pixelcam can improve WSOL baseline methods over both tasks, but also make them robust to OOD data. However, there is still room to improve our method by better optimizing both tasks while reducing their mutual negative impact. In addition, dealing with OOD in WSOL scenario is still an ongoing issue, and our method still requires more improvements despite its progress. 

\newpage
\subsubsection*{Acknowledgements}{This research was supported in part
by the Canadian Institutes of Health Research, the Natural
Sciences and Engineering Research Council of Canada, and
the Digital Research Alliance of Canada.
}

\newpage


\appendix
\addcontentsline{toc}{section}{Appendix} 
\part{Appendix} 
\parttoc 


\section{Related Work} 
\label{sec:related}

WSOL has emerged as a low-cost training setup to localize objects while classifying an image content~\cite{choe2020evaluating,rpwsol,zhou2017brief}. WSOL has also been extended to videos~\cite{belharbi23tcam,belharbi25colocam}. Moreover, there has been a recent focus on designing WSOL methods for histology image analysis~\cite{rony23}. This allows us to reduce the large annotation cost, in addition to building visually interpretable classifiers.

\noindent \textbf{Single-step WSOL}. Several approaches address the WSOL problem in a single step~\cite{choe2020evaluating,rpwsol,rony23} where a single model is trained to do both tasks, classification and localization. Usually, this is achieved by using a localization head followed by a spatial pooling layer to extract per-class probability. Early~\cite{deepmil,gradcam,zhou2016learning} but also recent WSOL works~\cite{SAT,zhu24,tscam} follow this strategy. Although this has brought significant improvements into the field, these models still face several limitations. Since only image-class labels are used as supervision without any localization cues, CAM localization can have poor performance, more so when dealing with less salient objects such as in histology images. Recent work~\cite{rony23} showed that without localization cues over histology images, these models often lead to highly unbalanced localization. This manifests either by under- or over-activations leading to high false negative/positive rates. In addition, this approach faces a great challenge in model selection as both tasks proceed in an asynchronous convergence~\cite{choe2020evaluating,rpwsol,rony23}. For instance, the model selected for best localization often yields poor classification.

\noindent \textbf{Two-step WSOL}.This strategy has emerged as a new line of research. In particular, a dual-model approach is used in combination with the usage of localization cues under the form of pseudo-labels~\cite{rpwsol,rony23}. This provides direct guidance for the localization task, bypassing the issue of task convergence. Several works simply create a dedicated model per task: one for classification and another for localization~\cite{wei2021shallowspol,zhang2020rethinking,zhao2023generative}. This leads to better performance since tasks are divided between two large models. However, this is a cumbersome approach as it drastically increases the number of parameters and training cycles. Most importantly, the localization CAMs are completely disconnected from the classification decision, making this strategy unreliable and less interpretable. A parallel direction overcomes this issue by using a decoder to act as a localizer on top of the classifier~\cite{negev,fcam,Murtaza2023dips}, in a similar way to a U-Net architecture~\cite{unet}. This creates a direct relation between both tasks. However, the decoder has a limited learning capacity since it is tied to a frozen encoder at several layers via skip connections. This prevents the localizer from a better adaptation, and it limits its performance leading to a sub-optimal solution. In addition, it adds a significant number of parameters to the classifier.

\noindent \textbf{WSOL vs OOD}. In addition to these limitations, WSOL methods face challenges when dealing with domain shift which degrades the performance of both tasks~\cite{sfdawsol}. Such shift is common in histology due to variations in stains, objects' structure, microscope type, and imaging centers. Further analysis shows that this issue could be rooted in the pixel features of the image encoder. Both classification and localization tasks depend heavily on these spatial features. Less flexible and poorly discriminant pixel features can lead to poor performance in both tasks since they both rely on these embeddings.

In summary, while existing WSOL methods have achieved great progress they still face several limitations. Single-step WSOL lacks the leverage of localization cues making them vulnerable to wrong localization when dealing with complex and less salient data such as histology images. This adds to the  well known issue of asynchronous convergence of localization and classification tasks. On the other hand, two-step methods are cumbersome in terms of training cycles, and number of parameters. In addition, either both tasks are disconnected leading to misaligned decisions or the localization has a tied capacity due to frozen backbone.
Our method comes as simple yet efficient alternative. It performs WSOL in a single-step within a multi-task framework where both tasks are optimized simultaneously. This leads to a single training cycle and facilitate convergence issue. It allows sharing parameters between both tasks making it parameter-efficient. But also, it can leverage localization cues such as pseudo-labels. Finally, well separating features at pixel level makes our model robust to OOD data.

\section{Datasets}
\label{sec:datasets-details}

\paragraph{GlaS.}  
The \glas  dataset is used for the diagnosis of colon cancer. The dataset consists of 165 images from 16 Hematoxylin and Eosin (H\&E) and includes labels at both pixel-level and image-level (benign or malign). The dataset consists of 67 images for training, 18 for validation, and 80 for testing.  We use the same protocol as in~\citep{rony23, sfdawsol, negev}. Similarly, we use 3 samples per class with full supervision in validation set for model selection for localization (\beloc). 

\paragraph{CAMELYON16.} 
A patch-based benchmark~\cite{rony23} is extracted from the \camsixteen dataset that contains 399 Whole slide images categorized into two classes (normal and metastasic) for the detection of breast cancer metastases in H\&E-stained tissue sections of sentinel lymph nodes. Patch extraction of size ${512\times 512}$ follows a protocol established by~\citep{rony23, sfdawsol, negev} to obtain patches with annotations at the image and pixel level. 
The dataset contains a total of 48870 images, including 24348 for training, 8850 for validation, and 15664 for testing. From the validation dataset, 6 examples per class are randomly selected to be fully supervised to perform model selection for localization (\beloc) similarly to~\citep{rony23, sfdawsol, negev}.

\section{WSOL Training Details}
\label{sec:training-details}

For pretraining a WSOL baseline method on the data, we use the same setup as in~\cite{sfdawsol}. 
The first part of the model training is defined using the backbone trained on ImageNet~\cite{ImageNet}. The training is done using SGD with a batch size of 32~\cite{sfdawsol}. For the \glas dataset, training is performed over 1000 epochs, and 20 epochs for \camsixteen. A weight decay of $10^{-4}$ is also used. During training, images are resized to  ${256\times 256}$, then randomly cropped to ${224\times 224}$. A hyperparameter search was conducted for the learning rate parameters among the values \{0.0001, 0.001, 0.01\}, and its decaying factor among \{0.1, 0.4, 0.9\} following~\cite{sfdawsol}. In the second phase of training \pixelcam, we use the CAMs generated by the previous method and continue the training using the same setup as defined previously. 
For the OOD scenario, the source model is trained using the same setup on a source dataset and evaluated on a target dataset.


\section{Class Separability Index}
\label{sec:class_separability}

In this section, we present the class separability index between foreground (FG) and background (BG) classes at pixel-feature level over test set of both datasets \glas and \camsixteen. To measure how well FG/BG classes are separated in the feature space of pixels, we resort to the class separability index~\cite{Duda2000separability}.
The class separability, $J$, is based on the Within-class scatter matrix ($\bm{S}_{W}$) and the Between-class scatter matrix ($\bm{S}_{B}$) of pixel-features, defined as follows,

\begin{equation}
\bm{S}_W = \sum_{i=1}^{c} \left[ \sum_{j=1}^{n_i} (\bm{x}_{i,j} - \bm{m}_i)(\bm{x}_{i,j} - \bm{m}_i)^\top \right]\, ,
\end{equation}

\begin{equation}
\bm{S}_B = \sum_{i=1}^{c} n_i (\bm{m}_i - \bm{m})(\bm{m}_i - \bm{m})^\top \, ,
\end{equation}

\noindent where $c$ is the number of class (in our case ${c=2}$ for foreground and background). Note that this index $J$ is computed over a single image using the pixel-labels. 
Let $n_{i}\; (i=0,..,c)$ be the number of pixels in the $i$-th class. The feature vector $\bm{x}_{i,j} \in \mathbb{R}^{d}$ denotes the $j$-th pixel of the $i$-th class.  It represents the vector ${\mathsf{F}_p}$ for the pixel $p$ in the main paper. The mean vector of the $i$-th class is given by 
$\bm{m_{i}}$, while $\bm{x}$ represents the mean vector computed over all feature vectors. The class separability denoted $J$ is defined as follows~\cite{Duda2000separability},

\begin{equation}
J = \frac{\tr(\bm{S}_B)}{\tr(\bm{S}_W)}\,.
\end{equation}

\begin{figure}[!htbp]
\centering
\includegraphics[width=0.999\linewidth]{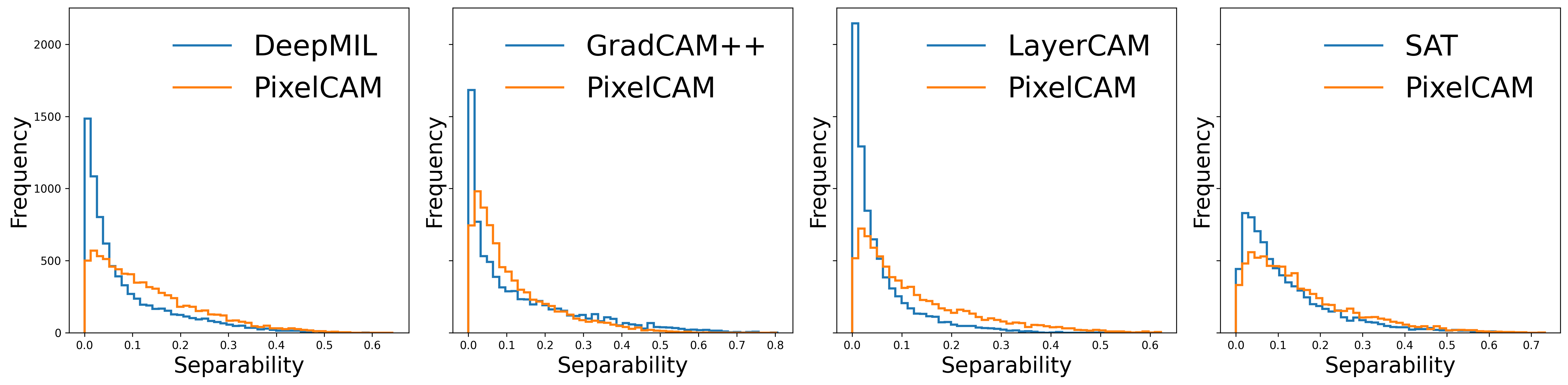}
\caption{Standard WSOL setup: Histogram of class separability index $J$ between FG/BG pixel-features on test set \camsixteen for WSOL baselines with and without \pixelcam. The higher $J$ is, the more classes are separated.}
\label{fig:histogram_camelyon_separability_source}
\end{figure}

\begin{figure}[!htbp]
\centering
\includegraphics[width=0.999\linewidth]{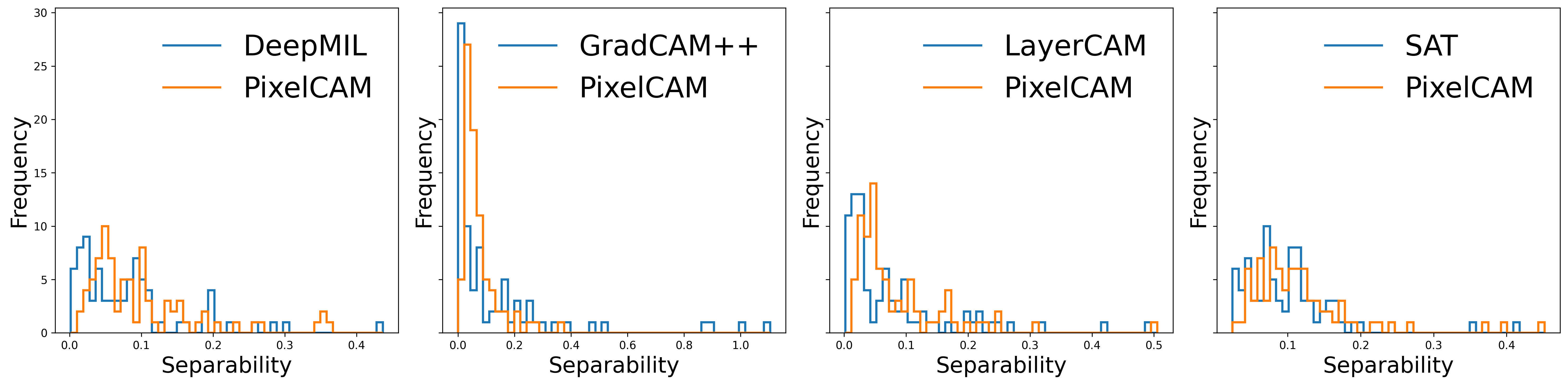}
\caption{OOD setup: Histogram of class separability index $J$ between FG/BG pixel-features on target test set \glas for the OOD case: \mbox{\camsixteen $\rightarrow$ \glas}  for WSOL baselines with and without \pixelcam. The higher $J$ is, the more classes are separated.}
  \label{fig:all_histogram_target_glas}
\end{figure}

\begin{figure}[!htbp]
\centering
\includegraphics[width=0.999\linewidth]{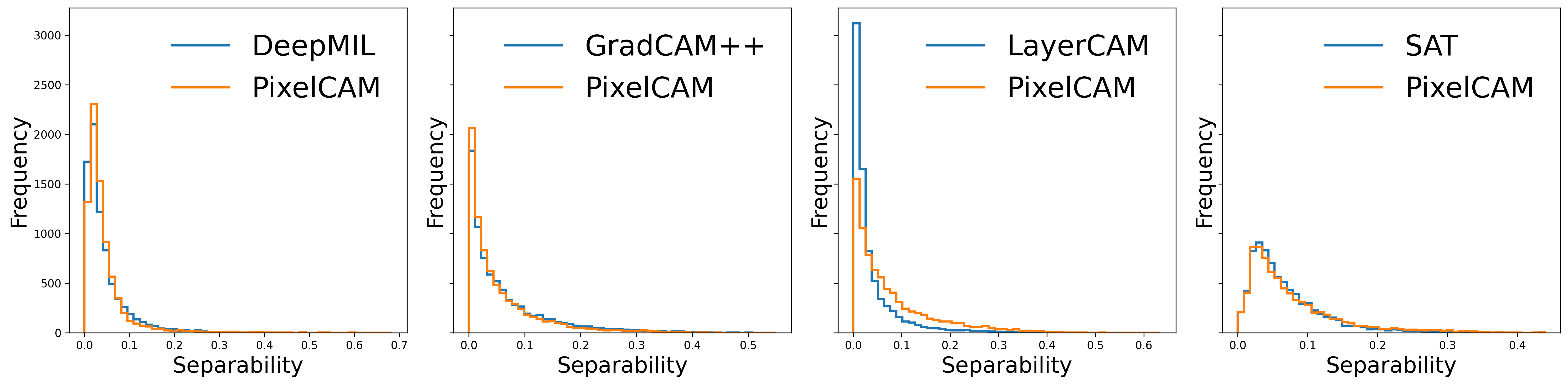}
\caption{OOD setup: Histogram of class separability index $J$ between FG/BG pixel-features on target test set \camsixteen for the OOD case: \mbox{\glas $\rightarrow$ \camsixteen}  for WSOL baselines with and without \pixelcam. The higher $J$ is, the more classes are separated.}
  \label{fig:all_histogram_target_camelyon}
\end{figure}

In Table~\ref{tab:accuracy-class-separability-bloc-source} and \ref{tab:accuracy-class-separability-bloc-target}, we provide the average class separability over normal and cancer classes and the overview over the entire dataset.  As we can observe, \pixelcam improves the class separability between FG and BG of WSOL baseline methods on both dataset except for GradCAM++ on \camsixteen. This separability is explained by the high number of images with a low separability as illustrated in Fig~\ref{fig:histogram_camelyon_separability_source}. The improvement of \pixelcam in the OOD scenario compared to the WSOL baseline methods such as DeepMIL and LayerCAM on \glas and \camsixteen can be attributed to better separability as shown in Figs~\ref{fig:all_histogram_target_glas} and \ref{fig:all_histogram_target_camelyon}. We provide examples of features separability at the pixel level in section~\ref{sec:appendix_visualization}.

\setlength{\tabcolsep}{3pt}
\renewcommand{\arraystretch}{1.0}
\begin{table}[!htbp]
\centering 
\resizebox{0.48\textwidth}{!}{
\small
\begin{tabular}{ | l  l | c c c | c c c |}
\hline
& & \multicolumn{3}{c|}{\glas} & \multicolumn{3}{c|}{\camsixteen} \\
 & \textbf{WSOL models} & Normal & Cancer & All & Normal & Cancer & All \\
\hline \hline

& DeepMil~\cite{deepmil}{\small \emph{(icml)}} & 0.13 & 0.07 & 0.10 & - & 0.11 & 0.11 \\
& DeepMil w/ \pixelcam & \textbf{0.21} & \textbf{0.09} & \textbf{0.14} & - & \textbf{0.17} & \textbf{0.17} \\
\hline
& GradCAM{\textit{++}}~\cite{gradcampp}{\small \emph{(wacv)}} & 0.22 & 0.13 & 0.17 & - & \textbf{0.22} & \textbf{0.22} \\
& GradCAM{\textit{++}} w/ \pixelcam & \textbf{0.25} & \textbf{0.15} & \textbf{0.20} & - & 0.16 & 0.16 \\
\hline
& LayerCAM~\cite{layercam}{\small \emph{(ieee tip)}} & \textbf{0.17}& 0.04 & 0.10 & - & 0.07 & 0.07 \\
& LayerCAM w/ \pixelcam & 0.14 & \textbf{0.12} & \textbf{0.13} & - & \textbf{0.17} & \textbf{0.17} \\
\hline
& SAT~\cite{SAT}{\small \emph{(iccv)}} & 0.21 & 0.06 & 0.13 & - & 0.17 & 0.17 \\
& SAT w/ \pixelcam & \textbf{0.25} & \textbf{0.10} & \textbf{0.17} & - & \textbf{0.22} & \textbf{0.22} \\
\hline
\end{tabular}
}
\caption{Standard WSOL setup: Average class separability index $J$ between FG/BG pixel-features on test set \glas and \camsixteen for WSOL baselines with and without \pixelcam. The higher $J$ is, the more classes are separated.}
\label{tab:accuracy-class-separability-bloc-source}
\vspace{-1em}
\end{table}

\setlength{\tabcolsep}{3pt}
\renewcommand{\arraystretch}{1.0}
\begin{table}[!htbp]
\centering 
\resizebox{0.48\textwidth}{!}{
\small
\begin{tabular}{ | l  l | c c c | c c c |}
\hline
& & \multicolumn{3}{c|}{\camsixteen $\rightarrow$ \glas} & \multicolumn{3}{c|}{\glas $\rightarrow$ \camsixteen} \\
 &\textbf{WSOL models} & Normal & Cancer & All & Normal & Cancer & All \\
\hline \hline

& DeepMil~\cite{deepmil}{\small \emph{(icml)}} & 0.05 & 0.11 & 0.08 & - & \textbf{0.05} & \textbf{0.05} \\
& DeepMil w/ \pixelcam & \textbf{0.06} & \textbf{0.13} & \textbf{0.10} & - & \textbf{0.05} & \textbf{0.05} \\
\hline
& GradCAM{\textit{++}}~\cite{gradcampp}{\small \emph{(wacv)}} & \textbf{0.08} & \textbf{0.19} & \textbf{0.14} & - & \textbf{0.08} & \textbf{0.08} \\
& GradCAM{\textit{++}} w/ \pixelcam & 0.05 & 0.09 & 0.07 & - & 0.07 & 0.07 \\
\hline
& LayerCAM~\cite{layercam}{\small \emph{(ieee tip)}} & 0.02 & \textbf{0.13} & 0.08 & - & 0.04 & 0.04 \\
& LayerCAM w/ \pixelcam & \textbf{0.06} & 0.11 & \textbf{0.09} & - & \textbf{0.09} & \textbf{0.09} \\
\hline
& SAT~\cite{SAT}{\small \emph{(iccv)}} & 0.11 & 0.09 & 0.10 & - & 0.08 & 0.08 \\
& SAT w/ \pixelcam & \textbf{0.12} & \textbf{0.12} & \textbf{0.12} & - & \textbf{0.09} & \textbf{0.09} \\
\hline
\end{tabular}
}
\caption{OOD setup: Average class separability index $J$ between FG/BG pixel-features on target test set \glas and \camsixteen for both OOD cases: \mbox{\camsixteen $\rightarrow$ \glas} and \mbox{\glas $\rightarrow$ \camsixteen} for WSOL baselines with and without \pixelcam. The higher $J$ is, the more classes are separated.}
\label{tab:accuracy-class-separability-bloc-target}
\end{table}


\section{More Localization Performance Analysis}

\pxap is the primary metric used to measure localization performance. However, following~\cite{rony23}, we include other pixel-wise performance measures  that is true/false positives/negative rates.
As shown previously, our method \pixelcam achieves better results in terms of \pxap performance. This improvement is observed in Tab.~\ref{tab:mtx-conf-best-loc-glas-camelyon} with a higher number of true positive/negative rates compared to GradCAM++, LayerCAM, SAT, and NEGEV on the \glas dataset.
For the \camsixteen dataset, we observe that \pixelcam increases the true positive rate compared to standard methods while being competitive to NEGEV. However, we note a significant improvement when applying \pixelcam to the SAT method, with a considerable increase in true positives, thereby reducing the over-activation issue.

{
\setlength{\tabcolsep}{3pt}
\renewcommand{\arraystretch}{1.1}
\begin{table*}[ht]
\centering
\resizebox{0.7\textwidth}{!}{%
\centering
\small
\begin{tabular}{l l c*{4}{c}c*{4}{c}{c}}
& &  & \multicolumn{4}{c}{\glas} & & \multicolumn{4}{c}{\camsixteen} & \\
\cline{1-2}\cline{4-7}\cline{9-12} \\
\textbf{Bottom-up WSOL} & Venue &  & \pxtp \(\uparrow\)& \pxfn \(\downarrow\)& \pxtn \(\uparrow\)& \pxfp  \(\downarrow\)&  & \pxtp \(\uparrow\)& \pxfn \(\downarrow\)& \pxtn \(\uparrow\)& \pxfp \(\downarrow\)  \\
\cline{1-2}\cline{4-7}\cline{9-12}\\
DeepMIL~\citep{deepmil} $\dagger$ & \textit{icml} &  & 63.4  & 36.6 & \underline{76.3} & \underline{23.6} &  & \textbf{66.9} & \textbf{33.1}& 89.6 &10.4&  \\
DeepMIL w/ NEGEV~\citep{negev} $\star$& \textit{midl} &  & \textbf{79.0}  & \textbf{21.0} & 75.5 & 24.5 &  & \underline{64.0} & \underline{36.0}& \underline{92.6}  &\underline{7.4}&  \\
DeepMIL w/ \pixelcam $\dagger$ & - &  & \underline{76.4} & \underline{23.6}  & \textbf{77.4} & \textbf{22.6} &  &59.0& 41.0& \textbf{93.3}&\textbf{6.7}&  \\
\cline{1-2}\cline{4-7}\cline{9-12}\\
GradCAM++~\citep{gradcampp} $\dagger$ & \textit{wacv} &  & \underline{62.0}  & \underline{38.0} & \textbf{79.9} &\textbf{20.1} &  & 42.1&57.8&  89.4 &10.6&  \\
GradCAM w/ NEGEV~\citep{negev} $\star$& \textit{midl} &  & 58.4  & 41.6 & 76.0 &24.0&  & \textbf{53.5}& \textbf{46.5}& \textbf{92.2}  & \textbf{7.8}&  \\
GradCAM w/ \pixelcam $\dagger$ & - &  & \textbf{76.9}  & \textbf{23.1} & \underline{78.8} & \underline{21.2} &  & \underline{49.2} & \underline{50.8} &  \underline{91.4} & \underline{8.6}&  \\
\cline{1-2}\cline{4-7}\cline{9-12}\\
LayerCAM~\citep{layercam} $\dagger$ & \textit{ieee tip} &  & 62.3  & 37.7 & \textbf{72.6} &  \textbf{27.4} &  &29.6& 70.3& 86.8 &13.2&  \\
LayerCAM w/ NEGEV~\citep{negev} $\star$& \textit{midl} &  & \underline{66.3}  & \underline{33.7} & 70.7 & 29.3 &  & \underline{54.4} & \underline{45.6}& \textbf{90.9}  & \textbf{9.1}&  \\
LayerCAM w/ \pixelcam $\dagger$ & - &  & \textbf{77.6}  & \textbf{22.4} & \underline{72.5} & \underline{27.5} &  & \textbf{56.5}& \textbf{43.5}& \underline{89.2}  &\underline{10.8}&  \\
\cline{1-2}\cline{4-7}\cline{9-12}\\
\cline{1-2}\cline{4-7}\cline{9-12}\\
U-Net~\citep{unet} & \textit{miccai} &  & 88.9  & 11.1 & 89.8  & 10.2  &  & 68.0& 32.0 & 94.5 &5.5&  \\
\cline{1-2}\cline{4-7}\cline{9-12}\\
\end{tabular}
}
\caption{Confusion matrix performance over \glas and \camsixteen test set with standard WSOL setup. $\dagger$ refers to model with a single-step approach while $\star$ refers to model from two-step family.}
\label{tab:mtx-conf-best-loc-glas-camelyon}
\end{table*}
}

\section{Additional Results With Different Backbones}
\setlength{\tabcolsep}{3pt}
\renewcommand{\arraystretch}{1.0}
\begin{table}[ht!]
\centering
\resizebox{.48\textwidth}{!}{
\begin{tabular}{ | l  l |l | c c c | c  c c c|}
\hline
& & &  & \multicolumn{2}{c}{\glas}  & & \multicolumn{2}{c}{\camsixteen} &\\
& \textbf{WSOL models} & Venue &  & \pxap  \(\uparrow\)& \cl \(\uparrow\) &  & \pxap \(\uparrow\) & \cl \(\uparrow\)& \\
\hline \hline
&  DeepMIL~\cite{deepmil} $\dagger$ & \textit{icml}   &  & 73.3 &\textbf{100.0}& &60.3 & \textbf{88.0}& \\
&  DeepMIL w/ NEGEV \cite{negev} $\star$ & \textit{midl}  &  &\underline{76.7} & \textbf{100.0}& &\underline{62.2}&86.6& \\
&  DeepMIL w/ \pixelcam  $\dagger$ & - &  & \textbf{78.6} &\textbf{100.0}& &\textbf{69.9}& \underline{85.4}& \\
\hline
&  GradCAM{\textit{++}}~\cite{gradcampp} $\dagger$ & \textit{wacv} &  & 72.9 &\textbf{100.0}& &27.7 & \underline{87.4}& \\
&  GradCAM{\textit{++}} w/ NEGEV \cite{negev} $\star$& \textit{midl} &  &\textbf{80.4}& \textbf{100.0}& &\textbf{68.3}& 87.3& \\
& GradCAM{\textit{++}} w/ \pixelcam  $\dagger$ & - & & \underline{76.9} &\textbf{100.0}& &\underline{64.9} & \textbf{87.7}& \\
\hline
&  LayerCAM~\cite{layercam} $\dagger$ & \textit{ieee tip} &  & 73.0&\textbf{100.0}& &24.5& \underline{88.7}& \\
&  LayerCAM w/ NEGEV\cite{negev} $\star$ & \textit{midl} &  &\textbf{80.7}& \textbf{100.0}& &\textbf{68.7}&\textbf{88.1}& \\
&  LayerCAM w/ \pixelcam $\dagger$ & - & & \underline{76.3}&\textbf{100.0}& &\underline{64.2}&87.1& \\
\hline
\end{tabular}
}
\caption{Localization (\pxap) and classification (\cl) accuracy on \glas and \camsixteen test sets using different WSOL methods with VGG16 architecture. $\dagger$ refers to model with a one-step approach while $\star$ refers to model from two-step family.}
\label{tab:accuracy-bloc-source-vgg16}
\end{table}

\setlength{\tabcolsep}{3pt}
\renewcommand{\arraystretch}{1.0}
\begin{table}[ht!]
\centering
\resizebox{.48\textwidth}{!}{
\begin{tabular}{ | l  l |l | c c c | c  c c c|}
\hline
& & &  & \multicolumn{2}{c}{\glas}  & & \multicolumn{2}{c}{\camsixteen} &\\
& \textbf{WSOL models} & Venue &  & \pxap  \(\uparrow\)& \cl \(\uparrow\) &  & \pxap \(\uparrow\) & \cl \(\uparrow\)& \\
\hline \hline
&  DeepMIL~\cite{deepmil} $\dagger$ & \textit{icml}   &  & 62.6 &87.5& &58.1 & \underline{89.0}& \\
&  DeepMIL w/ NEGEV \cite{negev} $\star$ & \textit{midl}  &  &\underline{66.1} & \textbf{91.2}& &\textbf{76.8}&\textbf{89.5}& \\
&  DeepMIL w/ \pixelcam  $\dagger$ & - &  & \textbf{72.0} &\underline{88.8}& &\underline{75.7}& 83.2& \\
\hline
&  GradCAM{\textit{++}}~\cite{gradcampp} $\dagger$ & \textit{wacv} &  & 70.0 &53.8& &64.0 & \underline{88.4}& \\
&  GradCAM{\textit{++}} w/ NEGEV \cite{negev} $\star$& \textit{midl} &  &\textbf{81.7}& \underline{93.7}& &\underline{74.1}& 86.9& \\
& GradCAM{\textit{++}} w/ \pixelcam  $\dagger$ & - & & \underline{81.0} &\textbf{96.3}& &\textbf{75.0} & \textbf{89.8}& \\
\hline
&  LayerCAM~\cite{layercam} $\dagger$ & \textit{ieee tip} &  & 68.3&92.5& &53.7& \textbf{83.8}& \\
&  LayerCAM w/ NEGEV\cite{negev} $\star$ & \textit{midl} &  &\underline{77.8}& \underline{93.8}& &\textbf{71.6}&\textbf{83.8}& \\
&  LayerCAM w/ \pixelcam $\dagger$ & - & & \textbf{78.3}&\textbf{100.0}& &\underline{65.5}&\underline{82.5}& \\
\hline
\end{tabular}
}
\caption{Localization (\pxap) and classification (\cl) accuracy on \glas and \camsixteen test sets using different WSOL methods with InceptionV3 architecture. $\dagger$ refers to model with a one-step approach while $\star$ refers to model from two-step family.}
\label{tab:accuracy-bloc-source-inceptionv3}
\vspace{-2.5em}
\end{table}
\setlength{\tabcolsep}{3pt}
\renewcommand{\arraystretch}{1.0}
\begin{table}[ht!]
\centering
\resizebox{0.48\textwidth}{!}{
\small
\begin{tabular}{ | l  l | l | c c | c c |}
\hline
& & & \multicolumn{2}{c|}{\glas} & \multicolumn{2}{c|}{\camsixteen} \\
& \textbf{WSOL models} & \textbf{Venue} & \textbf{T-stat} $\uparrow$ & \textbf{p-value} $\downarrow$ & \textbf{T-stat} $\uparrow$ & \textbf{p-value} $\downarrow$ \\
\hline \hline
& DeepMIL~\cite{deepmil} $\dagger$ & \textit{icml} & 16.8 & $3.4 \times 10^{-16}$ & 9.3 & $5.1 \times 10^{-10}$ \\
& GradCAM{\textit{++}}~\cite{gradcampp} $\dagger$ & \textit{wacv} & 11.2 & $7.9 \times 10^{-12}$ & 139.1 & $2.1\times 10^{-41}$ \\
& LayerCAM~\cite{layercam} $\dagger$ & \textit{ieee tip} & 6.6 & $3.9 \times 10^{-7}$ & 180.1 & $1.9 \times 10^{-44}$ \\
\hline
\end{tabular}
}
\caption{T-test statistics between baselines (DeepMIL,GradCAM{\textit{++}},LayerCAM) and our method (\pixelcam) for localization performance with VGG16 backbone.}
\label{tab:tstat-results-vgg16}
\end{table}

\setlength{\tabcolsep}{3pt}
\renewcommand{\arraystretch}{1.0}
\begin{table}[ht!]
\centering
\resizebox{0.48\textwidth}{!}{
\small
\begin{tabular}{ | l  l | l | c c | c c |}
\hline
& & & \multicolumn{2}{c|}{\glas} & \multicolumn{2}{c|}{\camsixteen} \\
& \textbf{WSOL models} & \textbf{Venue} & \textbf{T-stat} $\uparrow$ & \textbf{p-value} $\downarrow$ & \textbf{T-stat} $\uparrow$ & \textbf{p-value} $\downarrow$ \\
\hline \hline
& DeepMIL~\cite{deepmil} $\dagger$ & \textit{icml} & 17.8 & $8.2 \times 10^{-17}$ & 18.0 & $6.5 \times 10^{-17}$ \\
& GradCAM{\textit{++}}~\cite{gradcampp} $\dagger$ & \textit{wacv} & 19.3 & $1.0 \times 10^{-17}$ & 10.9 & $1.5\times 10^{-11}$ \\
& LayerCAM~\cite{layercam} $\dagger$ & \textit{ieee tip} & 13.7 & $5.8 \times 10^{-14}$ & 11.0 & $1.1 \times 10^{-11}$ \\
\hline
\end{tabular}
}
\caption{T-test statistics between baselines (DeepMIL,GradCAM{\textit{++}},LayerCAM) and our method (\pixelcam) for localization performance with InceptionV3 backbone.}
\label{tab:tstat-results-inceptionv3}
\end{table}

\section{Computational Complexity}
The computation overhead of the pixel classifier in \pixelcam is minimal in our context. It is suitable for analyzing whole slide images (WSIs) that contain millions of pixels \citep{rony23}. From a memory point of view, the impact is negligible since the number of supplementary parameters depends essentially on feature size. For instance, in the ResNet50-based architecture, the feature size is equal to 2048, and the number of classes is equal to two (FG and BG). The pixel classifier adds only 4,098 parameters to the CNN-based architecture, which already contains more than 23.5M parameters. The same conclusion applies to the transformer-based architecture ($>$ 5.5M parameters), where we add 386 parameters only.

For the inference computational cost, our pixel classifier is particularly advantageous, especially compared to WSOL gradient-based models for WSOL (see Tab.\ref{tab:inference-time-wsol}). Therefore, our model incurs negligible computation overhead, allowing for fast training and inference. The time efficiency, combined with the robustness of PixelCAM, makes it a key advantage for deployment in practical scenarios where WSI image sizes can be extremely large.

\begin{table}[!htbp]
    \centering
    \resizebox{.48\textwidth}{!}{%
    \begin{tabular}{|c|c|c|}
        \hline
       \textbf{WSOL Models} & \textbf{Inference time} \((\downarrow)\) & \textbf{No. parameters} \\
        \hline \hline
       DeepMIL & \textbf{9.1}ms & 24,036,932 \\
       DeepMIL w/ NEGEV & 12.3ms & 33,050,150 \\
       DeepMIL w/ PixelCAM & 9.2ms & 24,041,030 \\
        \hline
        GradCAM++ & 40.9ms & 23,514,179 \\
        GradCAM++ w/ NEGEV & 10.3ms & 32,527,397 \\
        GradCAM++ w/ PixelCAM & \textbf{8.8}ms & 23,518,277 \\
        \hline
        LayerCAM & 37.9ms & 23,514,179 \\
        LayerCAM w/ NEGEV & 10.4ms & 32,527,397 \\
        LayerCAM w/ PixelCAM & \textbf{8.4}ms & 23,518,277 \\
        \hline
        SAT & \textbf{10.5}ms & 5,528,267 \\
        SAT w/ PixelCAM & 11.5ms & 5,528,651 \\
        \hline
        U-Net & \textbf{10.1}ms & 32,521,250 \\
        \hline
    \end{tabular}%
    }
    \caption{Inference time required to produce CAMs using different WSOL methods with a ResNet50 architecture for CNN-based models and DeiT-Tiny for Transformer-based models. The time needed to build a full-size CAM is estimated using an NVIDIA RTX A6000 GPU for one random RGB image of size $224 \times 224$.}
    \label{tab:inference-time-wsol}
\end{table}

\section{Additional Results for Classification (\cl) and Localization (\pxap) Accuracy on \camsixteen Test Set with Different Stainings.}
\label{sec:staining}

In this section, we extend the experiment initially conducted on the \glas dataset to the \camsixteen dataset. We modify the stainings in the \camsixteen test set to evaluate the robustness of \pixelcam compare to the baseline method (LayerCAM). \pixelcam consistently outperforms the baseline in terms of robustness on both classification and localization tasks.

\begin{figure}[!ht]
  \includegraphics[width=\linewidth]{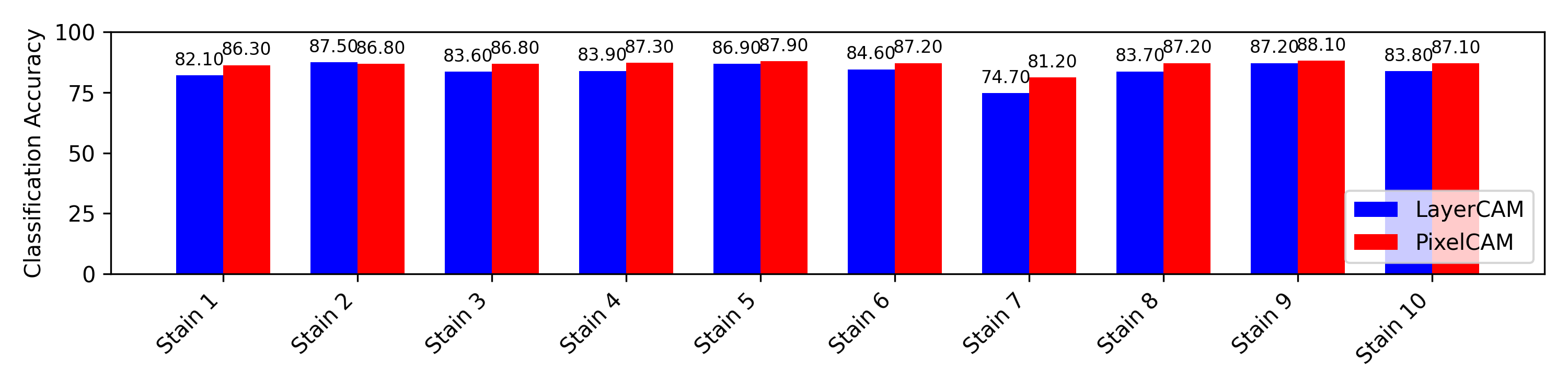}
  \caption{Classification (\cl) accuracy on \camsixteen test sets with LayerCAM and \pixelcam with different stainings.
  }
  \label{fig:visual-example-glas-cl}
\end{figure}

\begin{figure}[!ht]
  \includegraphics[width=\linewidth]{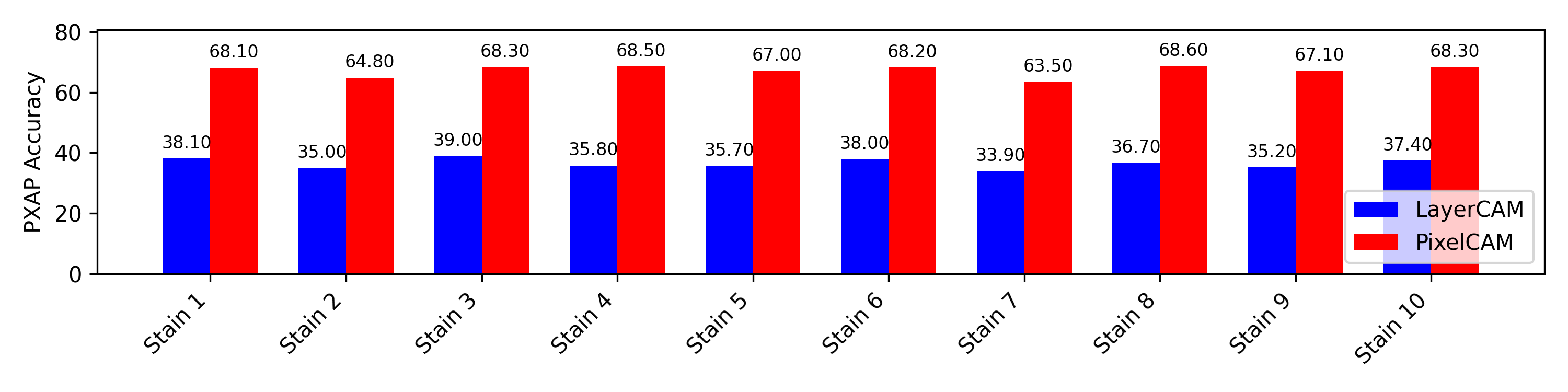}
  \caption{Localization (\pxap) accuracy on \camsixteen test sets with LayerCAM and \pixelcam with different stainings.
  }
  \label{fig:visual-example-glas-loc}
\end{figure}

\section{Limitations and Future Work}

WSOL methods are known to underperform when applied to new datasets due to domain shift \textcolor{red}{\citep{sfdawsol}}. Although our model is impacted by this issue, it demonstrates improved performance in terms of \pxap (localization) and \cl (classification). However, it may still face difficulties in maintaining the same level of performance as on the source dataset, particularly in the presence of extreme shifts (e.g., from \glas to \camsixteen).
To mitigate these limitations, several domain adaptation strategies could be explored. Many existing approaches rely on feature alignment techniques, such as contrastive learning or distribution alignment, to reduce domain shift. Our new pixel classifier can serve exactly as the image classifier by adapting such techniques at the pixel level.  Additionally, in domain adaptation context, most of the techniques use clustering techniques to refine pseudo labels. Since our model produces more discriminative features (as shown in Tables~\ref{tab:accuracy-class-separability-bloc-source} and \ref{tab:accuracy-class-separability-bloc-target}
), \pixelcam can improve clustering effectiveness leading to more reliable pixel pseudo-label adaptation approaches.

\section{Ablations}
\label{sec:ablations}

We provide in this section several ablations of our method \pixelcam.


\subsection{Impact of \pixelcam Depth}
\label{subsec:ablation-depth}

All the reported results of our method are obtained with a linear classifier that classifies pixel-embeddings into  FG/BG classes. In this section, we further explore the impact of using a multi-layer classifier composed of three 1×1 convolutional layers, which act as fully connected layers by reducing the dimensionality of the previous layer's output by a factor of 2.


\setlength{\tabcolsep}{3pt}
\renewcommand{\arraystretch}{1.0}
\begin{table}[!ht]
\centering
\resizebox{0.48\textwidth}{!}{
\small
\begin{tabular}{ | l  l  |l | c c c | c  c c c|}
\hline
& & &  & \multicolumn{2}{c}{\glas}  & & \multicolumn{2}{c}{\camsixteen} &\\
& \textbf{WSOL models} & Venue &  & \pxap \(\uparrow\)& \cl \(\uparrow\)&  & \pxap \(\uparrow\)& \cl \(\uparrow\)& \\
\hline \hline
&  DeepMIL~\cite{deepmil} & \textit{icml} &  & 79.9 &\textbf{100.0}& &71.3 &85.0& \\
& DeepMIL w/ linear \pixelcam & - &  & \underline{85.5} &\textbf{100.0}& &\textbf{75.7} & \underline{88.2}& \\
&  DeepMIL w/ multi-layer \pixelcam & - &  & \textbf{86.3} & 95.0 & &74.4 & \textbf{88.6}& \\
\hline
&  GradCAM{\textit{++}}~\cite{gradcampp} & \textit{wacv} &  & 77.9 &\textbf{100.0}& &49.1 & 63.4& \\
& GradCAM{\textit{++}} w/ linear \pixelcam & - &  & \textbf{86.6} &\textbf{100.0}& &\underline{63.4}&\textbf{88.7}& \\
& GradCAM{\textit{++}} w/ multi-layer \pixelcam & - &  & 86.5 &95.0& &\textbf{64.0}& \underline{85.8}& \\
\hline
\end{tabular}
}
\caption{Localization (\pxap) and classification (\cl) accuracy on \glas and \camsixteen test set with linear and multi-layer \pixelcam for Standard WSOL setup.
}
\label{tab:accuracy-ml-bloc-source}

\end{table}

As shown in Tab.~\ref{tab:accuracy-ml-bloc-source}, adding hidden layers to the pixel classifier is not necessarily an advantage for the localization task. Indeed, models with multiple layers perform worse on \glas and \camsixteen for the WSOL baseline methods GradCAM++~\cite{gradcampp} and DeepMIL~\cite{deepmil}, respectively. In the case of OOD data, Tab.\ref{tab:accuracy-ml-bloc-target} shows that large performance degradation can observed when using multi-layer classifier. Therefore, as a general rule, we recommend using simply a linear pixel-classifier.

\setlength{\tabcolsep}{3pt}
\renewcommand{\arraystretch}{1.0}
\begin{table}[!ht]
\vspace{2em}
\centering
\resizebox{0.48\textwidth}{!}{
\small
\begin{tabular}{ | l  l | l | c c c | c  c c c|}
\hline
& & &  & \multicolumn{2}{c}{\camsixteen $\rightarrow$ \glas}  & & \multicolumn{2}{c}{\glas  $\rightarrow$ \camsixteen} &\\
& \textbf{WSOL models} & \textbf{Venue} &  & \pxap \(\uparrow\)& \cl \(\uparrow\)&  & \pxap \(\uparrow\)& \cl \(\uparrow\)& \\
\hline \hline
&  DeepMIL~\cite{deepmil} & \textit{icml}  &  &64.5 &\underline{81.2}& &29.0 &\textbf{55.2}& \\
&   DeepMIL w/ linear \pixelcam & - &  &\textbf{69.1} &\textbf{83.8}& &\textbf{30.2} &\underline{52.5}& \\
&   DeepMIL w/ multi-layer \pixelcam & - &  &\underline{67.4} &80.0& &25.0 &51.6& \\
\hline
&  GradCAM{\textit{++}}~\cite{gradcampp} & \textit{wacv} &  & 52.9 &53.7& &\textbf{39.1} &\underline{52.4}& \\
&   GradCAM{\textit{++}} w/ linear \pixelcam & - &  & \textbf{56.2} &\textbf{71.2}& &\underline{36.9}& \textbf{63.3}& \\
&   GradCAM{\textit{++}} w/ multi-layer \pixelcam & - &  & \underline{55.8} &\textbf{71.2}& &33.8& 50.6& \\
\hline
\end{tabular}
}
\caption{Localization (\pxap) and classification (\cl) accuracy on \glas and \camsixteen test set with linear and multi-layer \pixelcam for OOD setup.
}
\label{tab:accuracy-ml-bloc-target}
\end{table}



\subsection{Impact of Model Selection of CAM Pre-trained Model} 
\label{subsec:ablation_cams}

Our method leverages pseudo-labels extracted from a pretrained WSOL CAM-based model. However, due to the asynchronous convergence of classification and localization tasks~\cite{rony23}, the criterion used for model selection is expected to have an impact on CAM quality, and therefore, pseudo-labels accuracy. Typically, models are selected based either on their classification performance on validation set by taking the best classifier (\becl), or their localization performance and considering the best localizer (\beloc). In Tab.\ref{tab:accuracy-bclcam-source}, we present the impact of such choice on our method \pixelcam.


\setlength{\tabcolsep}{3pt}
\renewcommand{\arraystretch}{1.0}
\begin{table}[!ht]
\centering
\resizebox{0.48\textwidth}{!}{
\small
\begin{tabular}{ | l  l | l | c c c | c  c c c|}
\hline
& & &  & \multicolumn{2}{c}{\glas}  & & \multicolumn{2}{c}{\camsixteen} &\\
& \textbf{WSOL models} & Venue &  & \pxap \(\uparrow\)& \cl \(\uparrow\)&  & \pxap \(\uparrow\)& \cl \(\uparrow\)& \\
\hline \hline
&  LayerCAM~\cite{layercam} & \textit{ieee tip} &  &75.1 &\textbf{100.0}& &33.2 &84.8& \\
&  LayerCAM w/ \pixelcam  (\beloc CAM) & - &  &\textbf{83.6}&\textbf{100.0}& &\underline{66.2}& \underline{89.1}& \\
&   LayerCAM w/ \pixelcam (\becl CAM) & - &  & \underline{80.6} &98.8& &\textbf{67.5}& \textbf{89.4}& \\
\hline
& SAT~\cite{SAT} & \textit{iccv} &  &65.9 &98.8& &32.8 &83.2& \\
&  SAT w/ \pixelcam (\beloc CAM) & - &  & \textbf{79.1}&\textbf{100.0}& &\textbf{51.2}& \textbf{87.2}& \\
&  SAT w/ \pixelcam (\becl CAM) & - &  & \underline{75.5} &\textbf{100.0}& &\underline{35.6}& \underline{86.4}& \\
\hline
\end{tabular}
}
\caption{Impact of model selection of CAM-based model to build pseudo-labels: Comparison of \pixelcam on \glas and \camsixteen datasets on test set by using CAMs from \beloc and \becl models respectively. The performance is compared to the standard WSOL setup.
}
\label{tab:accuracy-bclcam-source}
\end{table}

Initially, the CAMs used for pixel selection were generated from the \beloc model. We compared the performance compared to using \becl's CAMs. Table~\ref{tab:accuracy-bclcam-source} shows the results of various experiments with different WSOL baseline methods. We noticed that \pixelcam improves localization performance regardless of whether \beloc or \becl is used. However, for techniques that use gradients, such as GradCAM++~\cite{gradcampp} and LayerCAM~\cite{layercam}, CAMs generated with \becl provide better performance. This can be explained by the nature of the CAMs produced by \beloc for these methods. Specifically, \beloc generates CAMs with better localization performance due to higher true positive rates but also introduces more false positives. This negatively impacts training, as incorrect labels are incorporated, potentially reducing overall performance.


We also observe a significant difference for SAT method~\cite{SAT}. This is mainly explained by the fact that the CAMs generated by the \becl model produce a high number of false positives, similar to those generated by \beloc. However, the \beloc model generates a higher true positive rate than \becl, which is crucial for the strong performance of \pixelcam.

\subsection{Impact of Pixel Sampling Technique}
\label{subsec:ablation_sampling}

Our method \pixelcam uses pseudo-annotation for the pixel classifier's training. In particular, we consider random sampling of pixel locations to generate pseudo-labels as it has shown to yield better performance than fitting static regions~\citep{negev}. This avoids overfitting regions and promote exploring potential ROIs.
In this section, we investigate the impact of various sampling techniques to identify pixels associated to FG and BG. We consider two different sampling approaches namely \emph{Threshold-based}~\cite{fcam} and \emph{Probability-based (PB)}~\cite{negev}. The \emph{Threshold-based (TH)} approach automatically thresholds the CAM. Then, it considers all the pixels with activation above the threshold as foreground and valid for FG sampling. This is delineated by a mask. FG pixels are sampled uniformly within that mask, while BG pixels are sampled from outside the mask. 
The \emph{Probability-based} approach samples FG pixels proportionally to the probability values obtained from the CAM using a multinomial distribution. However, the BG pixels are samples from (1 - CAM) activations so regions with low activations will have higher sampling chance.

\setlength{\tabcolsep}{3pt}
\renewcommand{\arraystretch}{1.0}
\begin{table}[!ht]
\centering
\resizebox{0.48\textwidth}{!}{
\small
\begin{tabular}{ | l  l  | l | c c c | c  c c c|}
\hline
& & &  & \multicolumn{2}{c}{\glas}  & & \multicolumn{2}{c}{\camsixteen} &\\
& \textbf{WSOL models} & \textbf{Venue} &  & \pxap \(\uparrow\)& \cl \(\uparrow\)&  & \pxap \(\uparrow\)& \cl \(\uparrow\)& \\
\hline \hline
&  GradCAM{\textit{++}}~\cite{gradcampp} &\textit{wacv} &  & 76.8 &\textbf{100.0}& &49.1& 63.4& \\
& GradCAM{\textit{++}} w/ TH \pixelcam & - &  & 85.4 & \textbf{100.0} & &\underline{54.2}& \textbf{89.9}& \\
& GradCAM{\textit{++}} w/ PB \pixelcam & - &  & \textbf{86.6} &\textbf{100.0}& &\textbf{64.1}& \underline{85.1}& \\
\hline
&  SAT~\cite{SAT} & \textit{iccv} &  & 65.9 &\underline{98.8}& &32.8 &\underline{83.2}& \\
& SAT w/ TH \pixelcam & - &  & \underline{71.5} &92.5& &\underline{50.9} & 81.0& \\
&  SAT w/ PB \pixelcam & - &  & \textbf{79.1} &\textbf{100.0} & &\textbf{51.2} & \textbf{87.2}& \\
\hline
\end{tabular}
}
\caption{Impact of pixel sampling technique on \pixelcam performance: Threshold-based (TH) vs. probability-based (PB), over standard WSOL setup. 
}
\label{tab:sampling_exps}
\end{table}

In Tab.~\ref{tab:sampling_exps}, we observe that pixel sampling technique has a significant impact on the localization performance of \pixelcam. For \glas dataset, using GradCAM++ CAMs has a small impact, but for CAMs generated from a transformer architecture as SAT the impact is significant with a difference of 7.6\%. Similarly, on a more challenging dataset as \camsixteen, using a probabilistic approach for sampling pixels can lead to an improvement of 9.9\%. This can be explained by the fact that sampling without relying on a threshold favors relevant pixels and most likely to have the correct pseudo-label. On the other hand, threshold-based method fixes a region for sampling with high likelihood of covering wrong regions. Learning with incorrect labels leads to poor models, and, therefore poor performance.


\subsection{Impact of the Number of Sampled Pixels  \texorpdfstring{$n$}{n}}
\label{subsec:ablation_nb_pixel}

We analyze the impact of the number of pixels selected as pseudo-labels to train \pixelcam. To avoid unbalanced pixel classification, we sample the same number $n$ of pixels as FG and BG. We consider the following cases ${n \in \{1, 5, 10, 20\}}$.
As observed in Tab.~\ref{tab:accuracy-pixels}, increasing $n$ can affect both localization and classification with different degree, and depending on the dataset. In terms of localization, both datasets can slightly be affected. However, \camsixteen is largely affected in term of classification as performance can vary between ${85.7\%}$ and ${90.1\%}$.


\setlength{\tabcolsep}{3pt}
\renewcommand{\arraystretch}{1.0}
\begin{table}[!ht]
\centering 
\resizebox{0.48\textwidth}{!}{
\small
\begin{tabular}{ | l  l |l | c c c | c  c c c|}
\hline
& & &  & \multicolumn{2}{c}{\glas}  & & \multicolumn{2}{c}{\camsixteen} &\\
& \textbf{WSOL models} & Venue &  & \pxap \(\uparrow\)& \cl \(\uparrow\)&  & \pxap \(\uparrow\)& \cl \(\uparrow\)& \\
\hline \hline
& GradCAM{\textit{++}}~\cite{gradcampp} & \textit{wacv} &  & 76.8 &\textbf{100.0}& &49.1& 63.4& \\
& GradCAM{\textit{++}} w/ \pixelcam: & - &  &   & & & & & \\
& $n=1$  &   &  & \textbf{86.7} &\textbf{100.0}& &63.5&\textbf{90.1}& \\
& $n=5$  &   &  & \underline{86.6} &\textbf{100.0}& &\underline{64.1}&85.1& \\
& $n=10$ &   &  & 86.0 &\textbf{100.0}& &63.2&\underline{88.8}& \\
& $n=20$ &   &  & 86.0 &\textbf{100.0}& &\textbf{64.7}&88.7& \\
\hline
\end{tabular}
}
\caption{Impact of the number of sampled pixel as pseudo-labels $n$ on \pixelcam performance. 
We measure the \pxap and \cl performance over the test set for \glas and \camsixteen for standard WSOL setup.
}
\label{tab:accuracy-pixels}
\end{table}



\subsection{Impact of \texorpdfstring{$\lambda$}{lambda}}
\label{subsec:ablation_lambda}

Table~\ref{tab:accuracy-lambda-bloc-source} shows the impact of $\lambda$ on the performance of our method \pixelcam. It achieves better results when using a higher value of $\lambda$ on \glas. As $\lambda$ decreases, the \pxap performance also declines. On the \camsixteen dataset, we note that using a low $\lambda$ value (0.001) significantly impacts localization performance. Therefore, we recommend using $\lambda$ values in the range of 0.1 to 1.0 for better performance.

\setlength{\tabcolsep}{3pt}
\renewcommand{\arraystretch}{1.0}
\begin{table}[!ht]
\centering 
\resizebox{0.48\textwidth}{!}{
\small
\begin{tabular}{ | l  l | c c c | c  c c c|}
\hline
& &  & \multicolumn{2}{c}{\glas}  & & \multicolumn{2}{c}{\camsixteen} &\\
 & \textbf{$\lambda$} &  & \pxap \(\uparrow\)& \cl \(\uparrow\)&  & \pxap \(\uparrow\)& \cl \(\uparrow\)& \\
\hline \hline

&  1 &  & \textbf{83.6}&\textbf{100.0}& &66.2&89.1& \\
& 0.5 &  & \underline{83.1}&\textbf{100.0}& &66.6&\textbf{89.7}& \\
& 0.1  &  & 82.2&\textbf{100.0}& &\textbf{67.4}&88.3& \\
& 0.01  &  & 79.3&96.3& &\underline{66.7}&89.1& \\
& 0.001  &  & 76.1&98.8& &61.6&\underline{89.2}& \\
\hline
\end{tabular}
}
\caption{Impact of hyper-parameter $\lambda$ over \pixelcam in terms of \pxap and \cl performance over test set. \pixelcam uses LayerCAM~\cite{layercam} for pseudo-labels.
}
\label{tab:accuracy-lambda-bloc-source}
\end{table}

\subsection{Impact of using a pretrained WSOL model on same and external dataset for pseudo-labeling}

As mentionned, PixelCAM use a WSOL CAM-based model to obtain pseudo label for foreground and background. In our paper, we consider training a standard classic WSOL CAM-based (DeepMIL, GradCAM++, LayerCAM, SAT) trained on the same dataset to obtain a robust model for the pseudo labeling to avoid the number of false positive. Considering a WSOL CAM-based model trained on an external dataset can considered to avoid computation time during training as it doesn't require a pre step training but tends to perform poorly on a unseen dataset which will cause a high number of false positive and negative leading to a poor pseudo labeling. To support this claim we trained LayerCAM on \glas to obtain pseudo label on \camsixteen and also trained the model on \camsixteen to obtain pseudo labels on \glas. 

\setlength{\tabcolsep}{3pt}
\renewcommand{\arraystretch}{1.0}
\begin{table}[htbp]
\vspace{-4pt}
\centering
\resizebox{0.48\textwidth}{!}{%
\begin{tabular}{|l|cc|cc|cc|cc|}
\hline
& \multicolumn{4}{c|}{\glas} & \multicolumn{4}{c|}{\camsixteen} \\
& \multicolumn{2}{c|}{Same} & \multicolumn{2}{c|}{External} & \multicolumn{2}{c|}{Same} & \multicolumn{2}{c|}{External} \\
& \pxap \(\uparrow\) & \cl \(\uparrow\) & \pxap \(\uparrow\) & \cl \(\uparrow\) & \pxap \(\uparrow\) & \cl \(\uparrow\) & \pxap \(\uparrow\) & \cl \(\uparrow\) \\
\hline \hline
LayerCAM & 75.1 & \textbf{100.0} & 58.1 & 83.8 & 33.2 &84.8& 22.7 & 50.4\\
LayerCAM w/ PixelCAM & \textbf{83.6} & \textbf{100.0} & \textbf{68.1} & \textbf{92.5} & \textbf{66.2} & \textbf{89.1} & \textbf{59.2} & \textbf{81.7} \\
\hline
\end{tabular}
}
\caption{Localization (\pxap) and classification (\cl) accuracy on \glas and \camsixteen test sets with VGG16 and ResNet50 backbones.}
\label{tab:ablation-external-ds}
\vspace{-1em}
\end{table}

As we can observe (Tab.\ref{tab:ablation-external-ds}), the performance of using a WSOL CAM-based method on a external avoid computation cost on the training part but significantly impact the training of PixelCAM and make it less suitable compare to a standard WSOL CAM-based method on GlaS dataset.



\section{More Visualization}
\label{sec:appendix_visualization}

\subsection{Visualization Results on Standard WSOL setup}
\begin{figure*}[!ht]
\centering
  \includegraphics[width=0.85\linewidth]{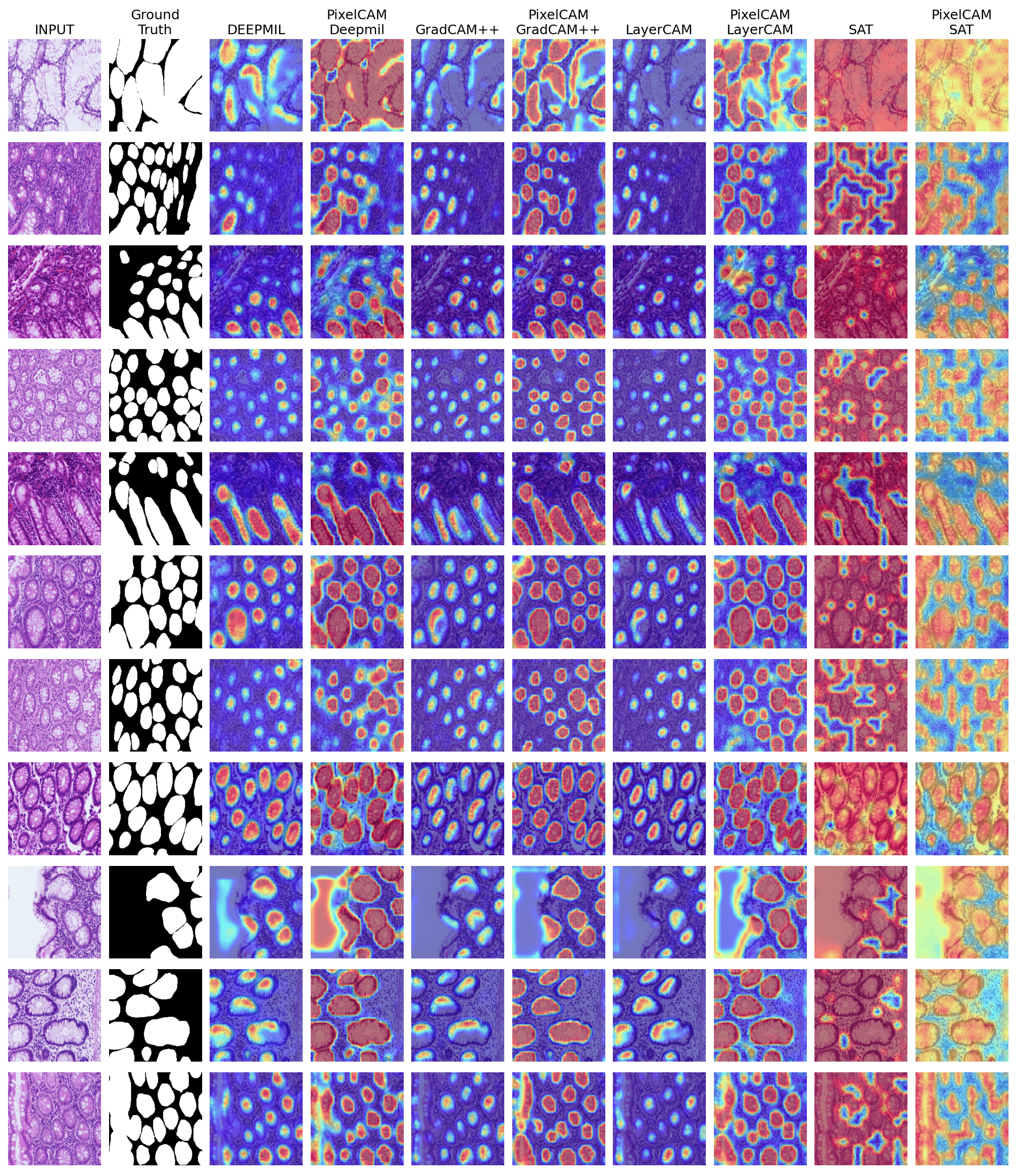}
  \caption{Standard WSOL setup. First column: \textbf{Normal} images from \glas. Second column: Ground truth. Next columns: We display the visual CAM results for WSOL baseline without and with \pixelcam, respectively.
  }
  \label{fig:example-normal-glas-wsol}
\end{figure*}

\begin{figure*}[!ht]
\centering
  \includegraphics[width=0.92\linewidth]{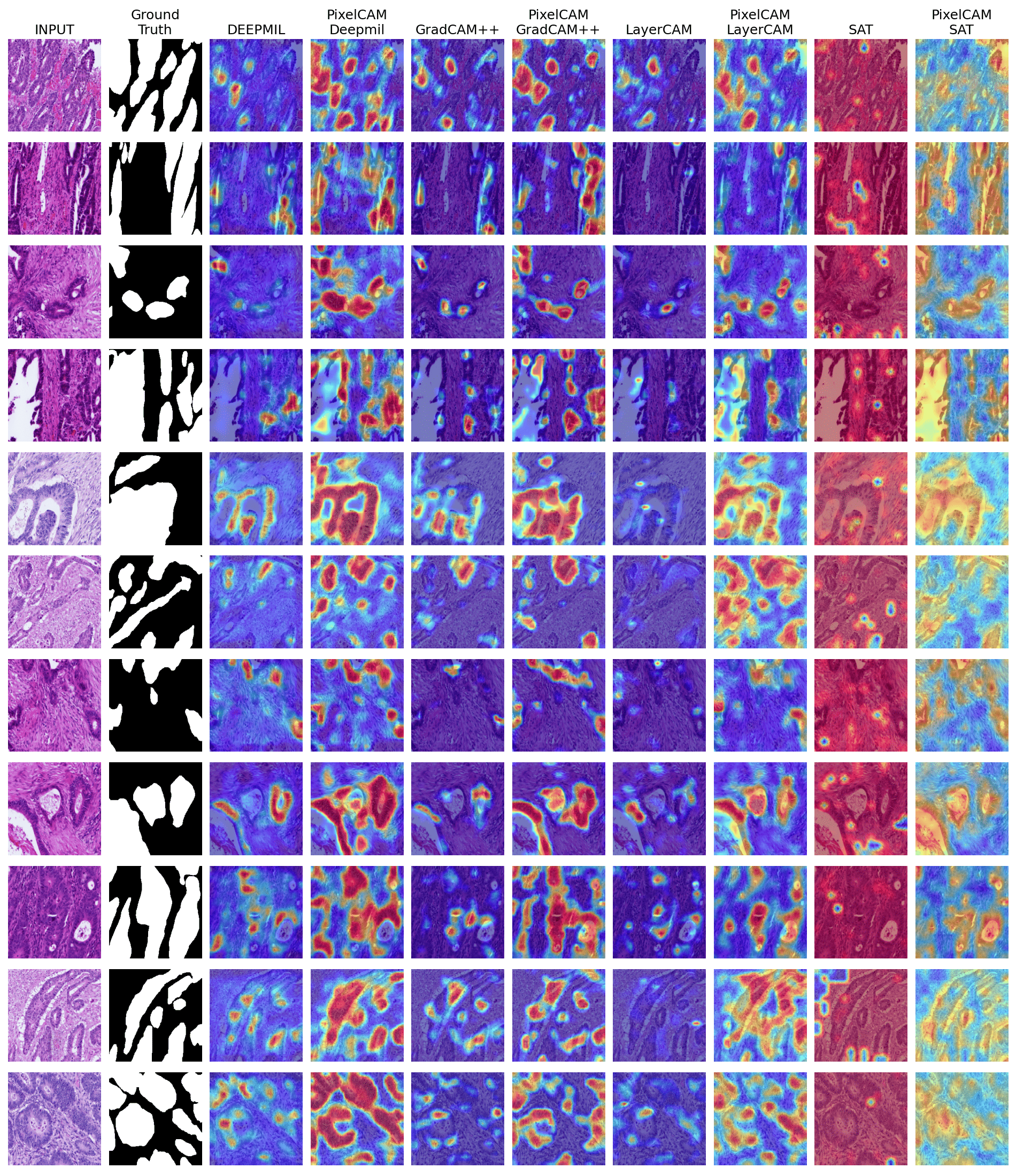}
  \caption{Standard WSOL setup. First column: \textbf{Cancerous} images from \glas. Second column: Ground truth. Next columns: We display the visual CAM results for WSOL baseline without and with \pixelcam, respectively.
  }
  \label{fig:example-cancer-glas-wsol}
\end{figure*}

\begin{figure*}[!ht]
\centering
  \includegraphics[width=0.999\linewidth]{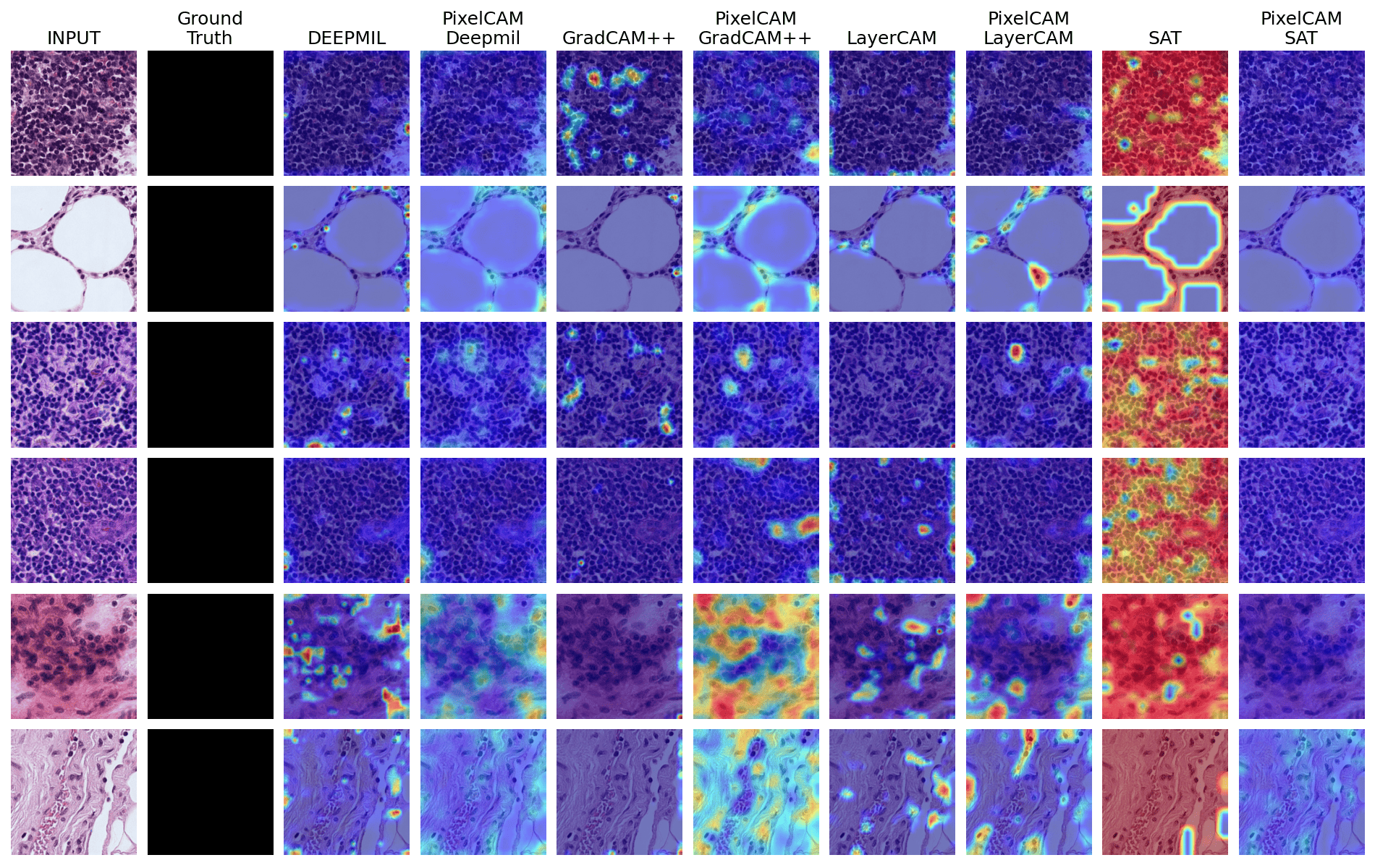}
  \caption{Standard WSOL setup. First column: \textbf{Normal} images from \camsixteen. Second column: Ground truth. Next columns: We display the visual CAM results for WSOL baseline without and with \pixelcam, respectively.
  }
  \label{fig:example-normal-cam16-wsol}
\end{figure*}

In this section, we present visual results of \pixelcam compared to WSOL baseline methods. In a standard WSOL setup, we observe that \pixelcam enhances the ROIs detected by WSOL baseline methods on the \glas dataset (Fig:~\ref{fig:example-normal-glas-wsol} and \ref{fig:example-cancer-glas-wsol}). Regarding \camsixteen, a challenging dataset, \pixelcam improves localization for cancer images by extending ROIs and reducing incorrect predictions from WSOL baseline methods as observed in (Fig:~\ref{fig:example-normal-cam16-wsol}). 

\begin{figure*}[!ht]
  \centering
  \includegraphics[width=0.999\linewidth]{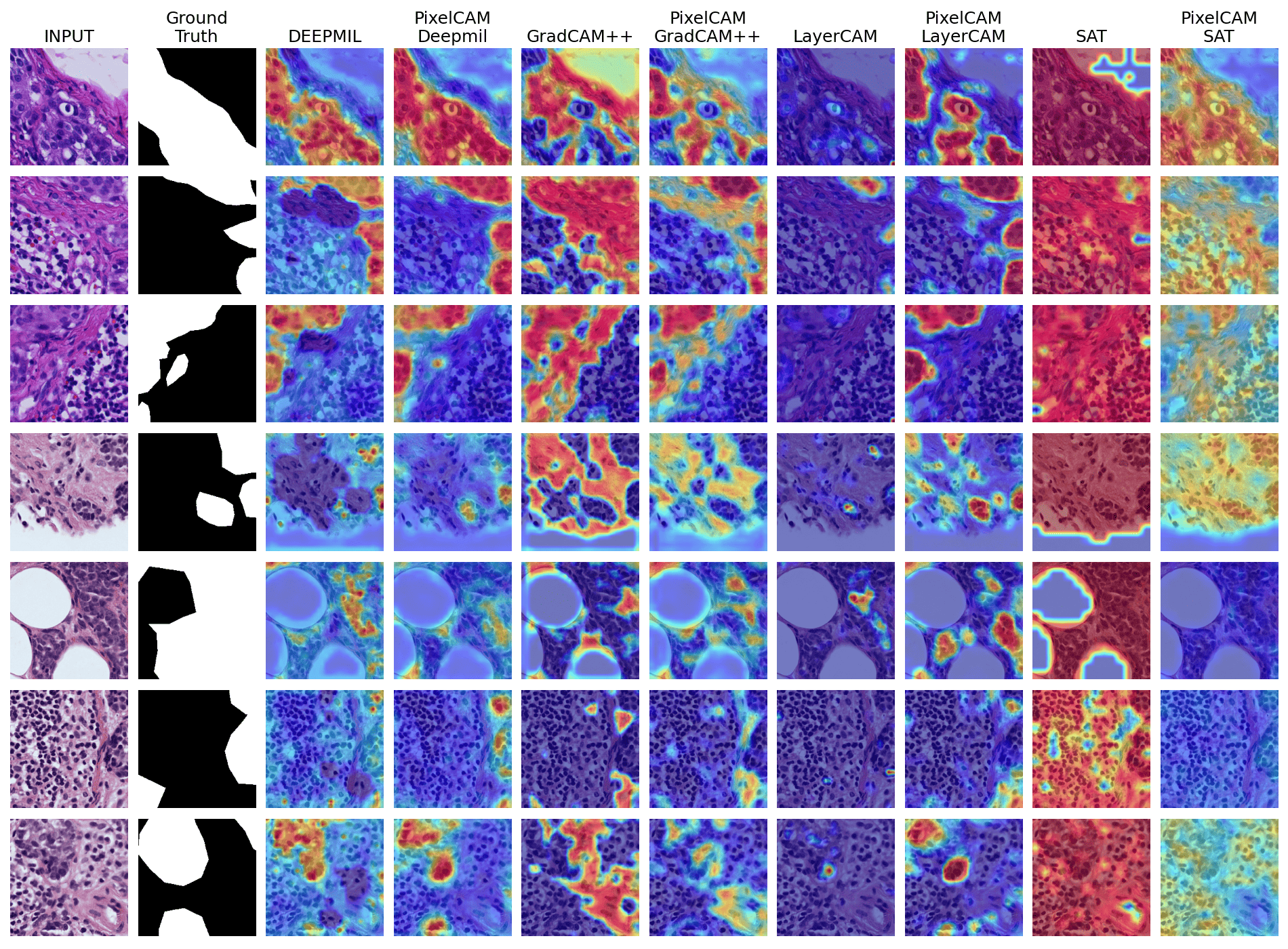}
  \caption{Standard WSOL setup. First column: \textbf{Cancerous} images from \camsixteen. Second column: Ground truth. Next columns: We display the visual CAM results for WSOL baseline without and with \pixelcam, respectively.
  }
  \label{fig:example-cancer-cam16-wsol}
\end{figure*}

\subsection{Visualization Results on Target set with OOD setup}
\begin{figure*}[!ht]
\centering
\includegraphics[width=0.9\linewidth]{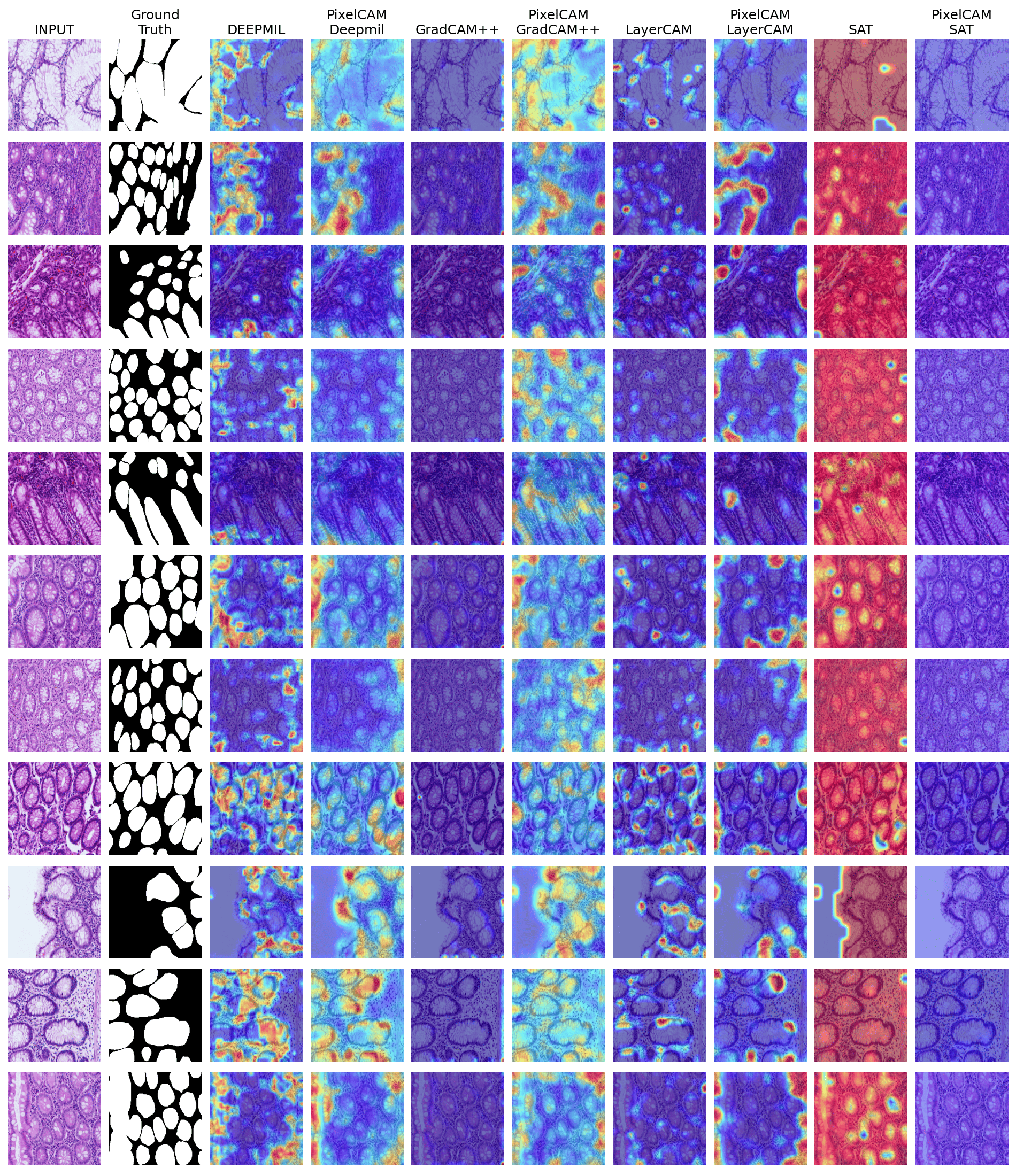}
  \caption{OOD setup:  \mbox{\camsixteen $\rightarrow$ \glas}. First column: \textbf{Normal} images from \glas. Second column: Ground truth. Next columns: We display the visual CAM results for WSOL baseline without and with \pixelcam, respectively.
  }
  \label{fig:example-normal-glas-ood}
\end{figure*}

\begin{figure*}[!ht]
\centering
  \includegraphics[width=0.9\linewidth]{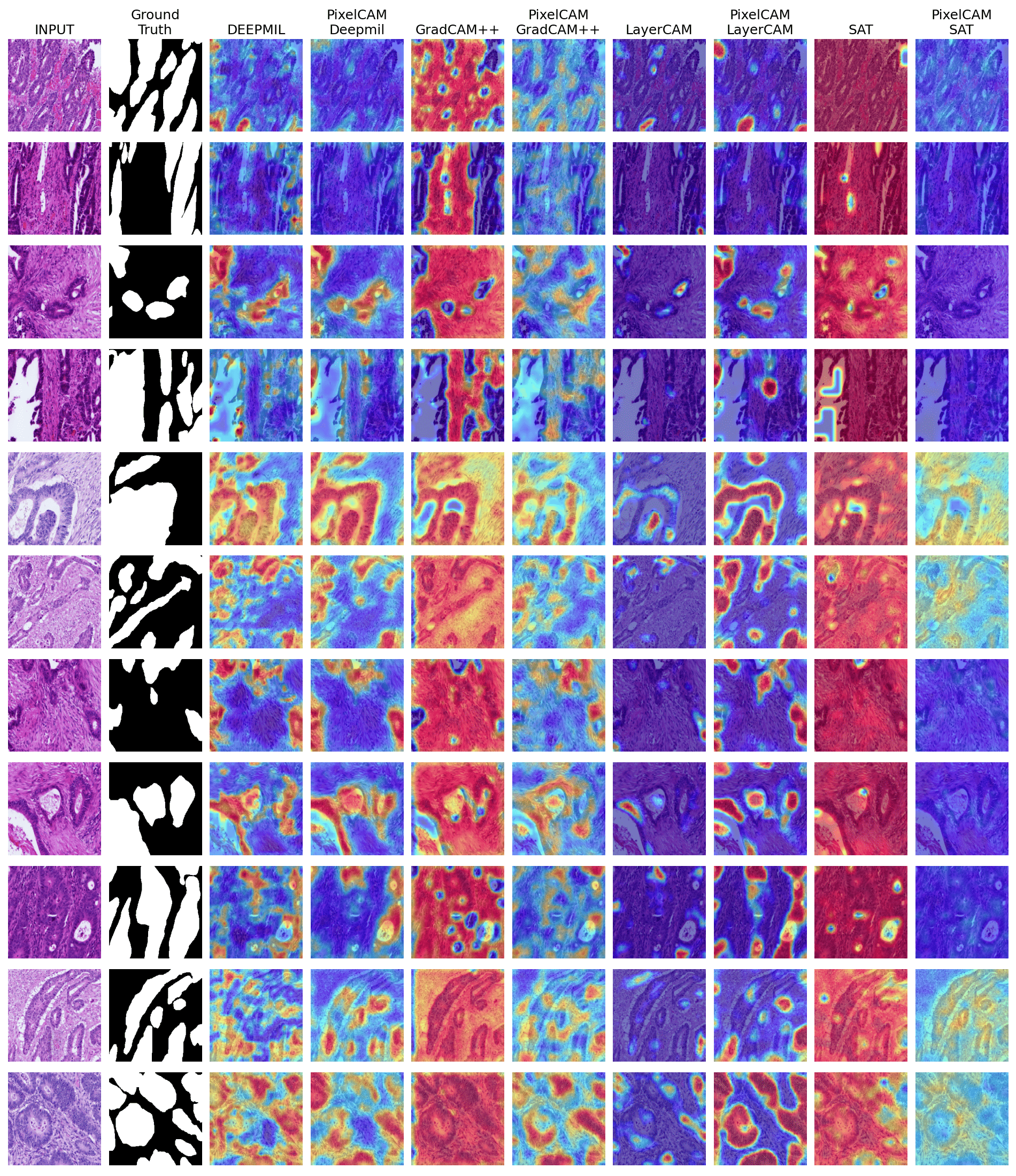}
  \caption{OOD setup:  \mbox{\glas $\rightarrow$ \camsixteen}. First column: \textbf{Cancerous} images from \glas. Second column: Ground truth. Next columns: We display the visual CAM results for WSOL baseline without and with \pixelcam, respectively.
  }  
  \label{fig:example-cancer-glas-ood}
\end{figure*}

\begin{figure*}[!ht]
\centering
  \includegraphics[width=0.9\linewidth]{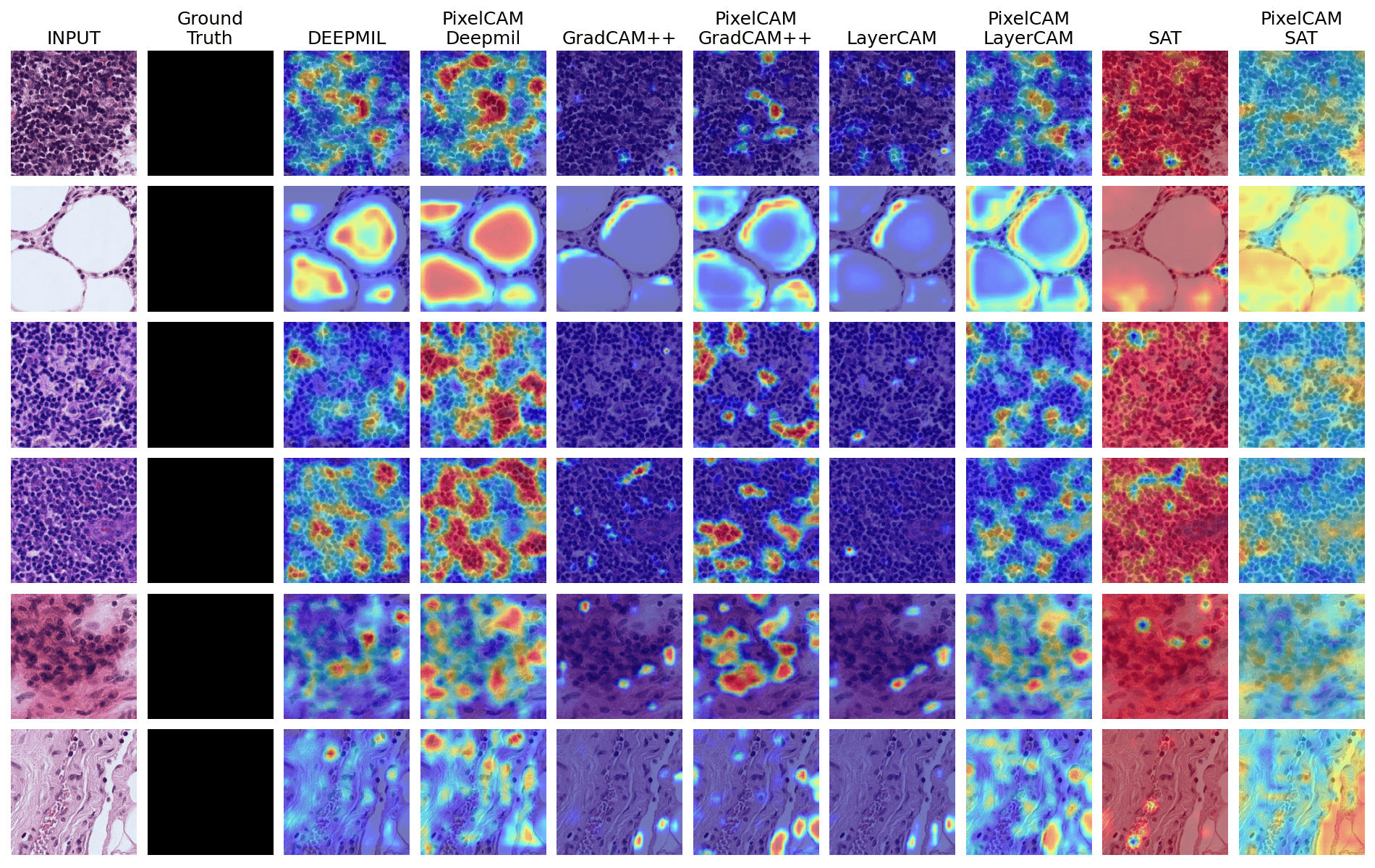}
  \caption{OOD setup:  \mbox{\glas $\rightarrow$ \camsixteen}. First column: \textbf{Normal} images from \camsixteen. Second column: Ground truth. Next columns: We display the visual CAM results for WSOL baseline without and with \pixelcam, respectively.
  }  
  \label{fig:example-normal-cam16-ood}
\end{figure*}

\begin{figure*}[!ht]
\centering
  \includegraphics[width=0.99\linewidth]{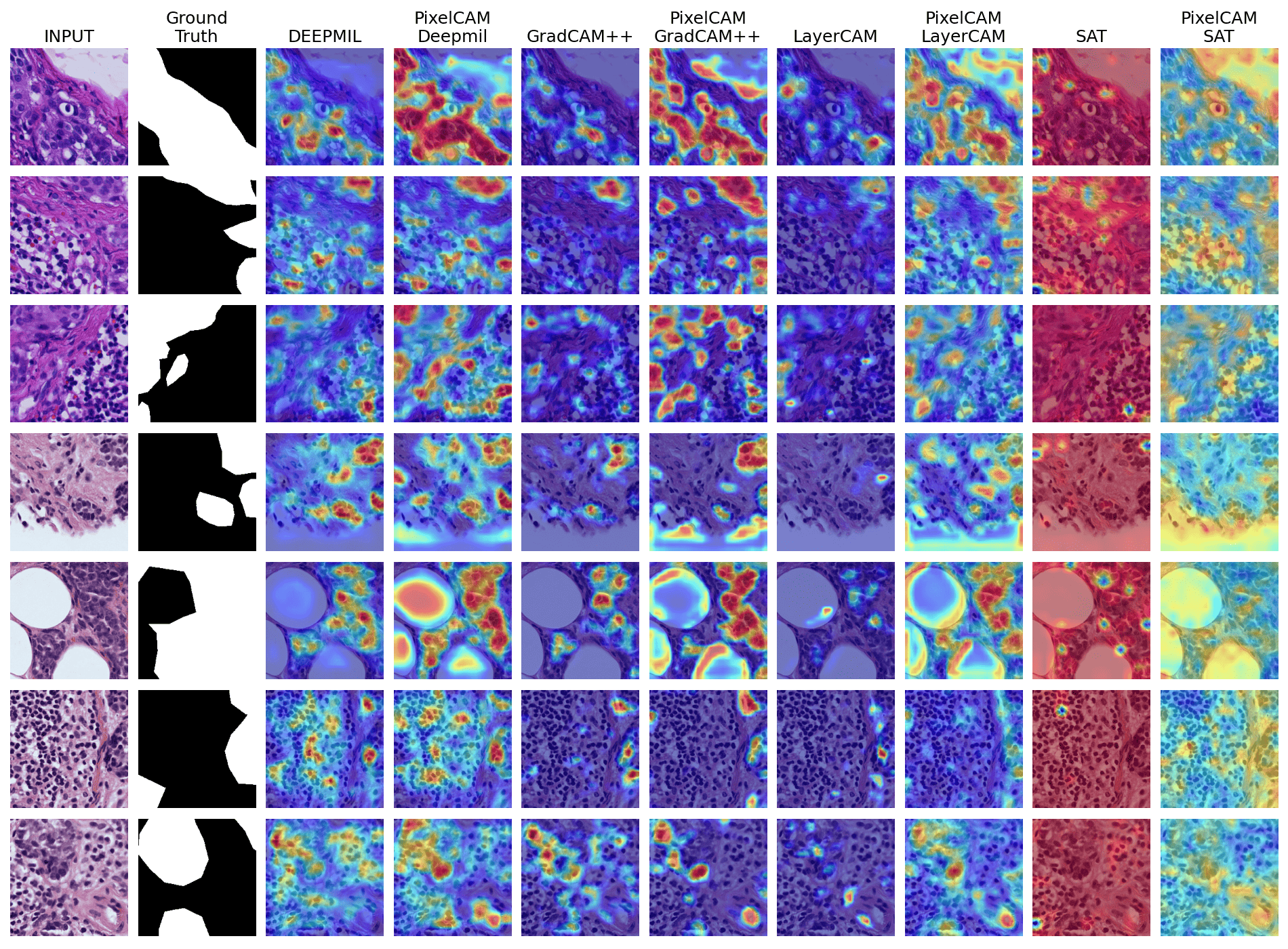}
  \caption{OOD setup:  \mbox{\glas $\rightarrow$ \camsixteen}. First column: \textbf{Cancerous} images from \camsixteen. Second column: Ground truth. Next columns: We display the visual CAM results for WSOL baseline without and with \pixelcam, respectively.
  }  
  \label{fig:example-cancer-cam16-ood}
\end{figure*}

We provide visual results on target set in the case of OOD for both scenarios: \mbox{\camsixteen $\rightarrow$ \glas} and \mbox{\glas $\rightarrow$ \camsixteen}. As we observe, \pixelcam can predict correctly in average cancer ROIs as illustrated in Fig:~\ref{fig:example-cancer-glas-ood} and \ref{fig:example-cancer-cam16-ood} but struggle with normal images as the WSOL baseline methods (Fig:~\ref{fig:example-normal-glas-ood} and \ref{fig:example-normal-cam16-ood}).  

\begin{figure*}[!ht]
\centering
  \begin{tabular}{@{}c@{\hskip 10pt}c@{\hskip 10pt}c@{\hskip 10pt}c@{\hskip 10pt}c@{\hskip 10pt}}
    \makecell{Input} & \makecell{DeepMil} & \makecell{GradCAM++} & \makecell{LayerCAM} & \makecell{SAT} \\
    \includegraphics[width=.12\textwidth]{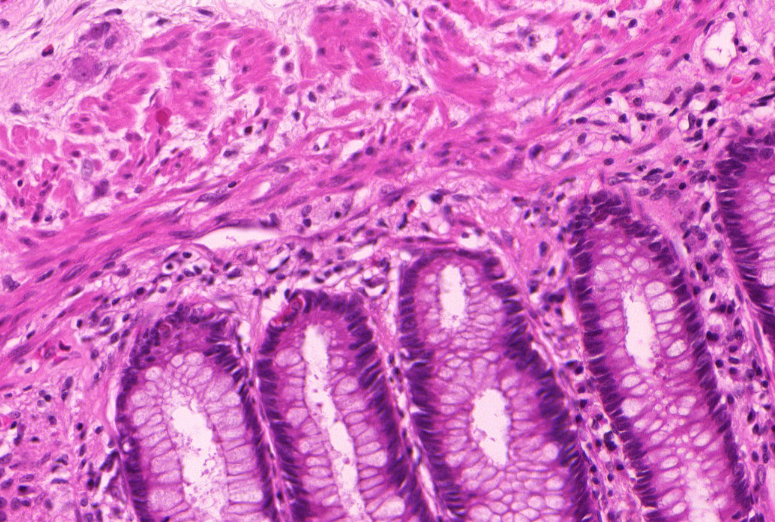} &
    \includegraphics[width=.12\textwidth]{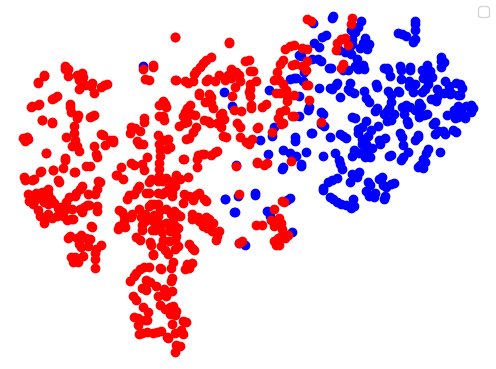} &
    \includegraphics[width=.12\textwidth]{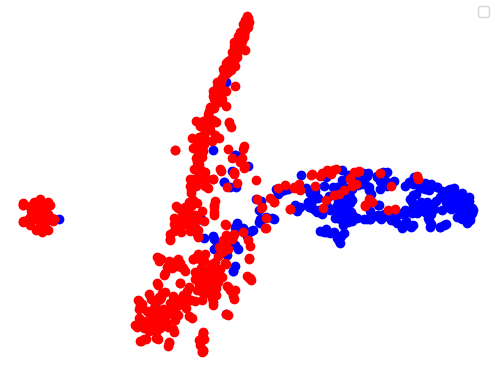} &
    \includegraphics[width=.12\textwidth]{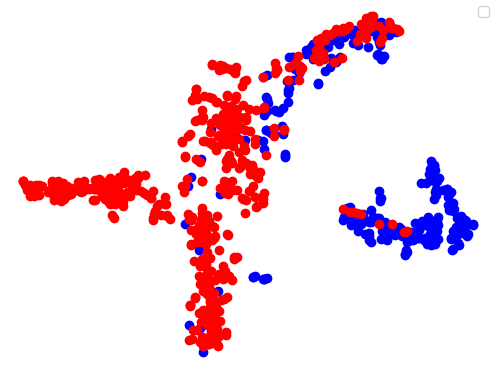} &
    \includegraphics[width=.12\textwidth]{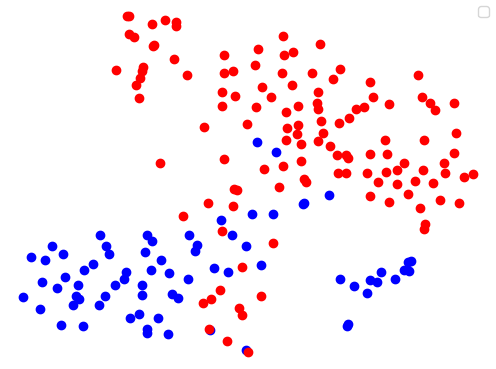} 
    \\
    &
    \includegraphics[width=.12\textwidth]{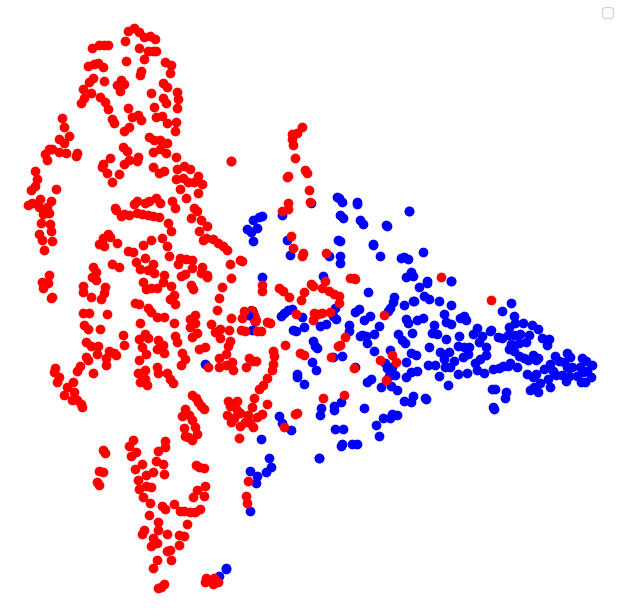} &
    \includegraphics[width=.12\textwidth]{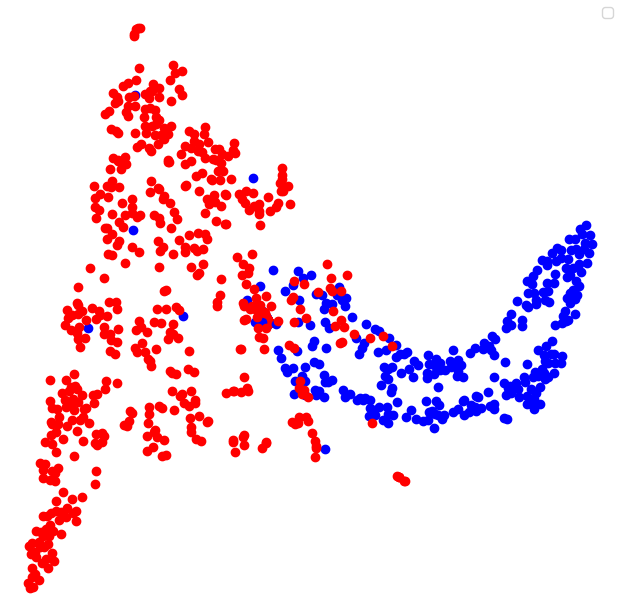} &
    \includegraphics[width=.12\textwidth]{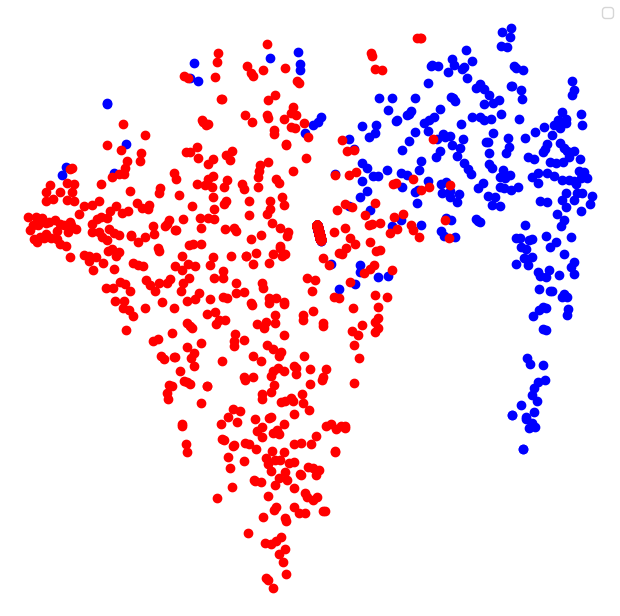} &
    \includegraphics[width=.12\textwidth]{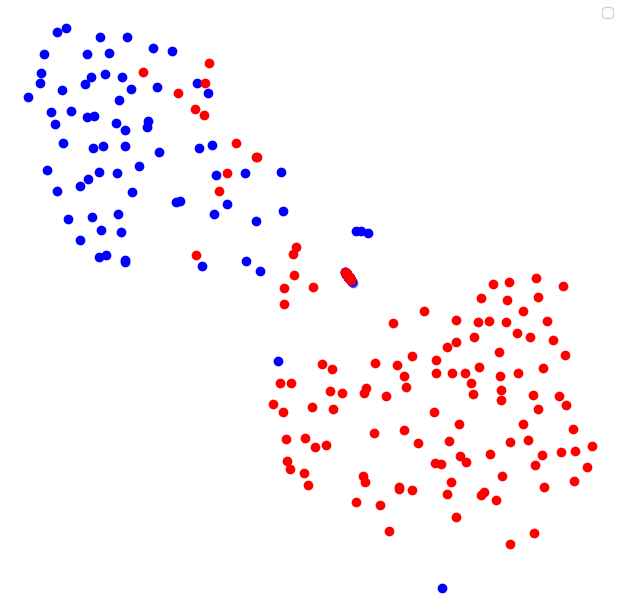} \\
     &
    \includegraphics[width=.12\textwidth]{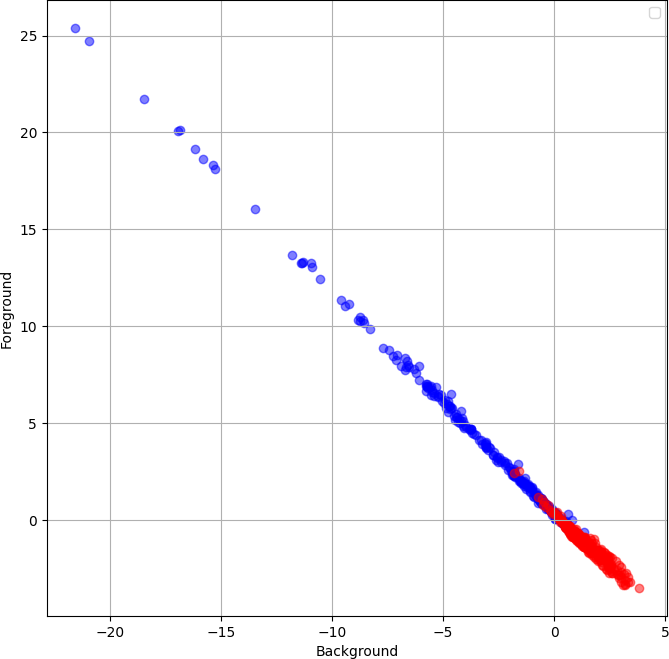} &
    \includegraphics[width=.12\textwidth]{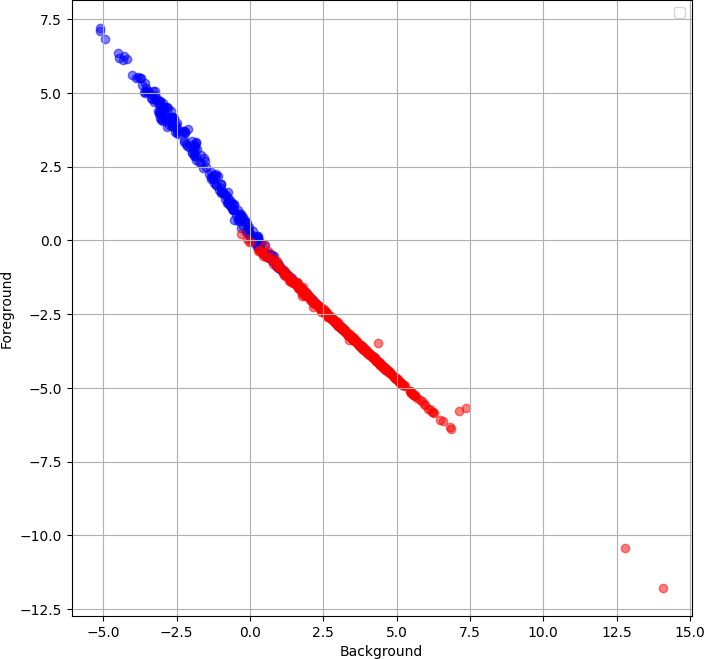} &
    \includegraphics[width=.12\textwidth]{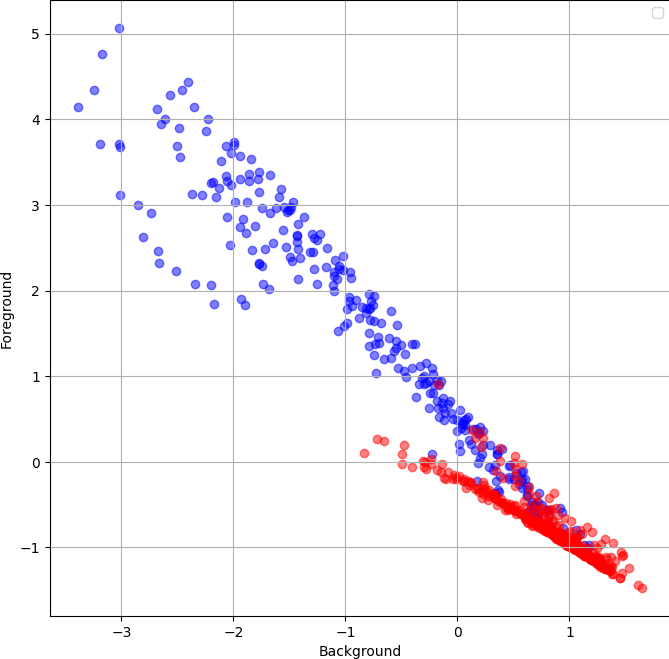} &
    \includegraphics[width=.12\textwidth]{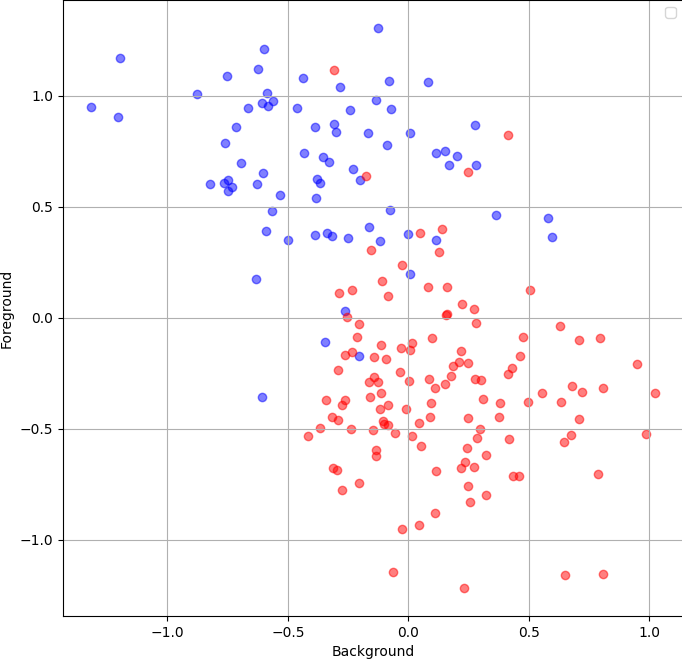} \\    
     \includegraphics[width=.12\textwidth]{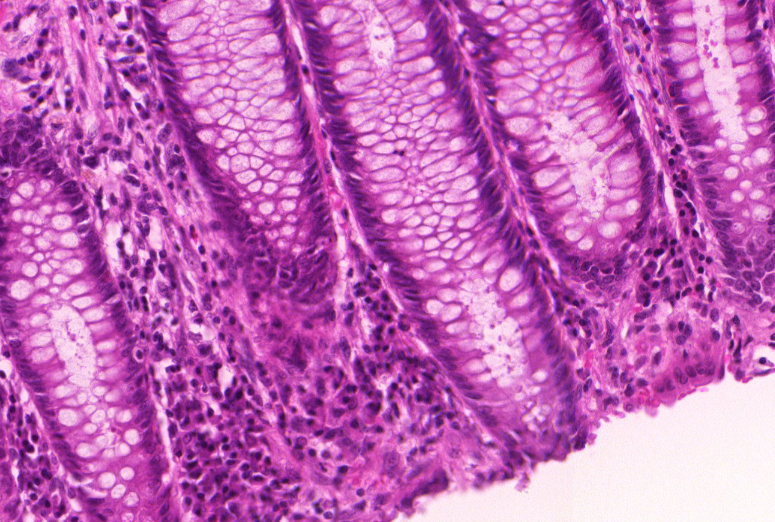} &
    \includegraphics[width=.12\textwidth]{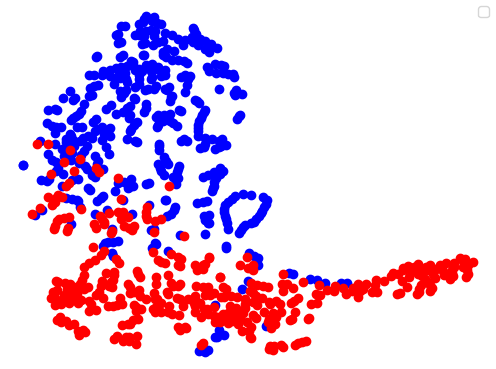} &
    \includegraphics[width=.12\textwidth]{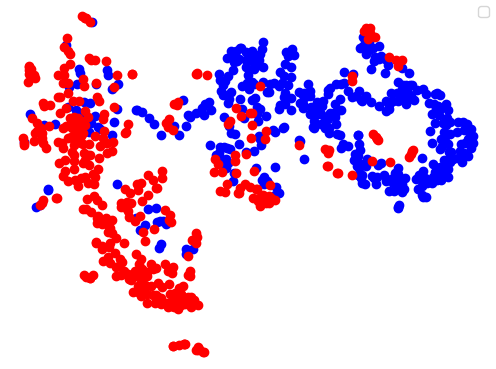} &
    \includegraphics[width=.12\textwidth]{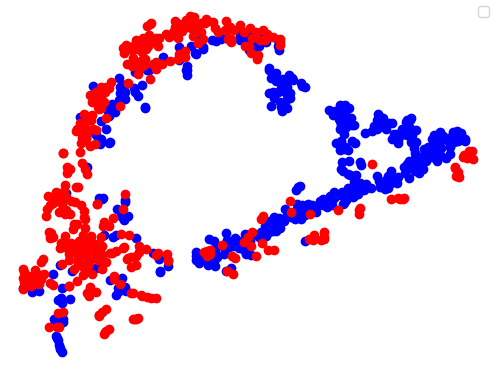} &
    \includegraphics[width=.12\textwidth]{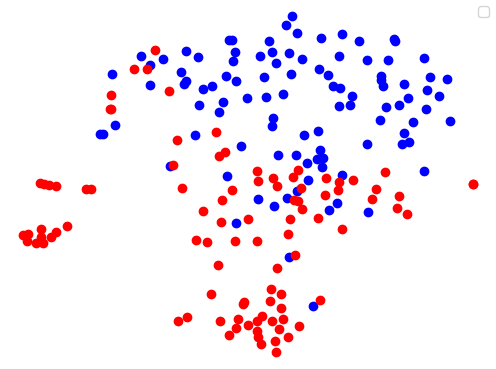} \\
     &
    \includegraphics[width=.12\textwidth]{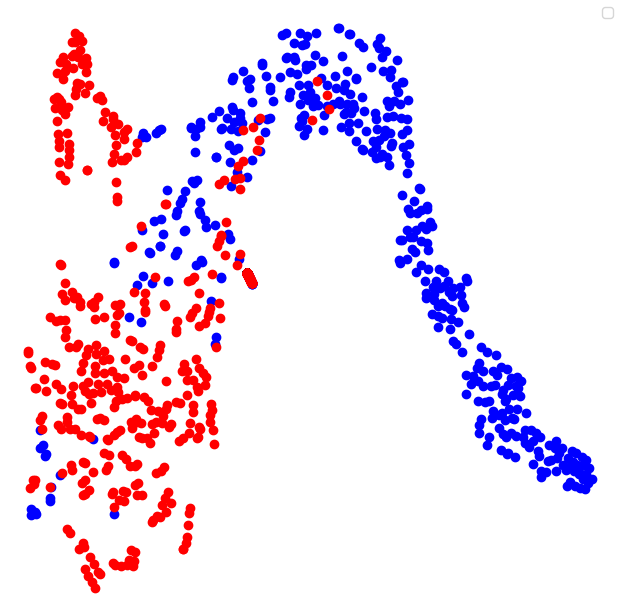} &
    \includegraphics[width=.12\textwidth]{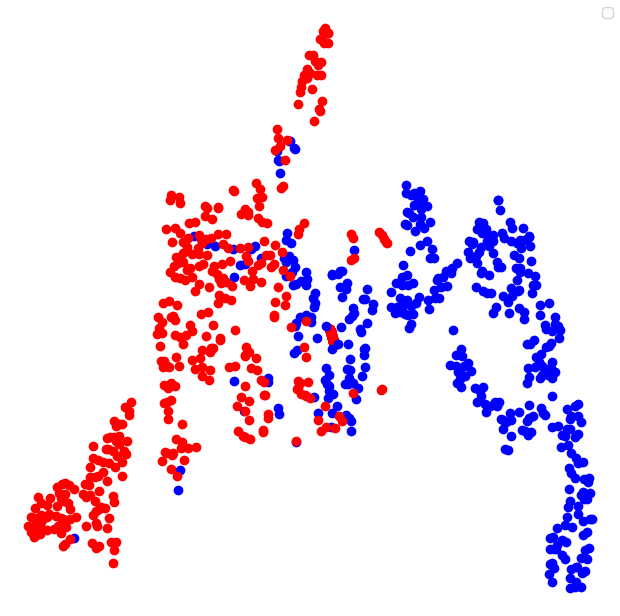} &
    \includegraphics[width=.12\textwidth]{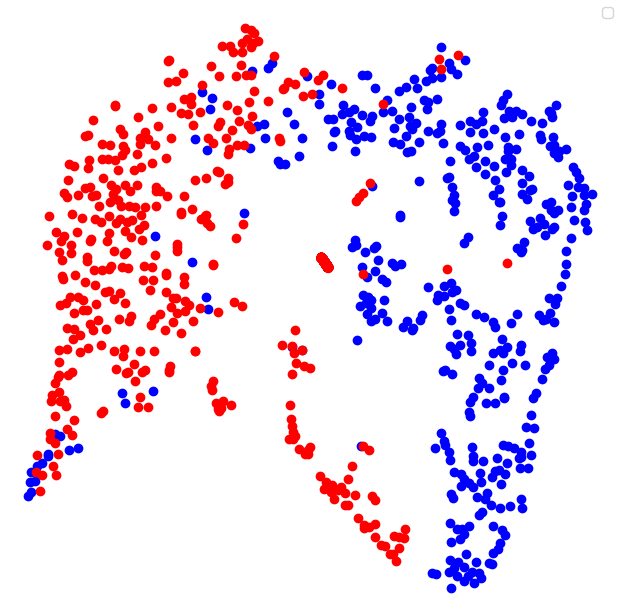} &
    \includegraphics[width=.12\textwidth]{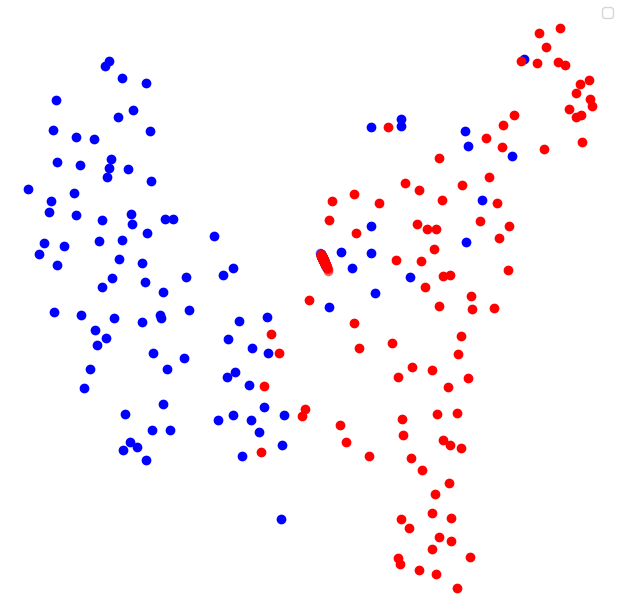} \\
     &
    \includegraphics[width=.12\textwidth]{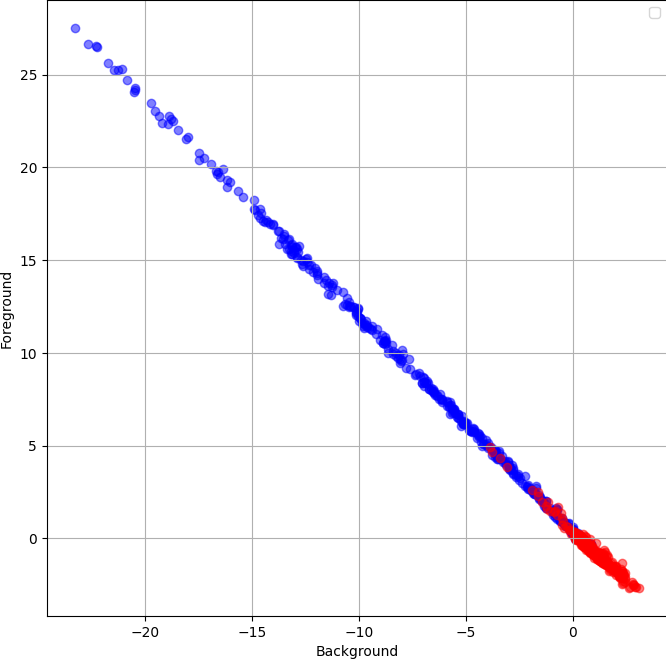} &
    \includegraphics[width=.12\textwidth]{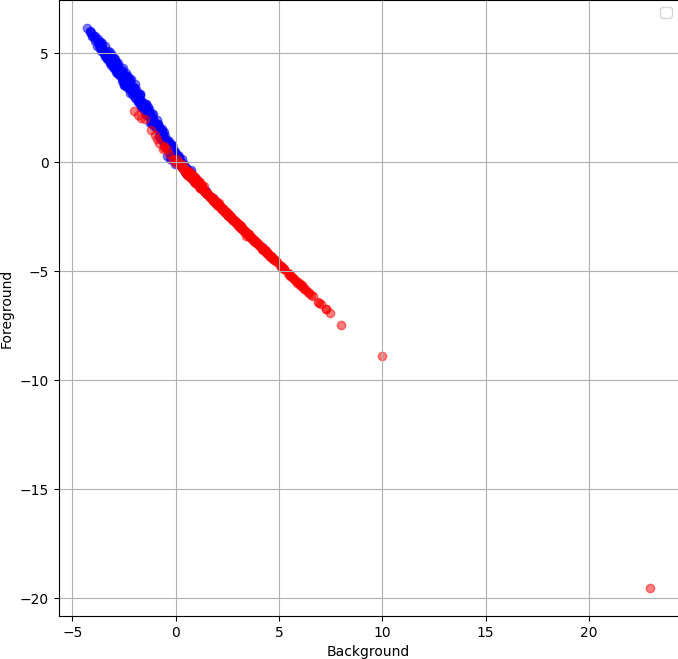} &
    \includegraphics[width=.12\textwidth]{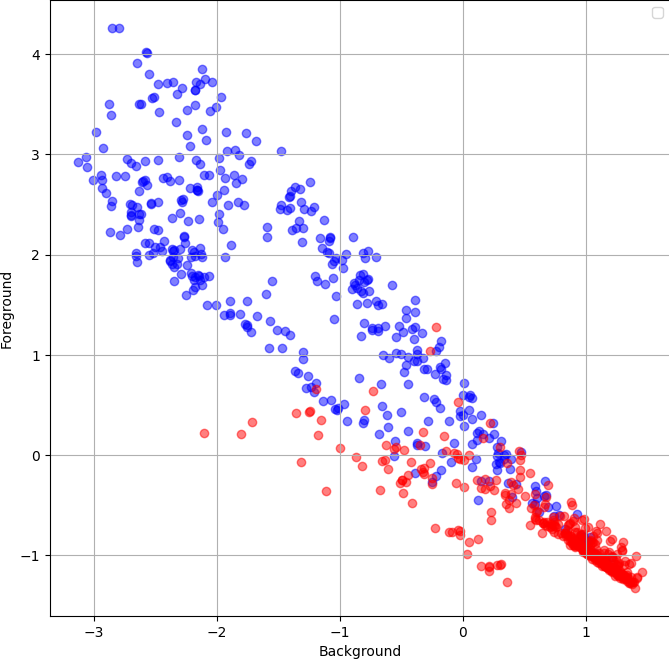} &
    \includegraphics[width=.12\textwidth]{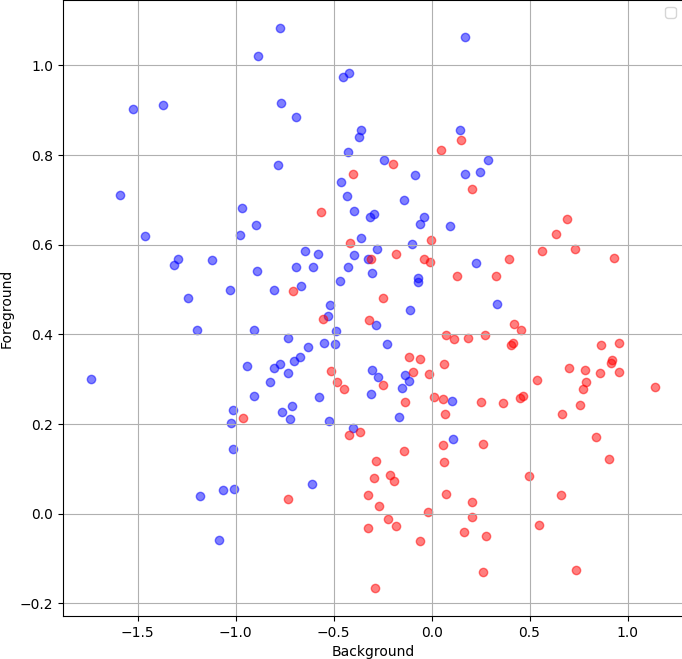} \\
    \includegraphics[width=.12\textwidth]{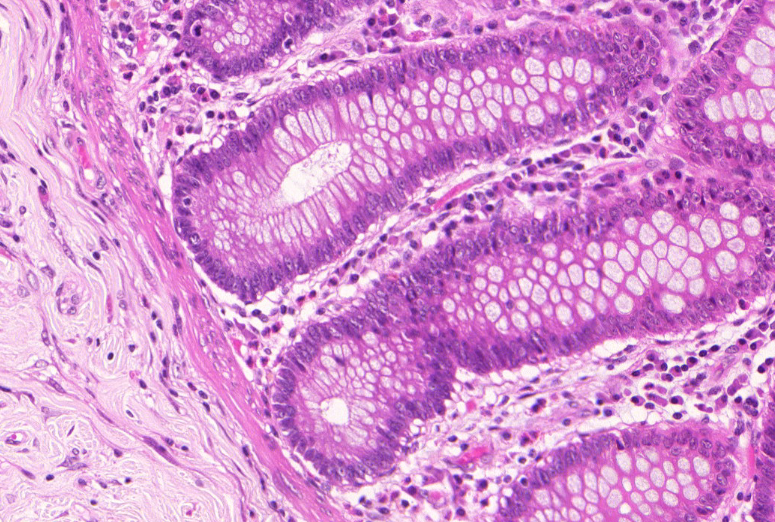} &
    \includegraphics[width=.12\textwidth]{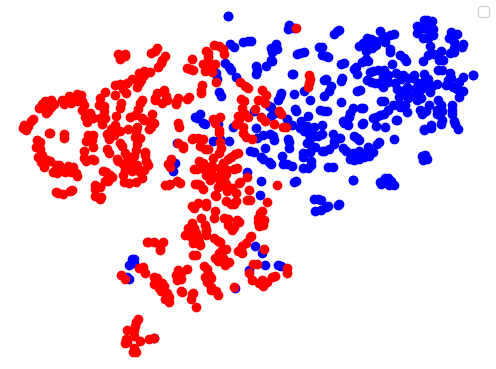} &
    \includegraphics[width=.12\textwidth]{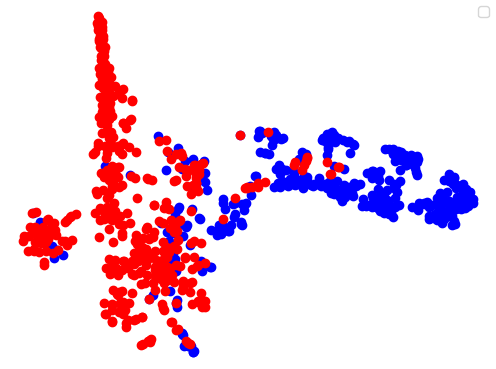} &
    \includegraphics[width=.12\textwidth]{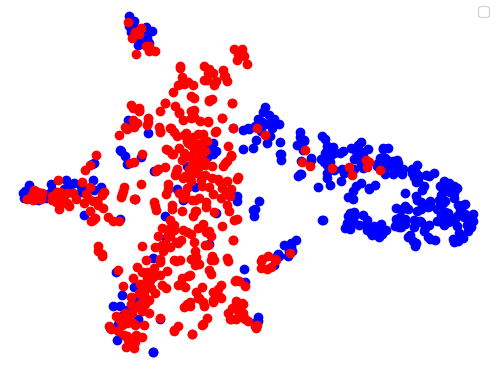} &
    \includegraphics[width=.12\textwidth]{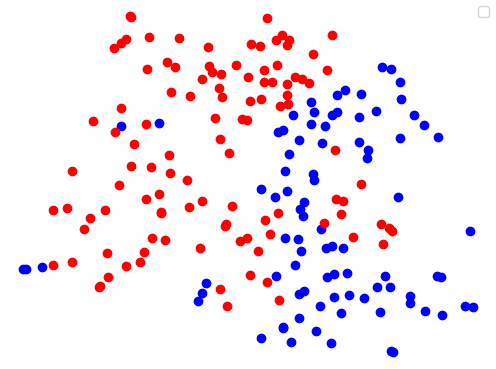} \\
    &
    \includegraphics[width=.12\textwidth]{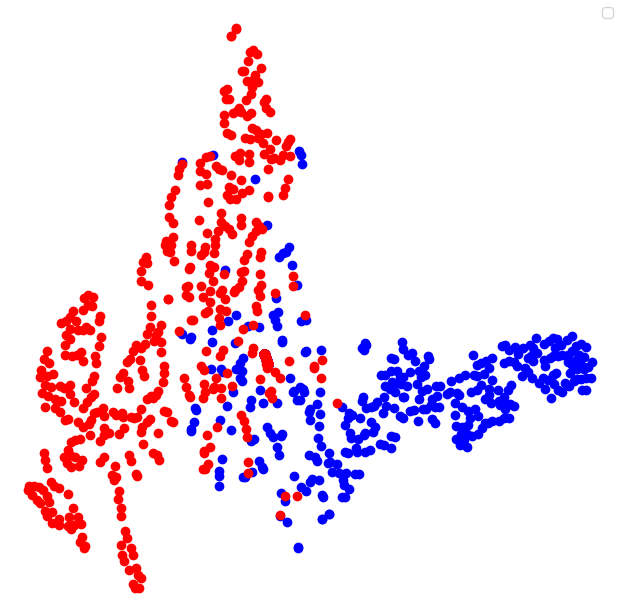} &
    \includegraphics[width=.12\textwidth]{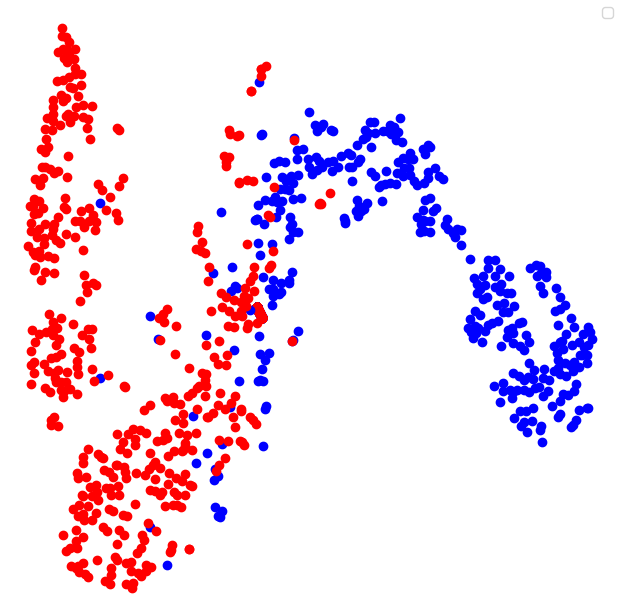} &
    \includegraphics[width=.12\textwidth]{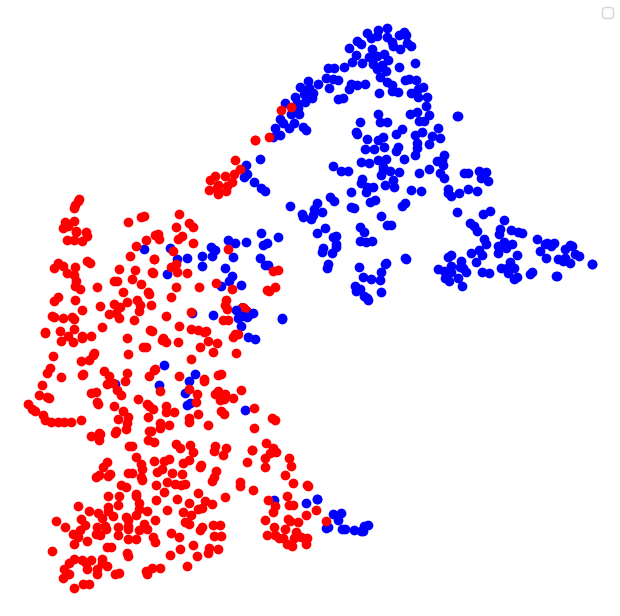} &
    \includegraphics[width=.12\textwidth]{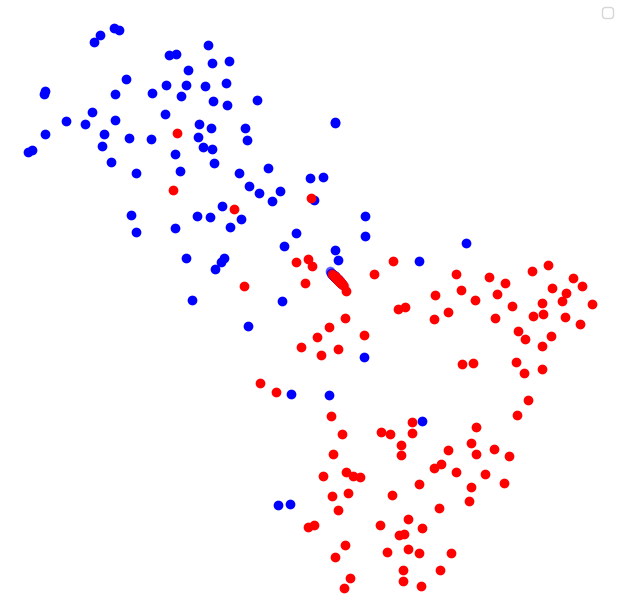} \\
    &
    \includegraphics[width=.12\textwidth]{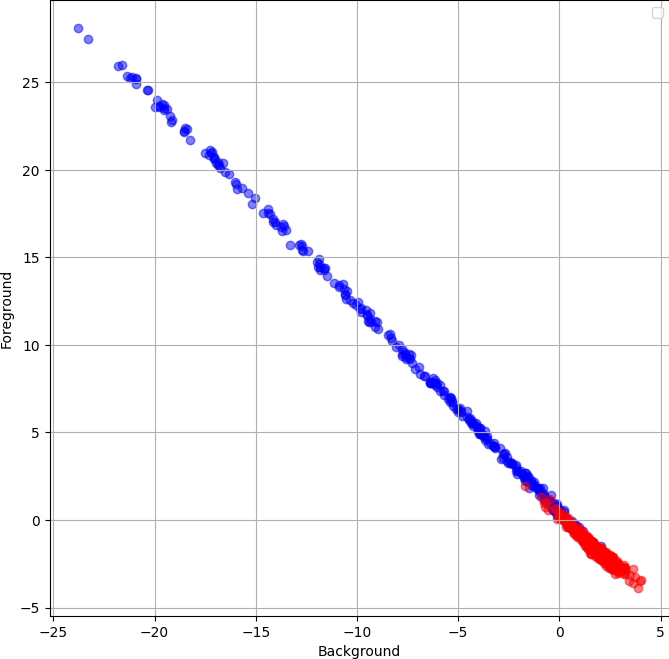} &
    \includegraphics[width=.12\textwidth]{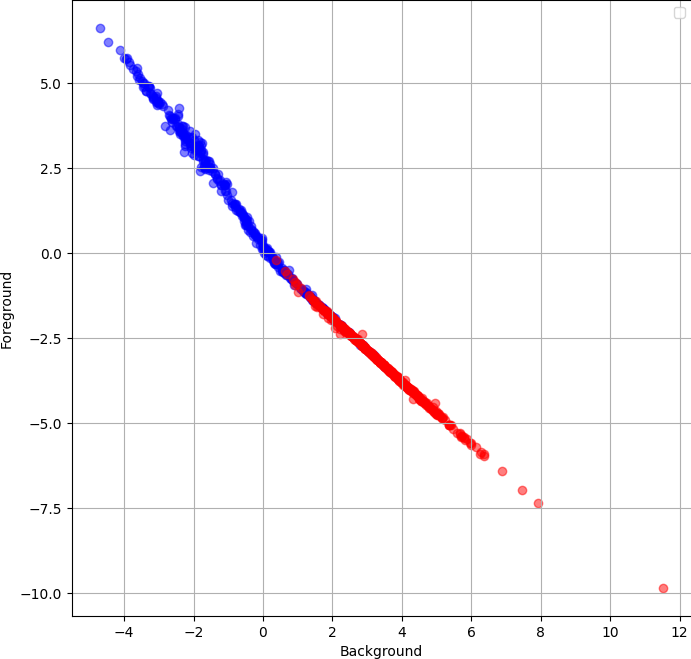} &
    \includegraphics[width=.12\textwidth]{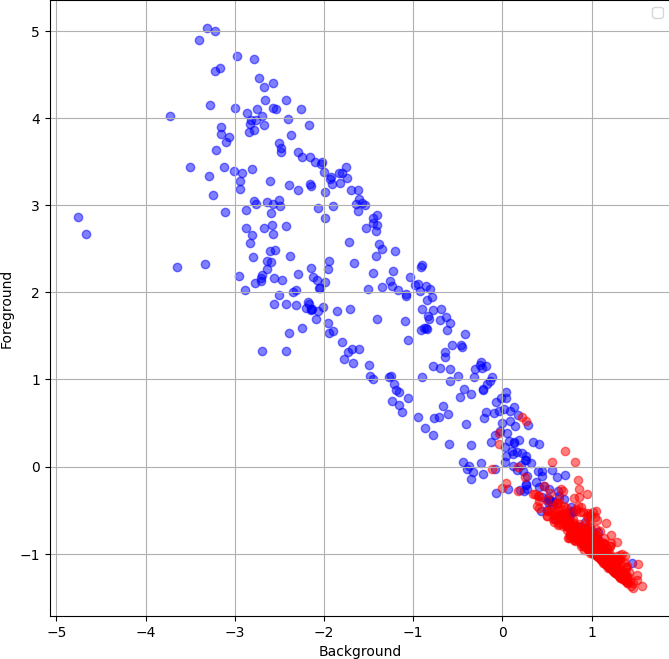} &
    \includegraphics[width=.12\textwidth]{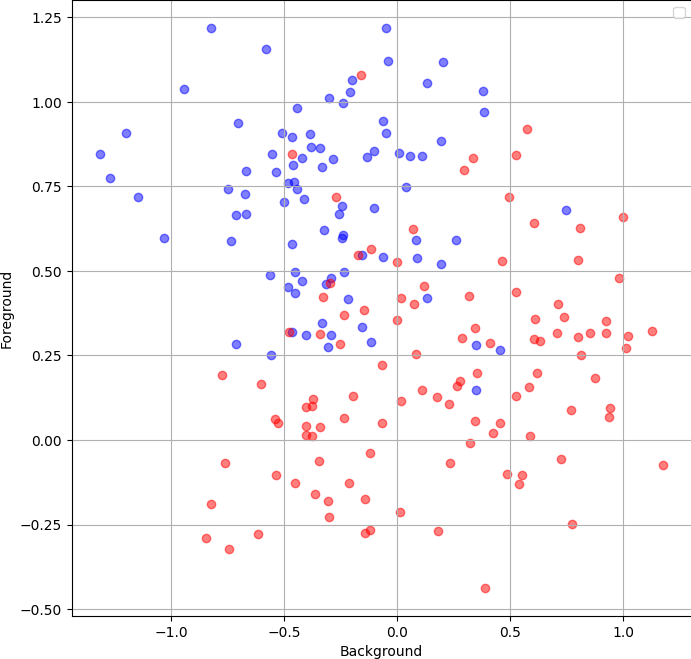}
  \end{tabular}
  \caption{
  Standard WSOL setup: For the three \textbf{normal} images from \glas, we display the t-SNE projection of foreground and background pixel-features of WSOL baseline methods without (1st row) and with (2nd row) \pixelcam. The 3rd row presents the logits of the pixel classifier of \pixelcam.
  }
  \label{fig:feature-tsne-source-normal-glas}
\end{figure*}

\begin{figure*}[!ht]
\centering
  \begin{tabular}{@{}c@{\hskip 10pt}c@{\hskip 10pt}c@{\hskip 10pt}c@{\hskip 10pt}c@{}}
    \makecell{Input} & \makecell{DeepMil} & \makecell{GradCAM++} & \makecell{LayerCAM} & \makecell{SAT} \\
   \includegraphics[width=.12\textwidth]{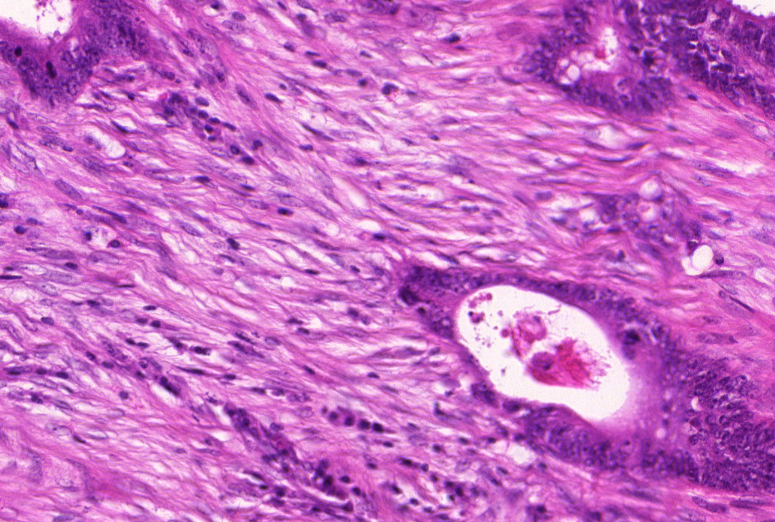} &
    \includegraphics[width=.12\textwidth]{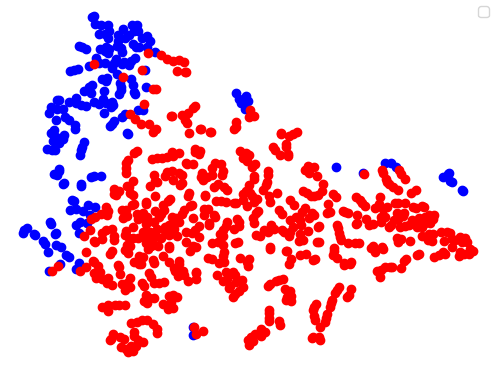} &
    \includegraphics[width=.12\textwidth]{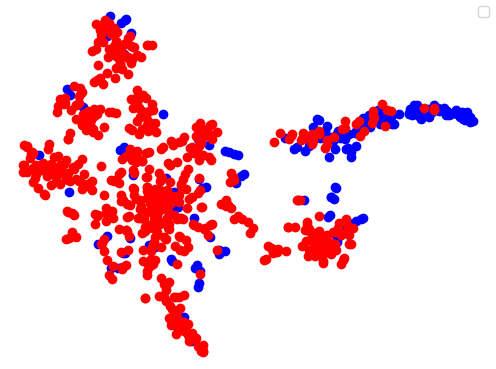} &
    \includegraphics[width=.12\textwidth]{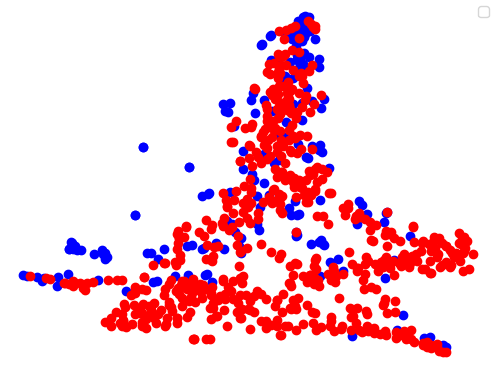} &
    \includegraphics[width=.12\textwidth]{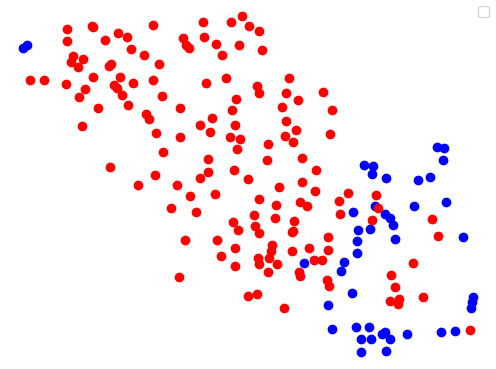} \\
     &
    \includegraphics[width=.12\textwidth]{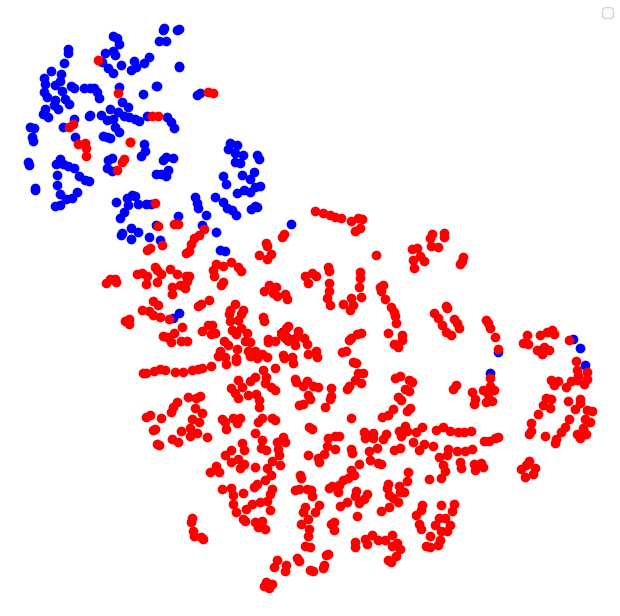} &
    \includegraphics[width=.12\textwidth]{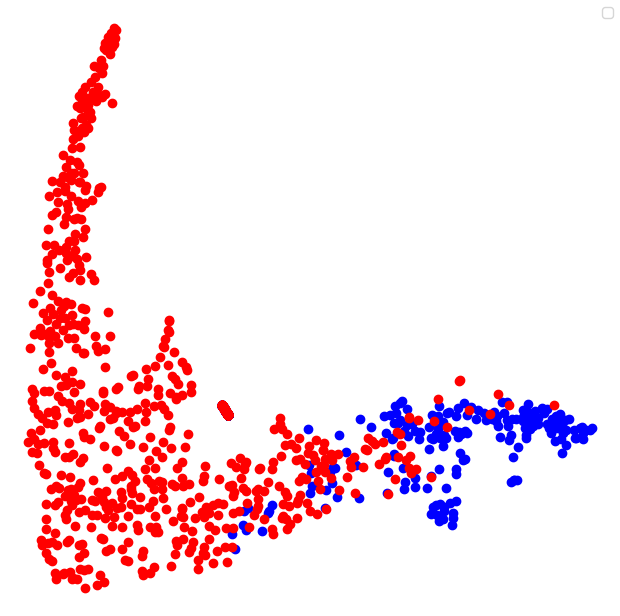} &
    \includegraphics[width=.12\textwidth]{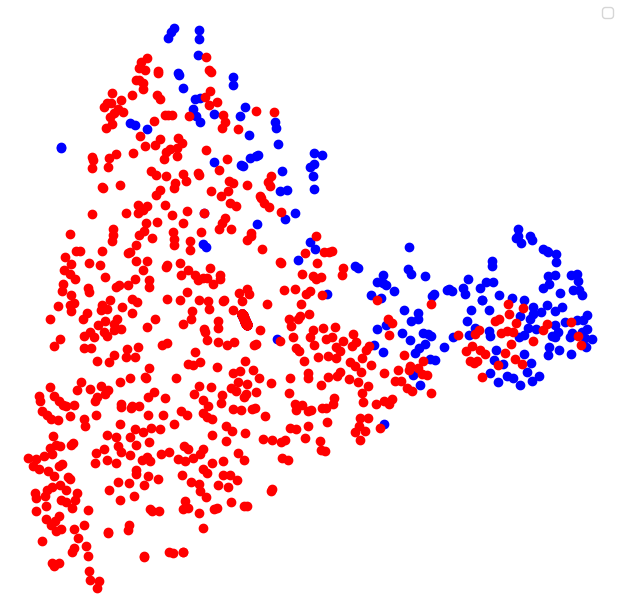} &
    \includegraphics[width=.12\textwidth]{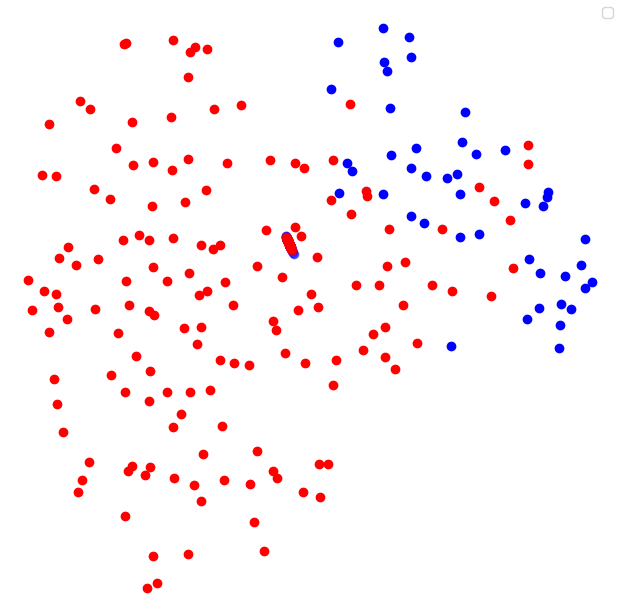} \\

     &
    \includegraphics[width=.12\textwidth]{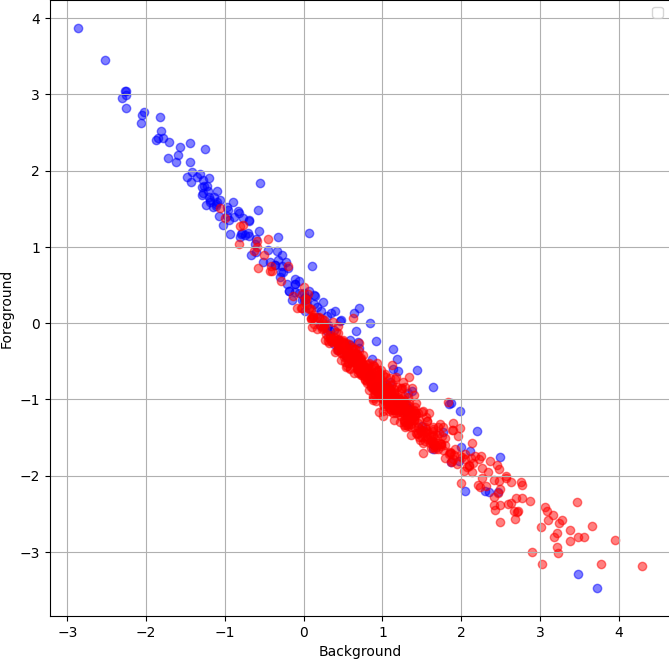} &
    \includegraphics[width=.12\textwidth]{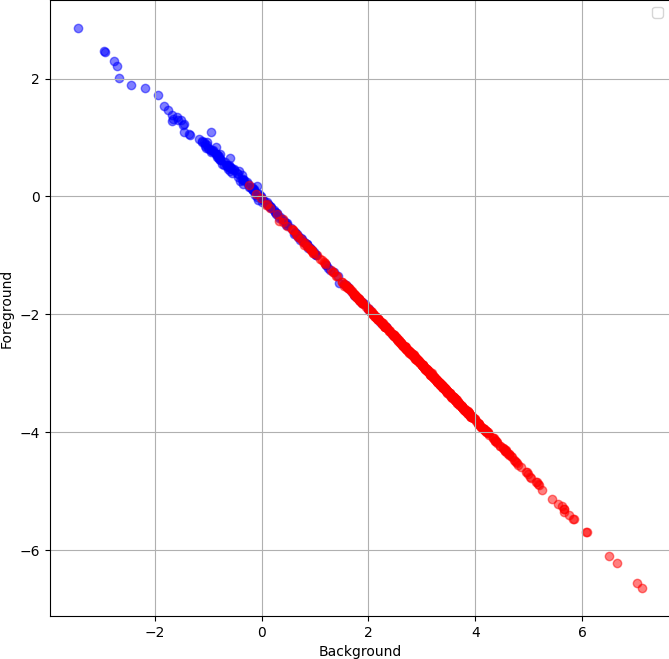} &
    \includegraphics[width=.12\textwidth]{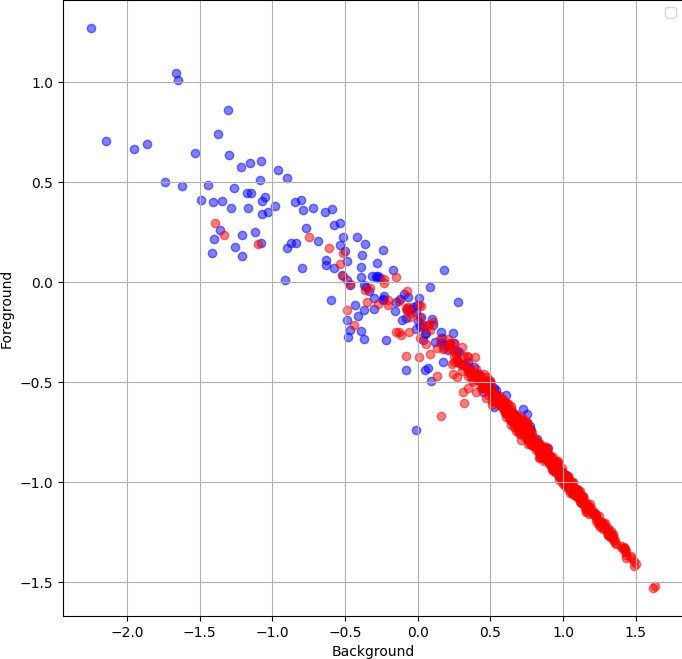} &
    \includegraphics[width=.12\textwidth]{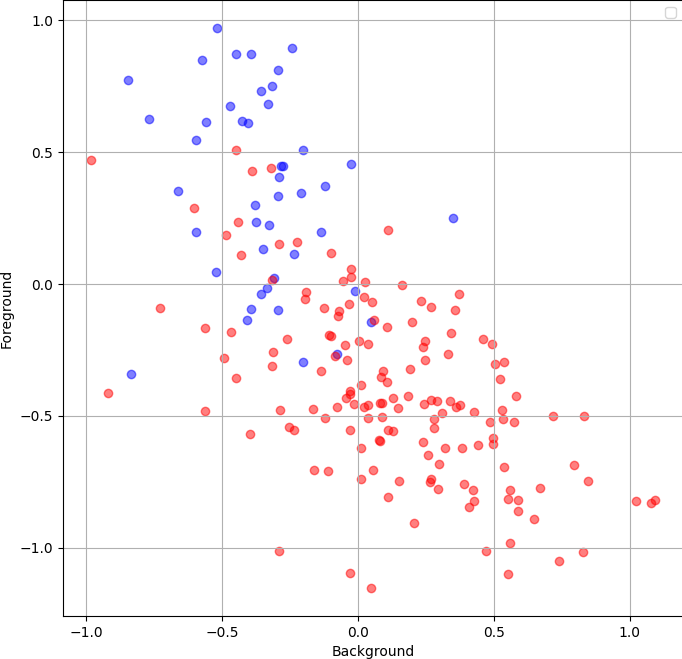} \\

     \includegraphics[width=.12\textwidth]{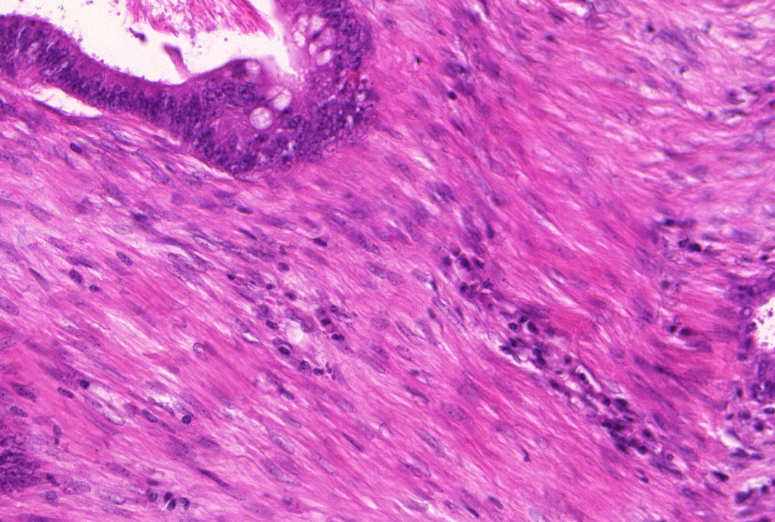} &
    \includegraphics[width=.12\textwidth]{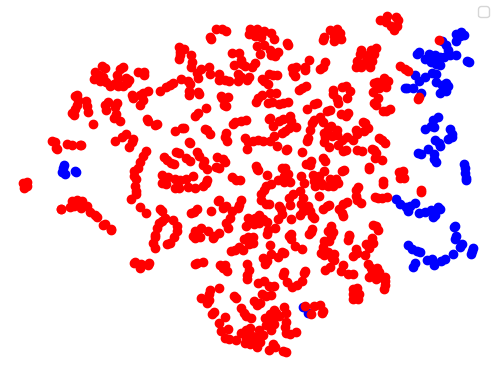} &
    \includegraphics[width=.12\textwidth]{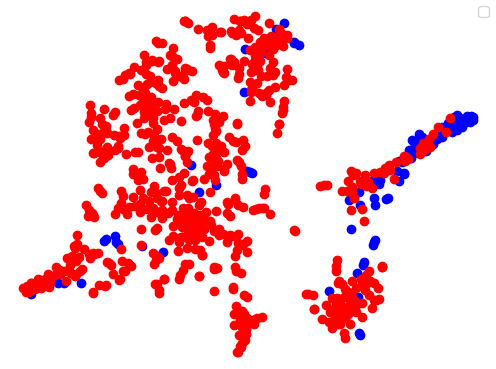} &
    \includegraphics[width=.12\textwidth]{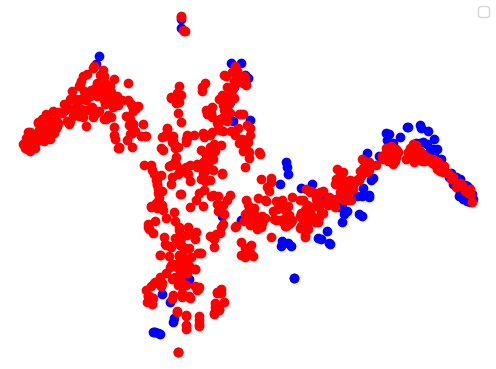} &
    \includegraphics[width=.12\textwidth]{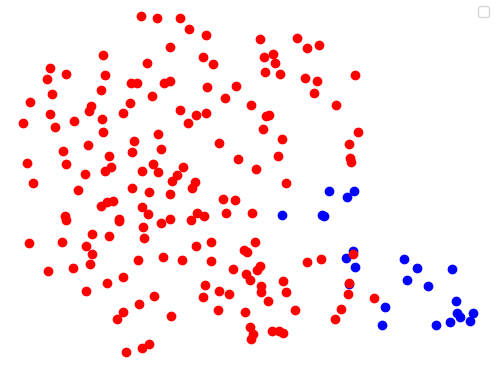} \\
     &
    \includegraphics[width=.12\textwidth]{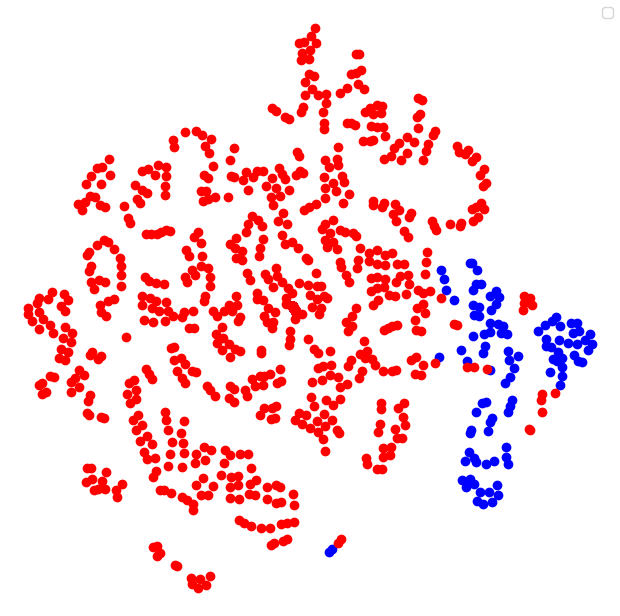} &
    \includegraphics[width=.12\textwidth]{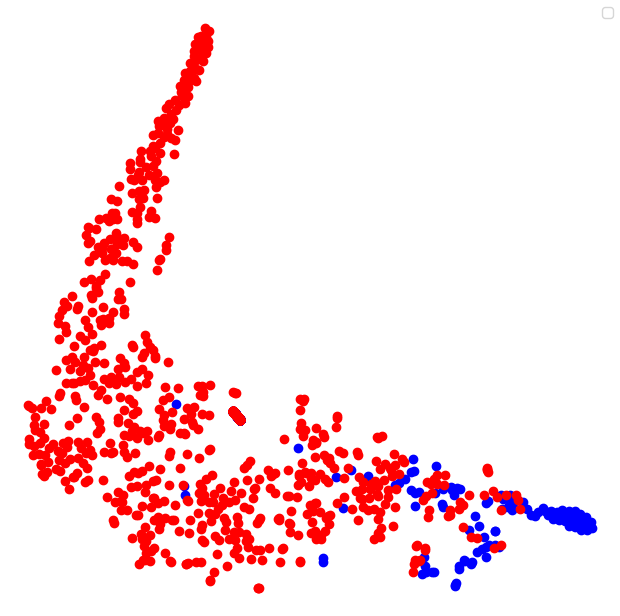} &
    \includegraphics[width=.12\textwidth]{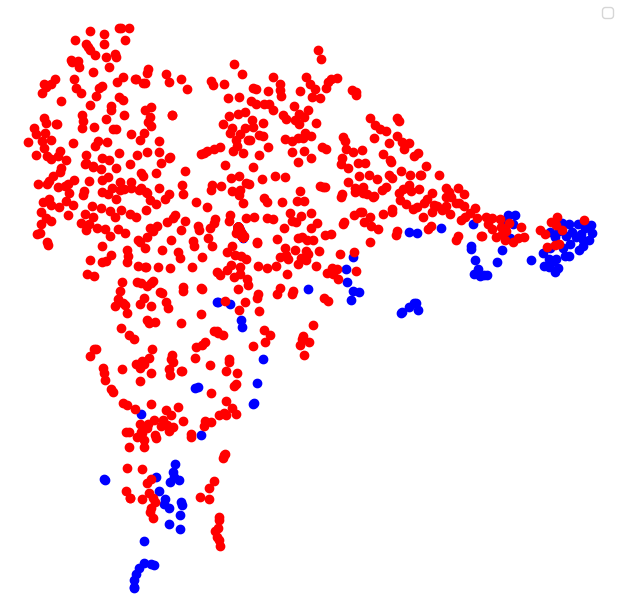} &
    \includegraphics[width=.12\textwidth]{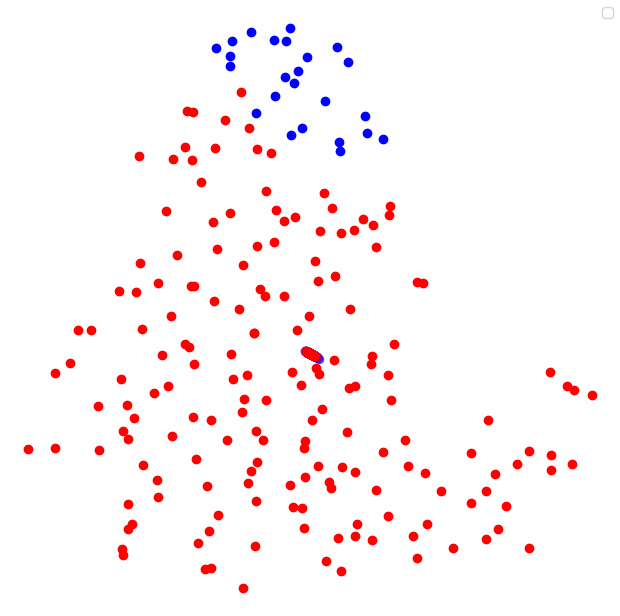} \\
     &
    \includegraphics[width=.12\textwidth]{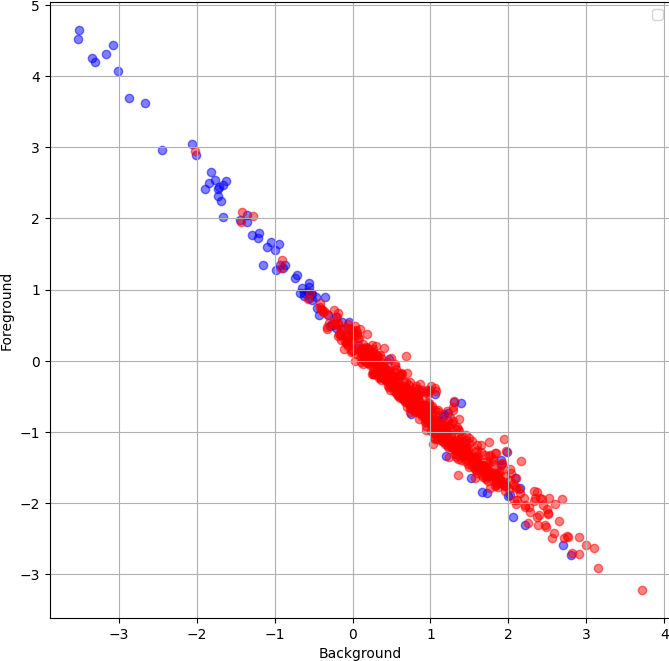} &
    \includegraphics[width=.12\textwidth]{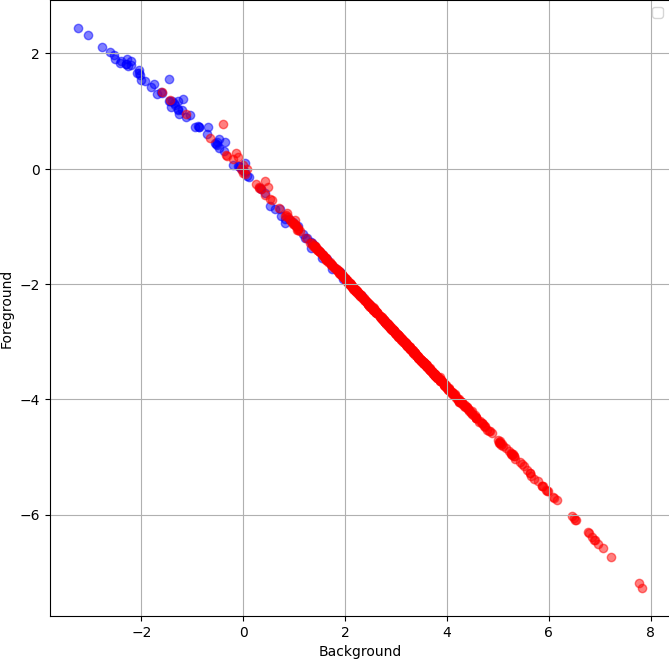} &
    \includegraphics[width=.12\textwidth]{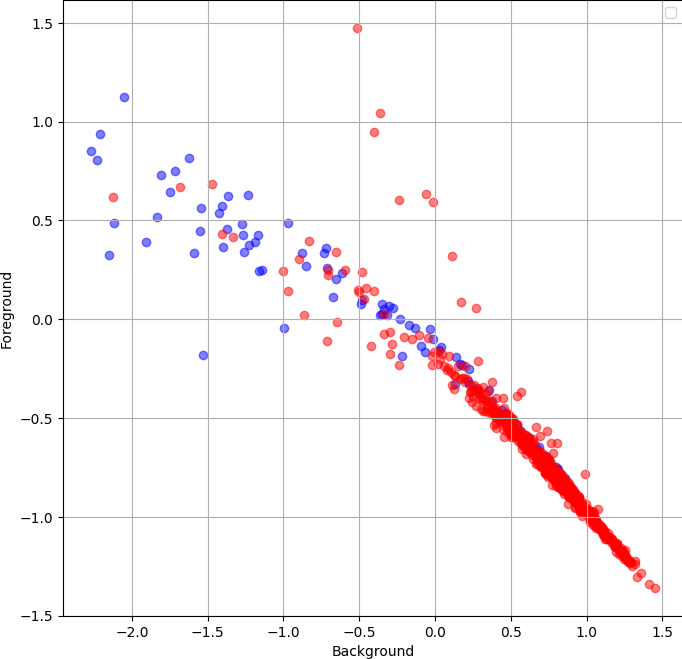} &
    \includegraphics[width=.12\textwidth]{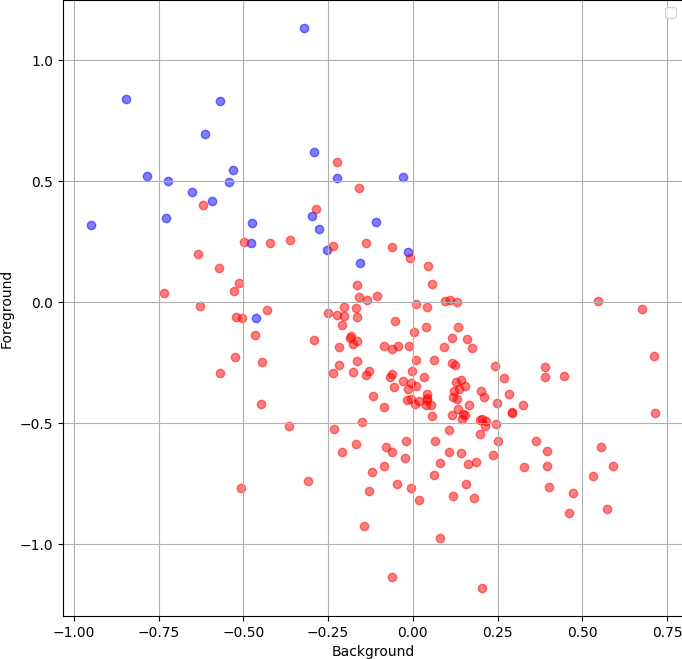} \\
    \includegraphics[width=.12\textwidth]{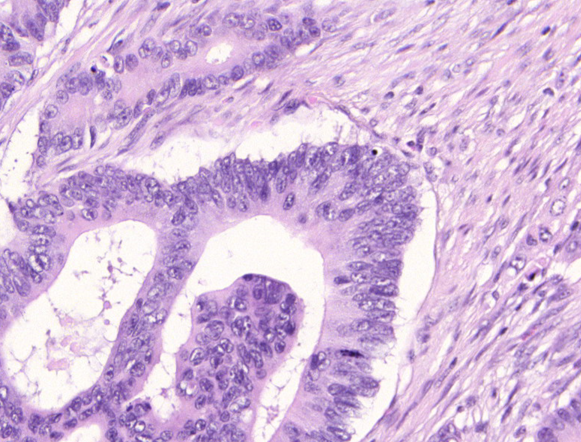} &
    \includegraphics[width=.12\textwidth]{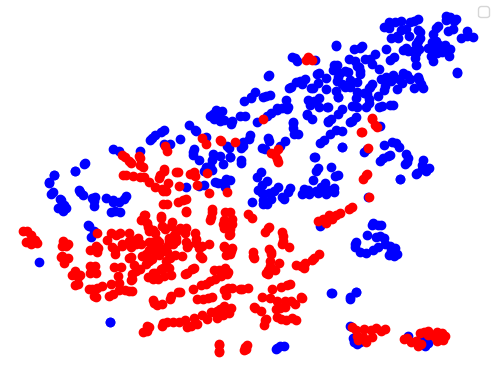} &
    \includegraphics[width=.12\textwidth]{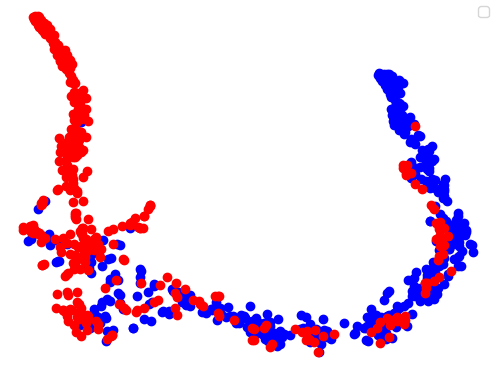} &
    \includegraphics[width=.12\textwidth]{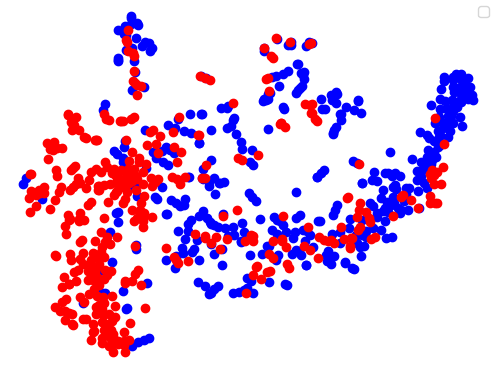} &
    \includegraphics[width=.12\textwidth]{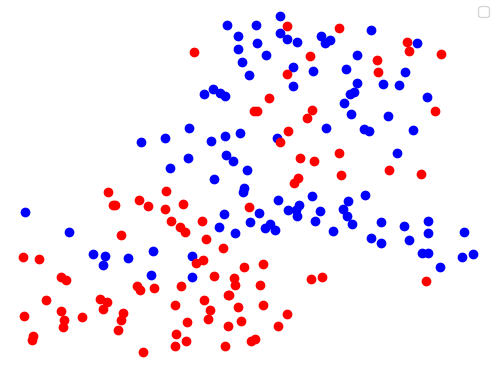} \\
    &
    \includegraphics[width=.12\textwidth]{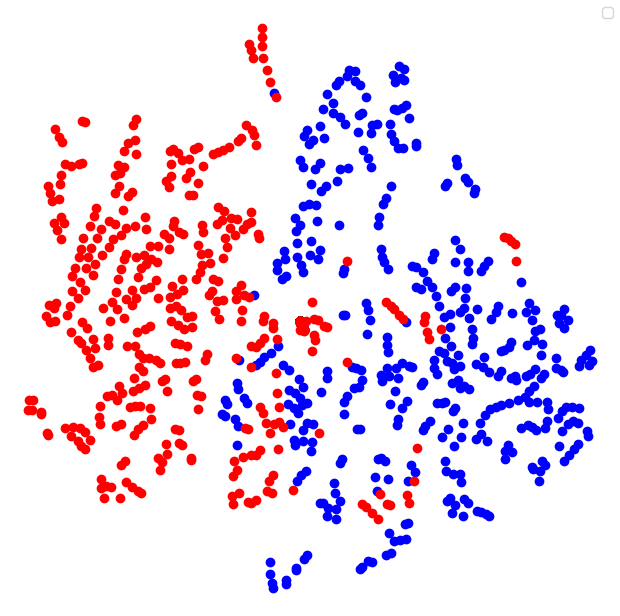} &
    \includegraphics[width=.12\textwidth]{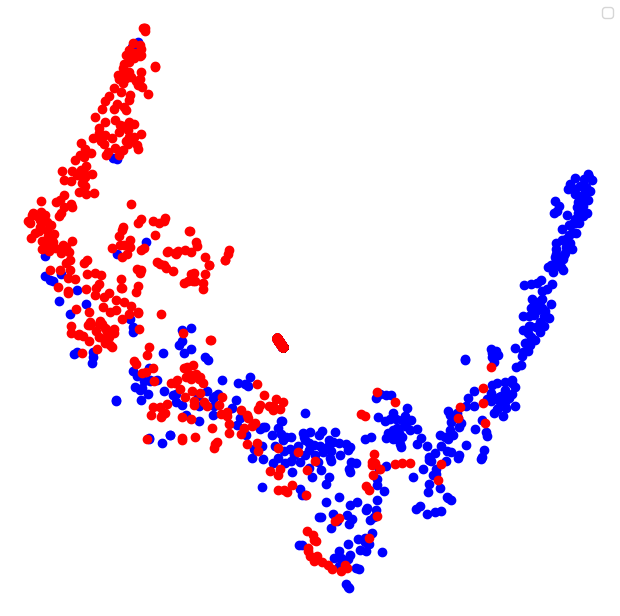} &
    \includegraphics[width=.12\textwidth]{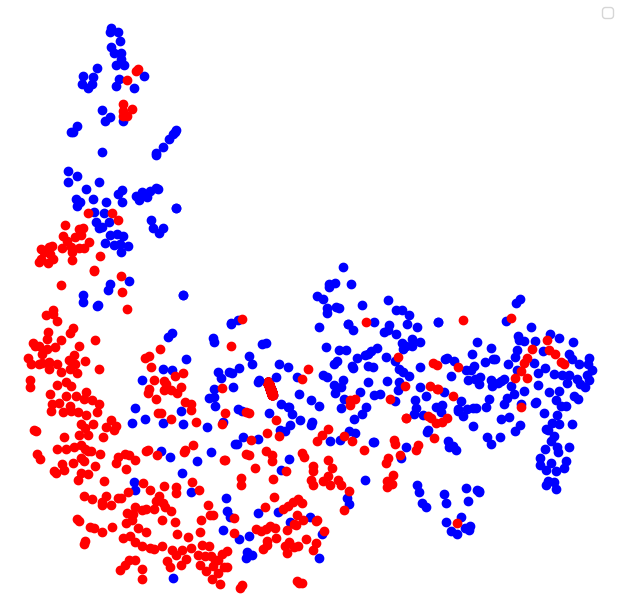} &
    \includegraphics[width=.12\textwidth]{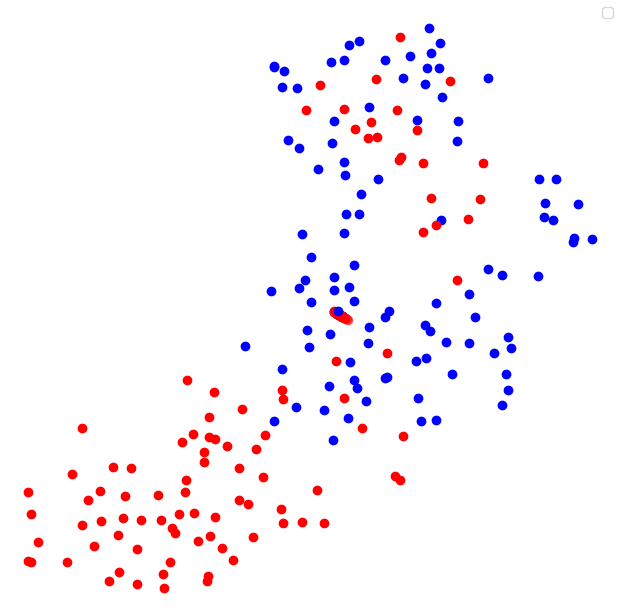} \\
     &
    \includegraphics[width=.12\textwidth]{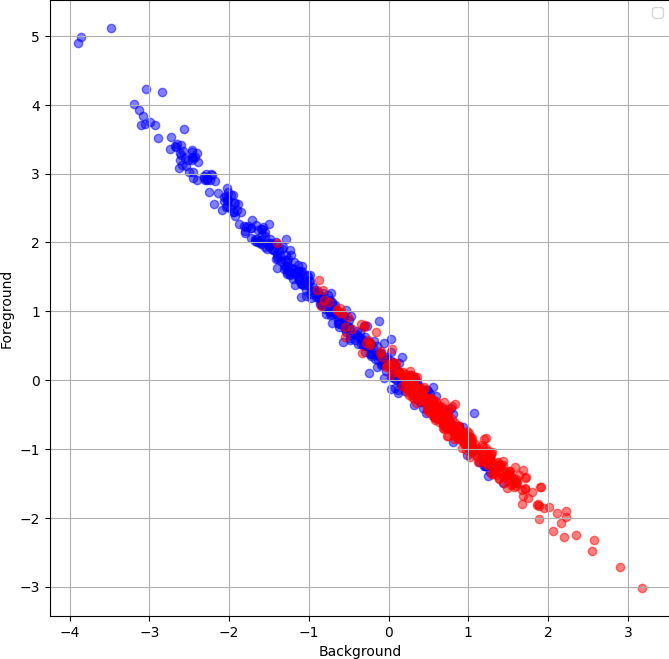} &
    \includegraphics[width=.12\textwidth]{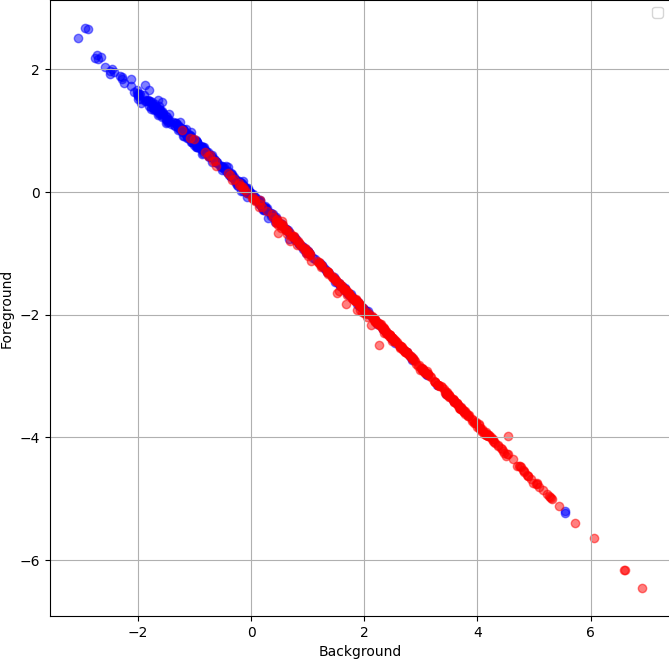} &
    \includegraphics[width=.12\textwidth]{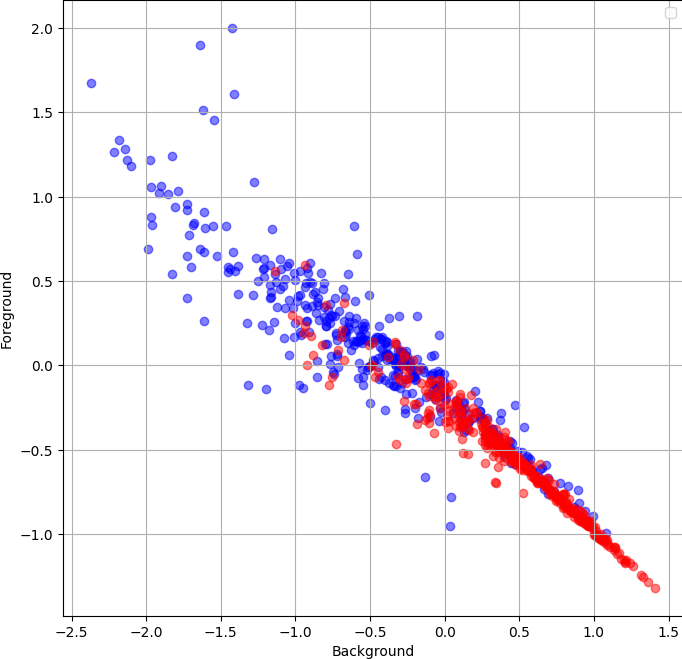} &
    \includegraphics[width=.12\textwidth]{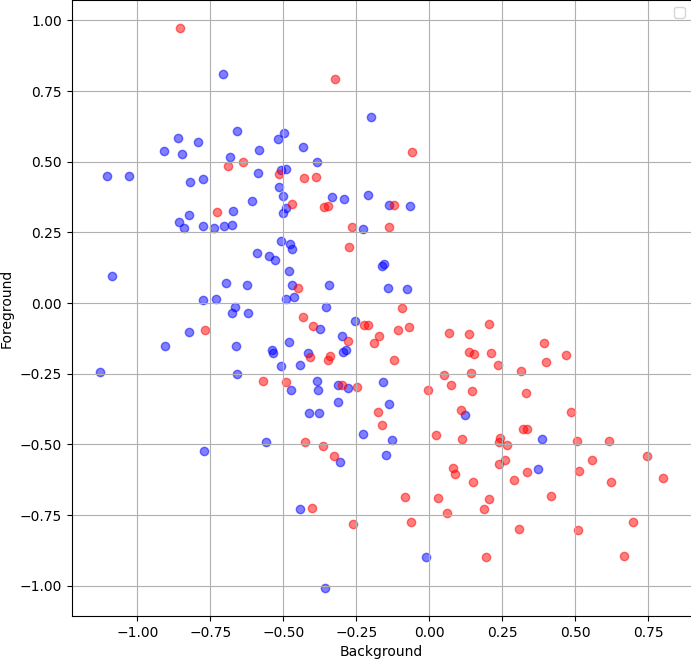} 
  \end{tabular}
  \caption{
  Standard WSOL setup: For the three \textbf{cancerous} images from \glas, we display the t-SNE projection of foreground and background pixel-features of WSOL baseline methods without (1st row) and with (2nd row) \pixelcam. The 3rd row presents the logits of the pixel classifier of \pixelcam.
  }
  \label{fig:feature-tsne-source-cancer-glas}
\end{figure*}

\begin{figure*}[!ht]
\centering
  \begin{tabular}{@{}c@{\hskip 10pt}c@{\hskip 10pt}c@{\hskip 10pt}c@{\hskip 10pt}c@{}}
    \makecell{Input} & \makecell{DeepMil} & \makecell{GradCAM++} & \makecell{LayerCAM} & \makecell{SAT} \\
    \includegraphics[width=.12\textwidth]{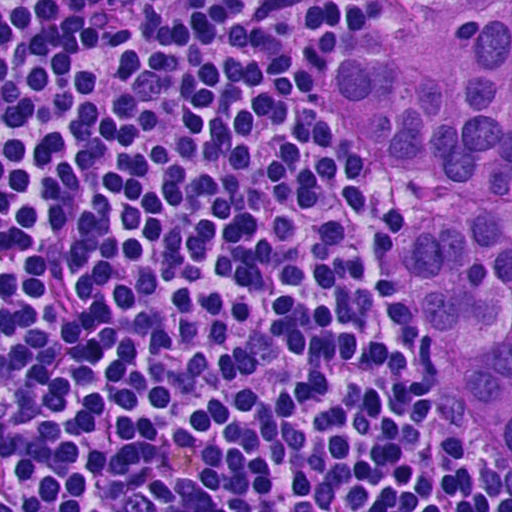} &
    \includegraphics[width=.12\textwidth]{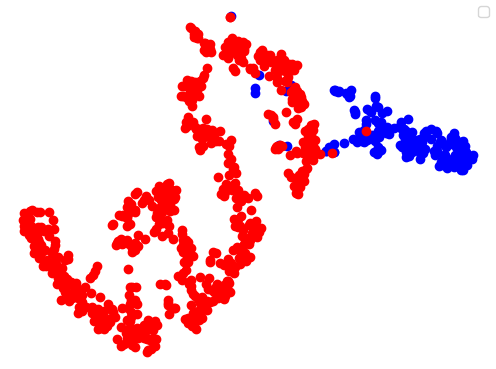} &
    \includegraphics[width=.12\textwidth]{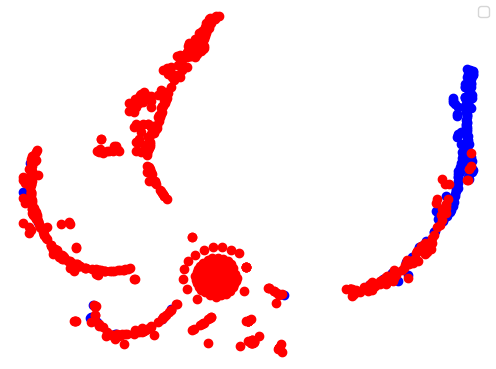} &
    \includegraphics[width=.12\textwidth]{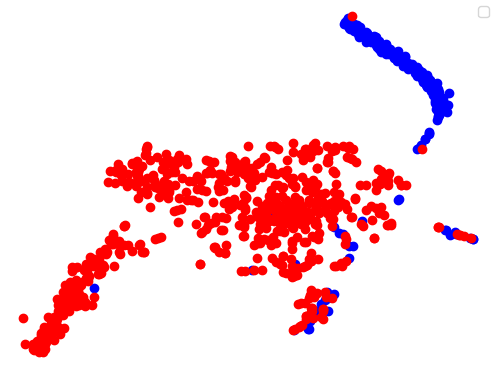} &
    \includegraphics[width=.12\textwidth]{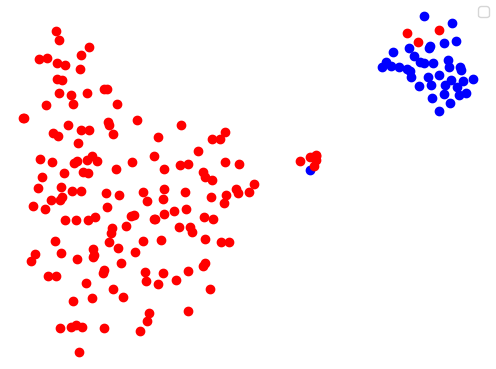} \\
    &
    \includegraphics[width=.12\textwidth]{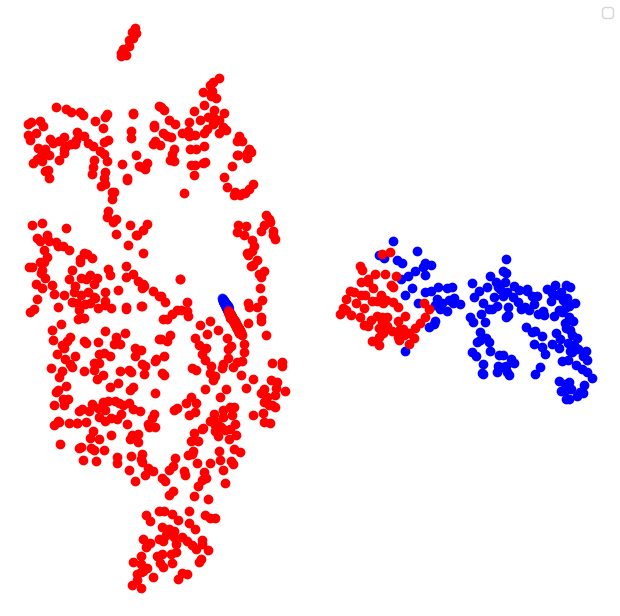} &
    \includegraphics[width=.12\textwidth]{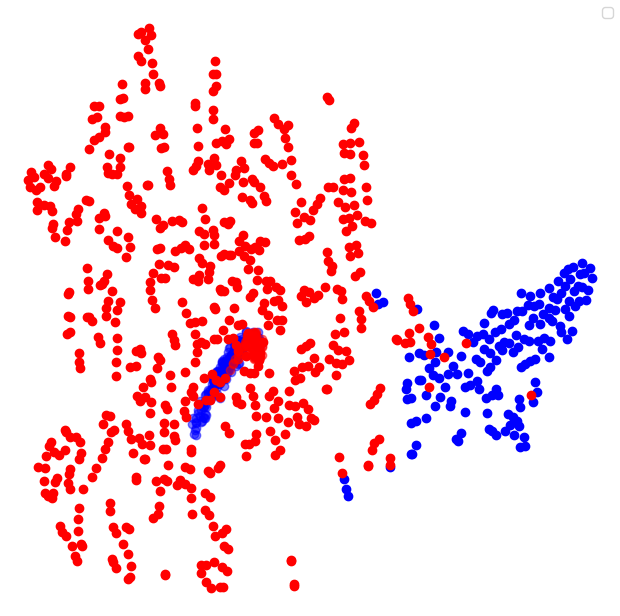} &
    \includegraphics[width=.12\textwidth]{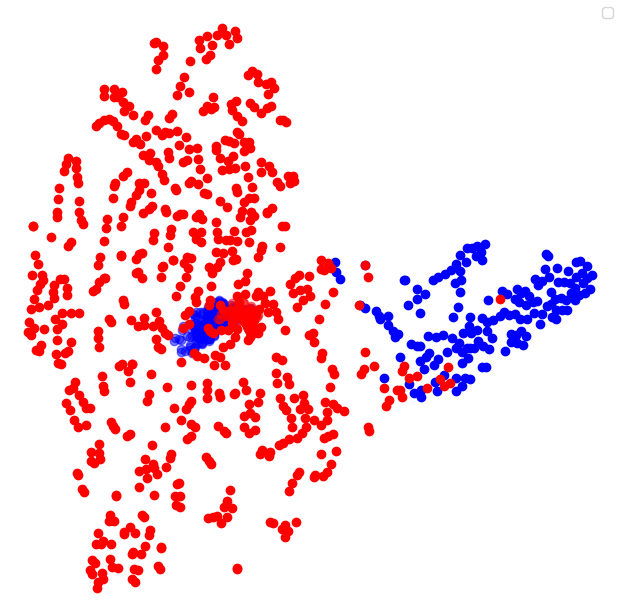} &
    \includegraphics[width=.12\textwidth]{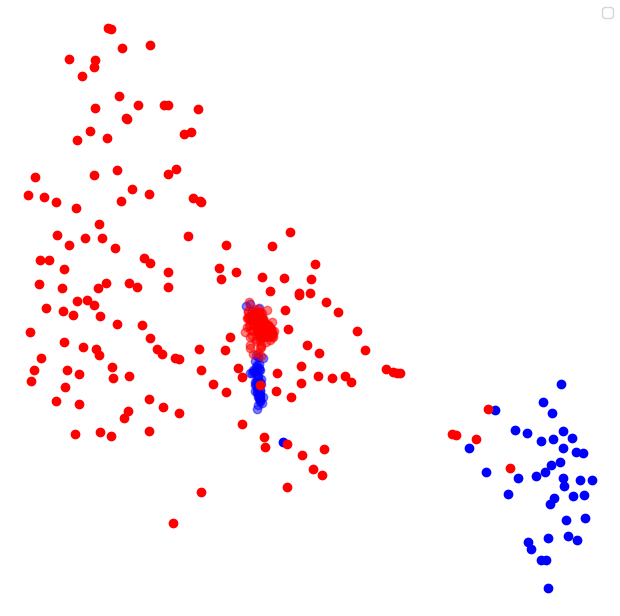} \\
    &
    \includegraphics[width=.12\textwidth]{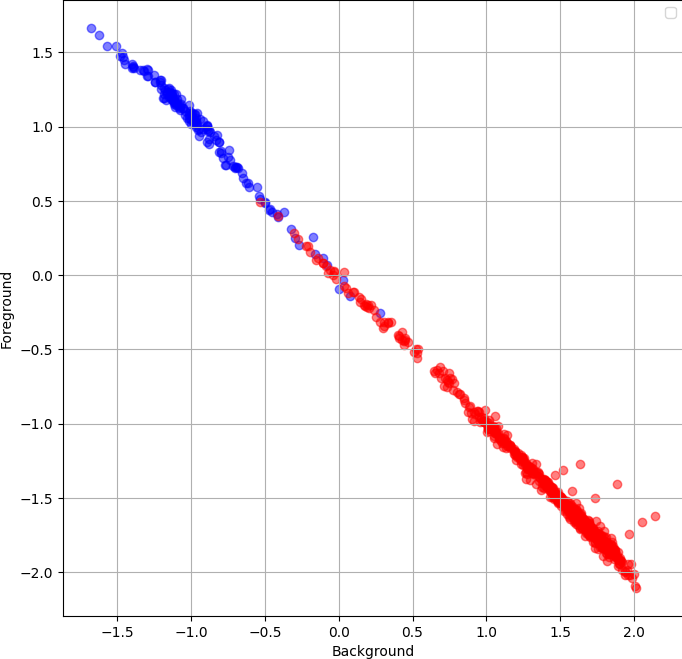} &
    \includegraphics[width=.12\textwidth]{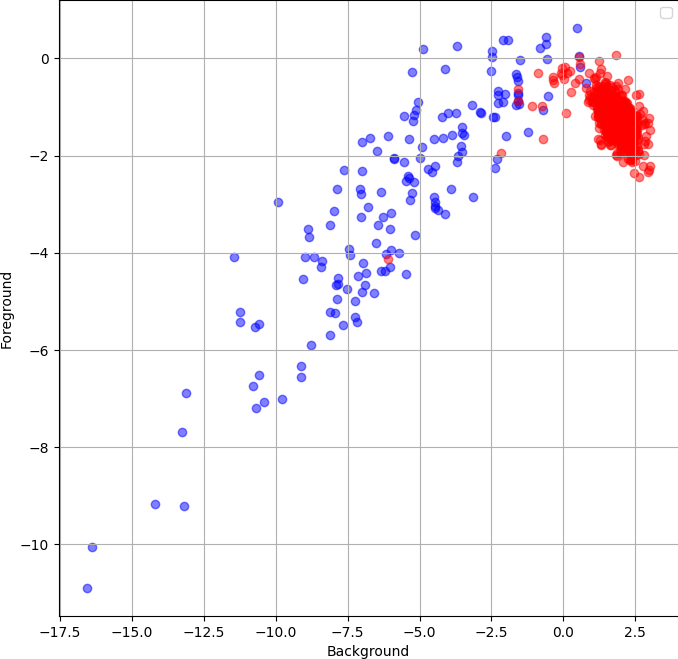} &
    \includegraphics[width=.12\textwidth]{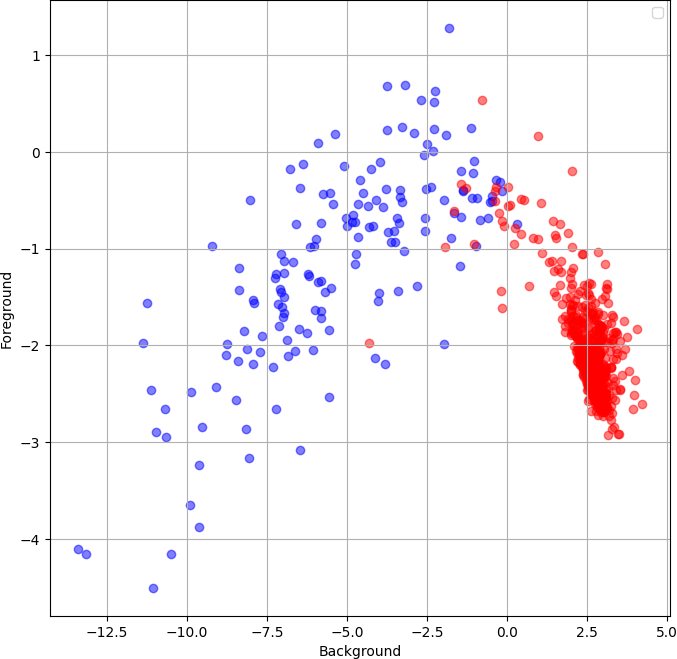} &
    \includegraphics[width=.12\textwidth]{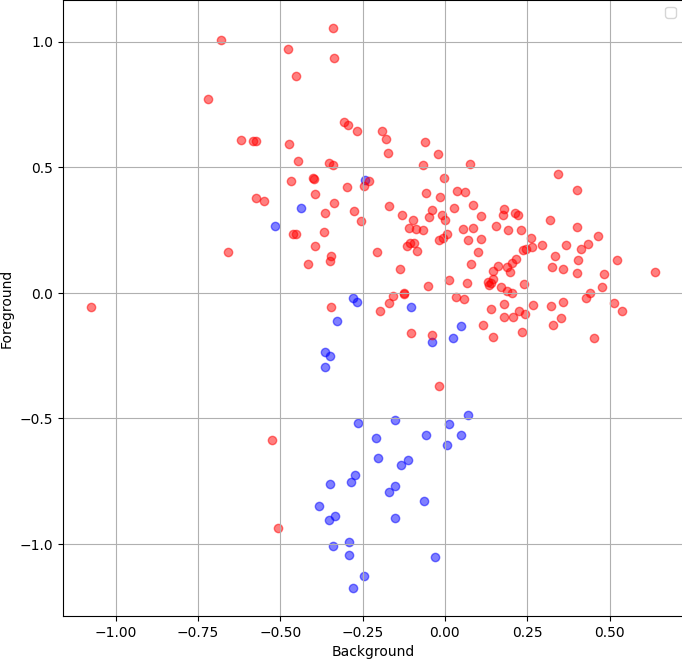} \\

     \includegraphics[width=.12\textwidth]{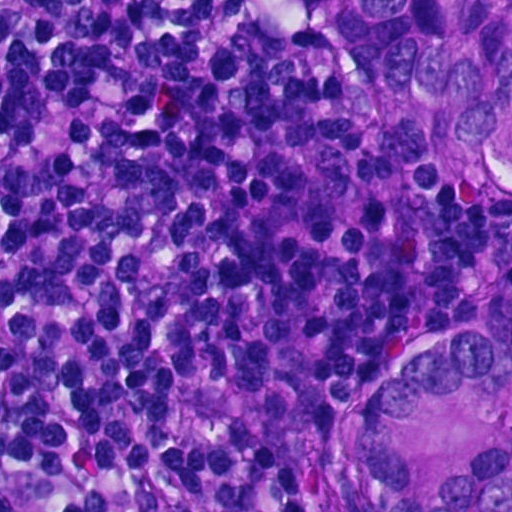} &
    \includegraphics[width=.12\textwidth]{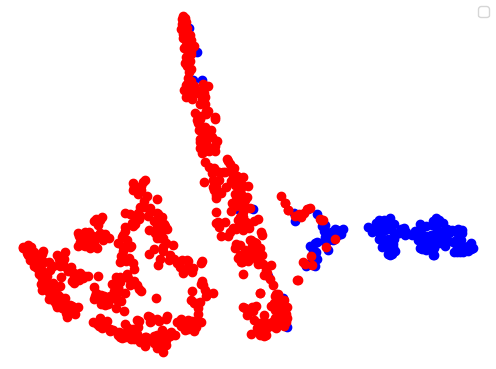} &
    \includegraphics[width=.12\textwidth]{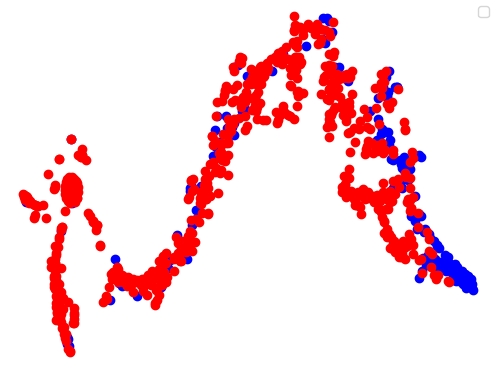} &
    \includegraphics[width=.12\textwidth]{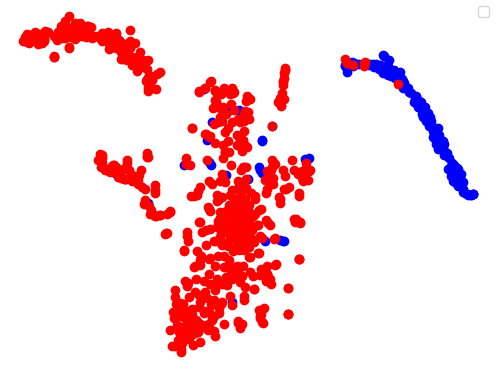} &
    \includegraphics[width=.12\textwidth]{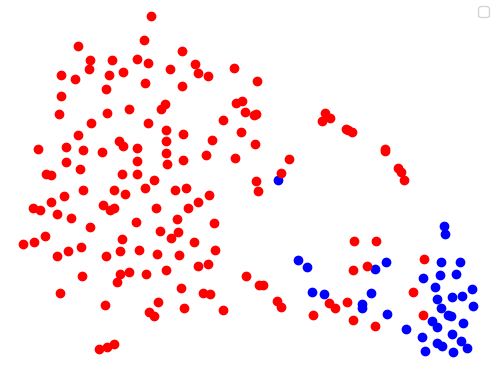} \\
     &
    \includegraphics[width=.12\textwidth]{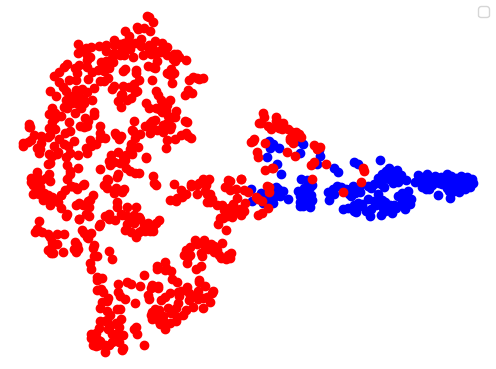} &
    \includegraphics[width=.12\textwidth]{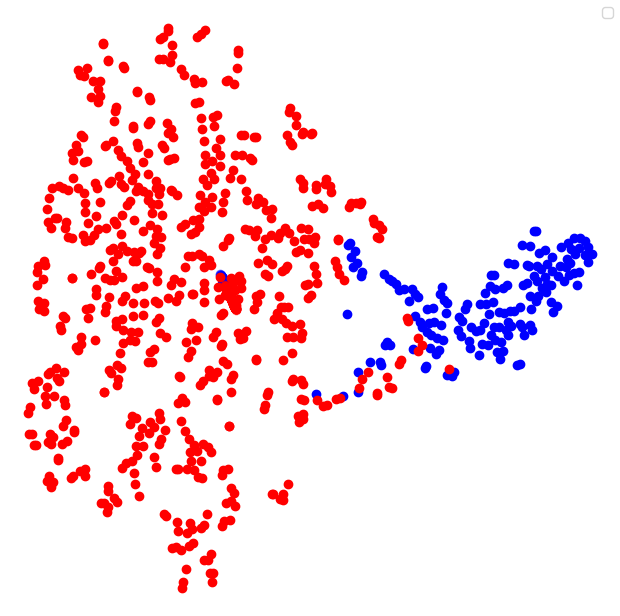} &
    \includegraphics[width=.12\textwidth]{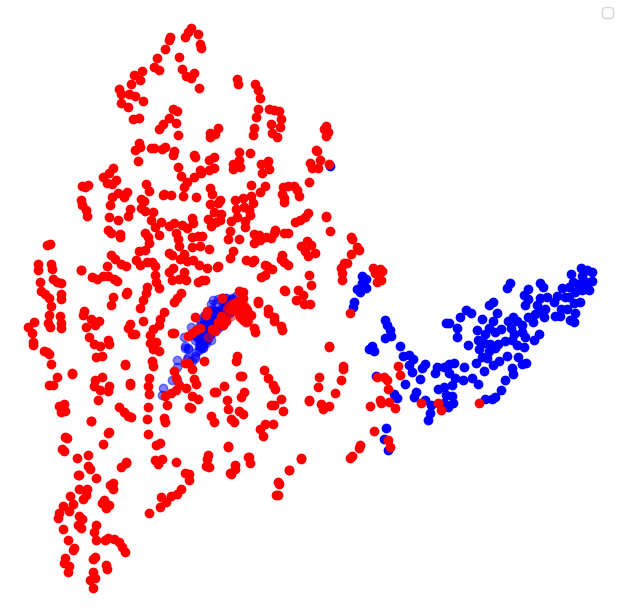} &
    \includegraphics[width=.12\textwidth]{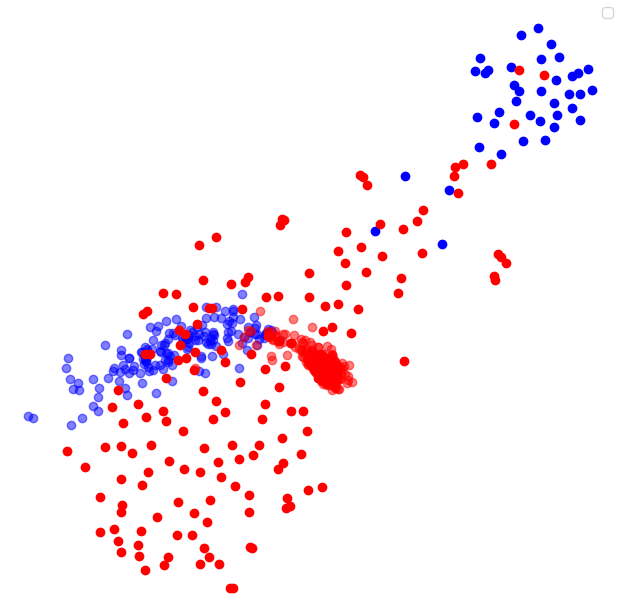} \\
     &
    \includegraphics[width=.12\textwidth]{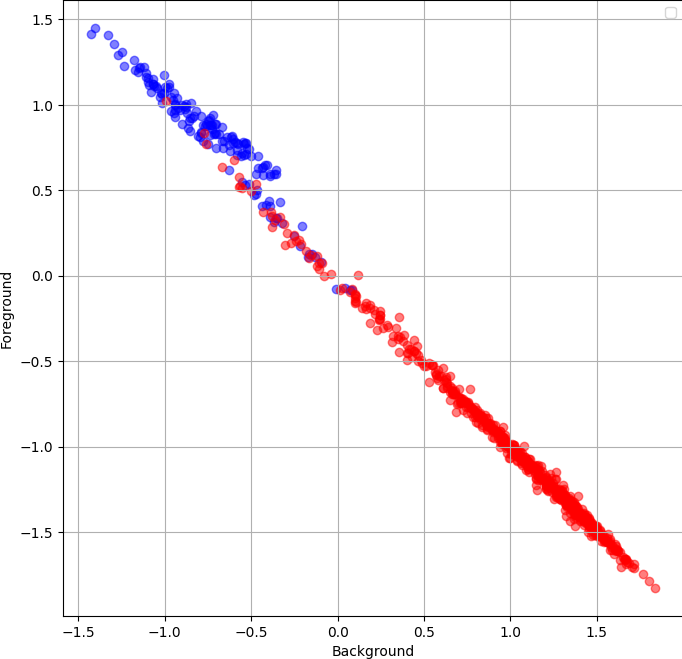} &
    \includegraphics[width=.12\textwidth]{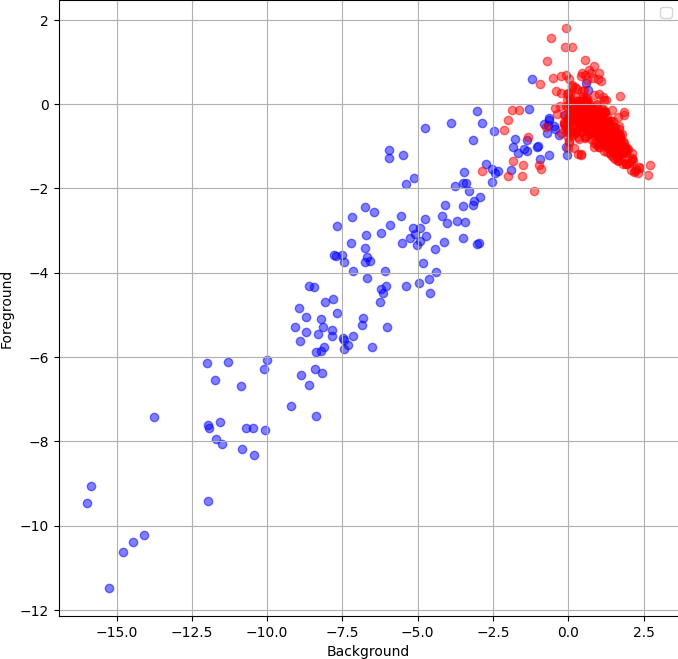} &
    \includegraphics[width=.12\textwidth]{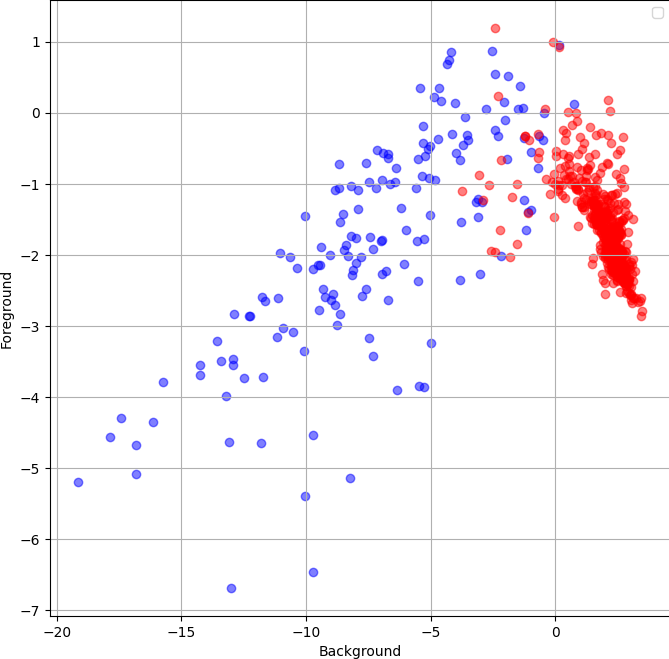} &
    \includegraphics[width=.12\textwidth]{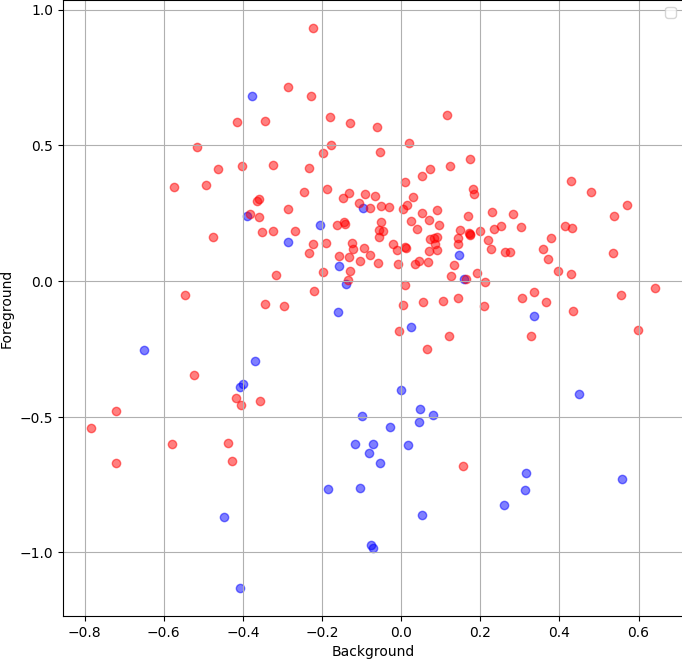} \\

    \includegraphics[width=.12\textwidth]{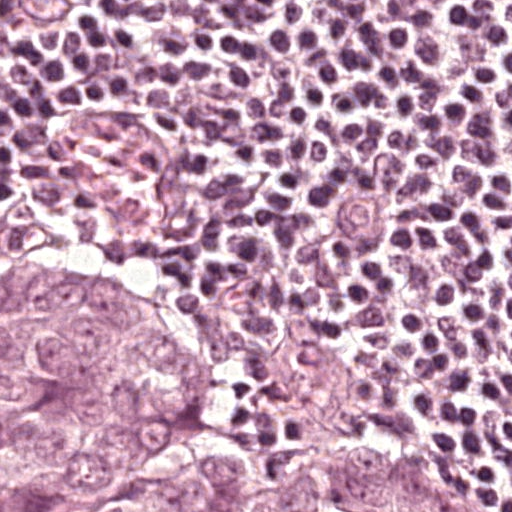} &
    \includegraphics[width=.12\textwidth]{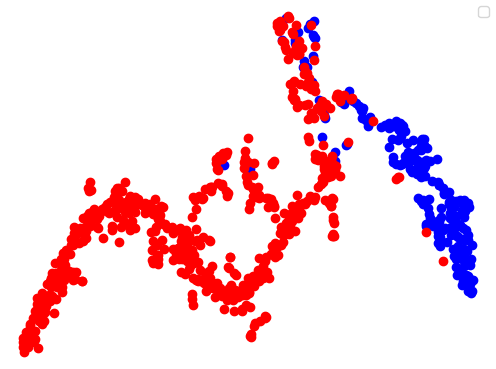} &
    \includegraphics[width=.12\textwidth]{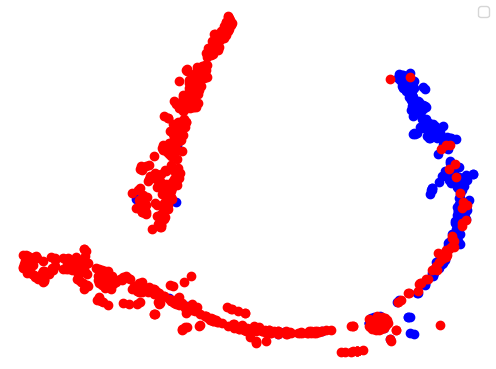} &
    \includegraphics[width=.12\textwidth]{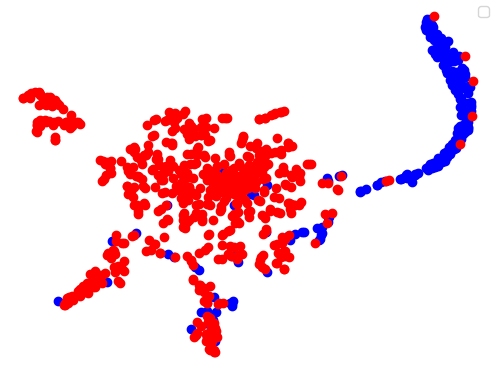} &
    \includegraphics[width=.12\textwidth]{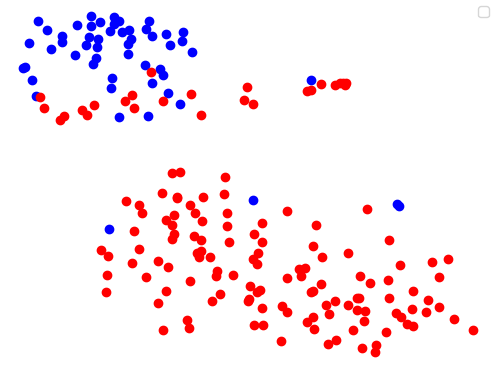} \\
     &
    \includegraphics[width=.12\textwidth]{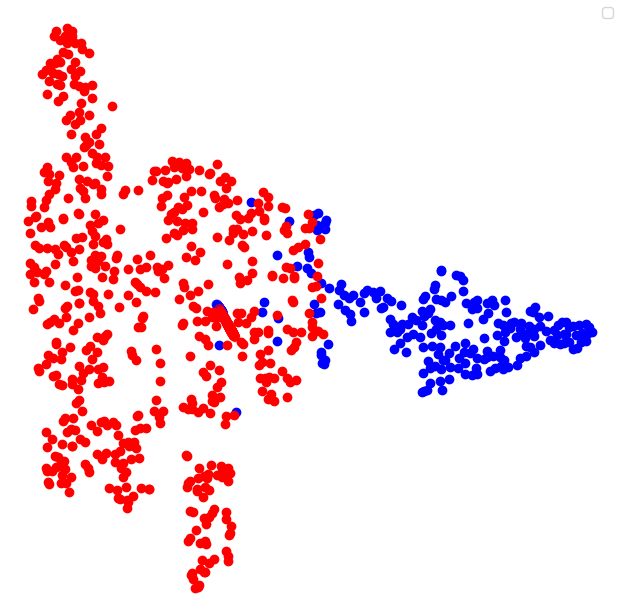} &
    \includegraphics[width=.12\textwidth]{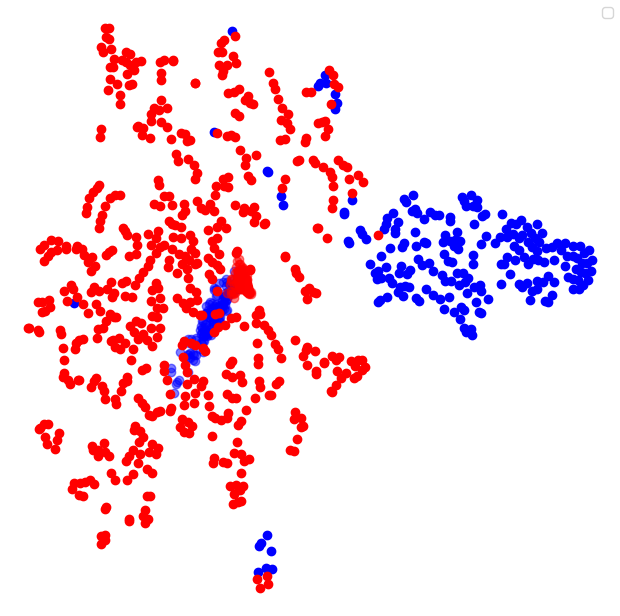} &
    \includegraphics[width=.12\textwidth]{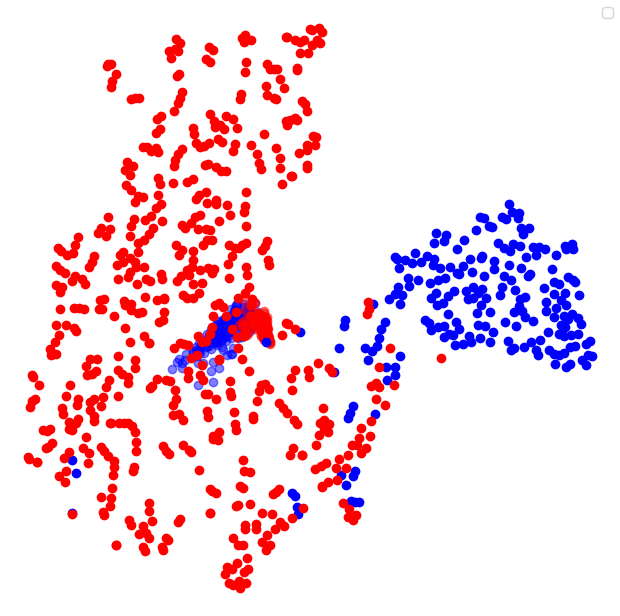} &
    \includegraphics[width=.12\textwidth]{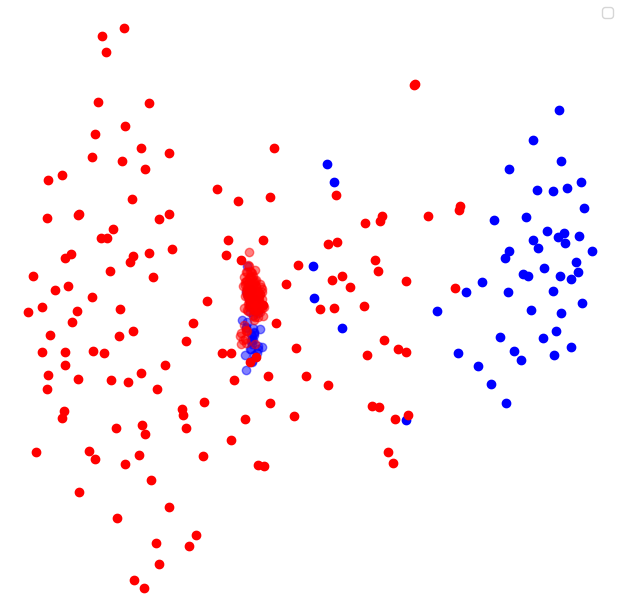} \\
    &
    \includegraphics[width=.12\textwidth]{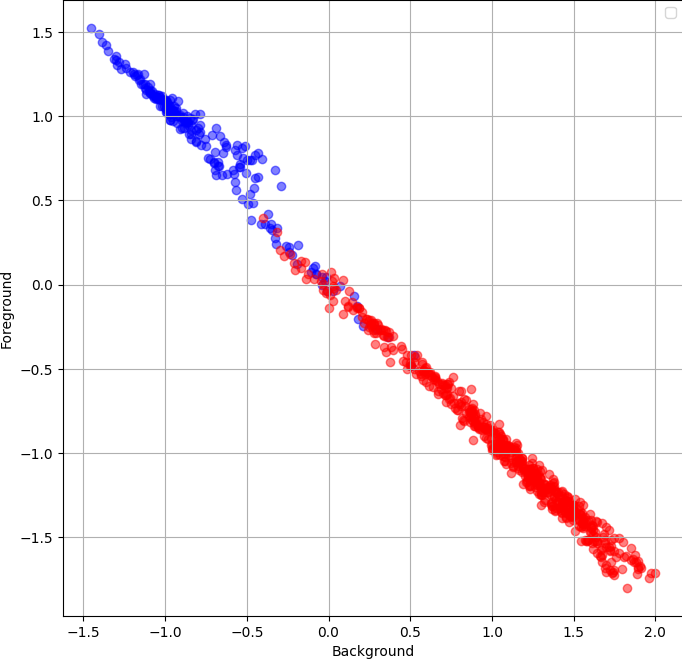} &
    \includegraphics[width=.12\textwidth]{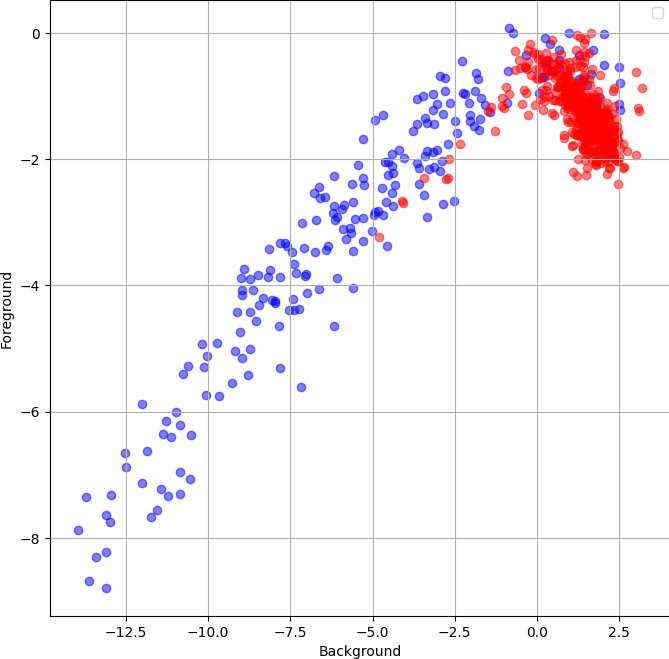} &
    \includegraphics[width=.12\textwidth]{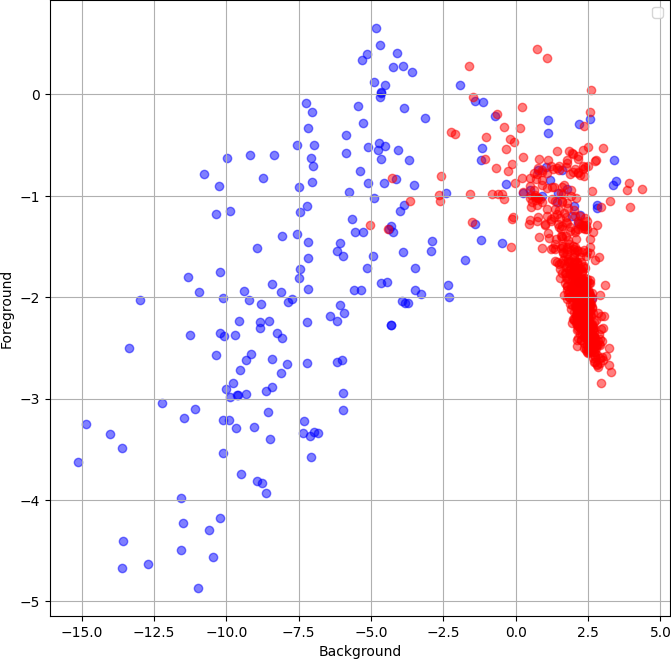} &
    \includegraphics[width=.12\textwidth]{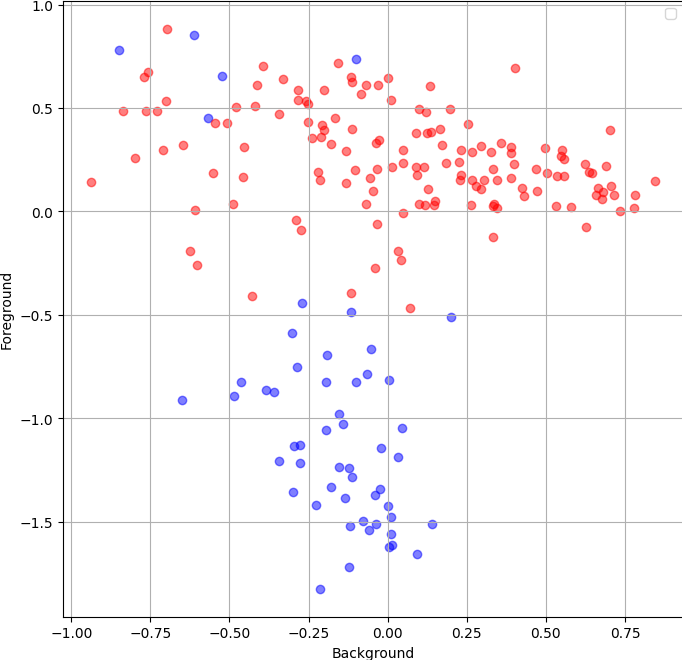} \\
  \end{tabular}
  \caption{Standard WSOL setup: For the three \textbf{cancerous} images from \camsixteen, we display the t-SNE projection of foreground and background pixel-features of WSOL baseline methods without (1st row) and with (2nd row) \pixelcam. The 3rd row presents the logits of the pixel classifier of \pixelcam.
  }
  \label{fig:feature-tsne-source-cancer-cam16}
\end{figure*}

\begin{figure*}[!ht]
\centering
  \begin{tabular}{@{}c@{\hskip 10pt}c@{\hskip 10pt}c@{\hskip 10pt}c@{\hskip 10pt}c@{}}
    \makecell{Input} & \makecell{DeepMil} & \makecell{GradCAM++} & \makecell{LayerCAM} & \makecell{SAT} \\
   \includegraphics[width=.12\textwidth]{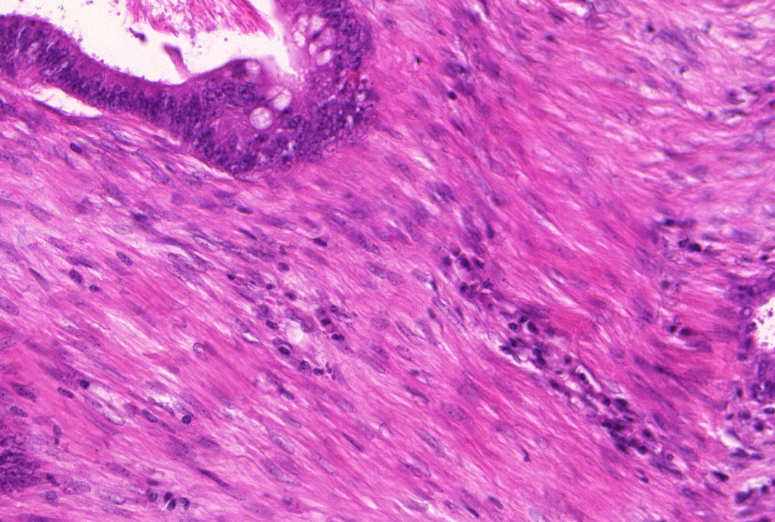} &
    \includegraphics[width=.12\textwidth]{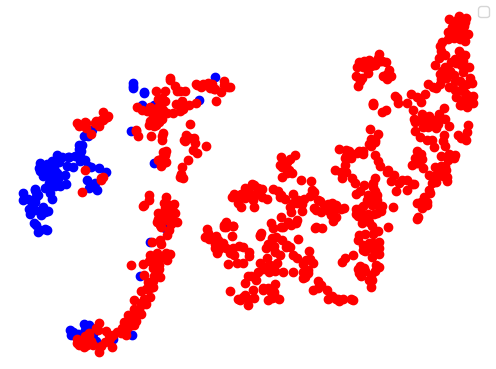} &
    \includegraphics[width=.12\textwidth]{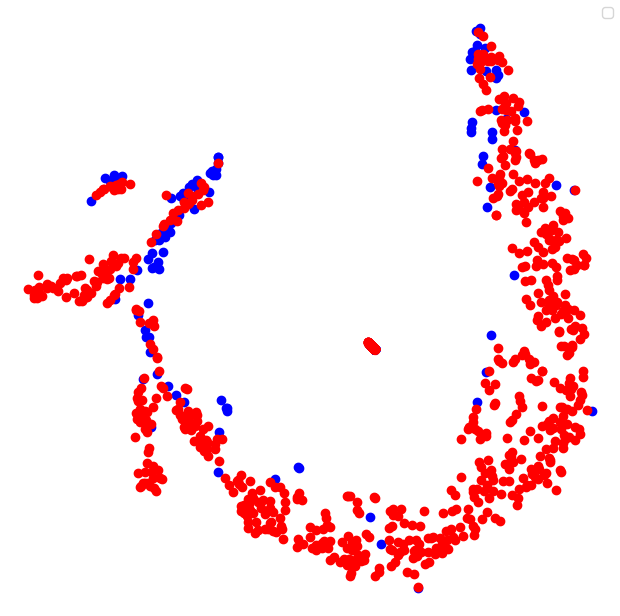} &
    \includegraphics[width=.12\textwidth]{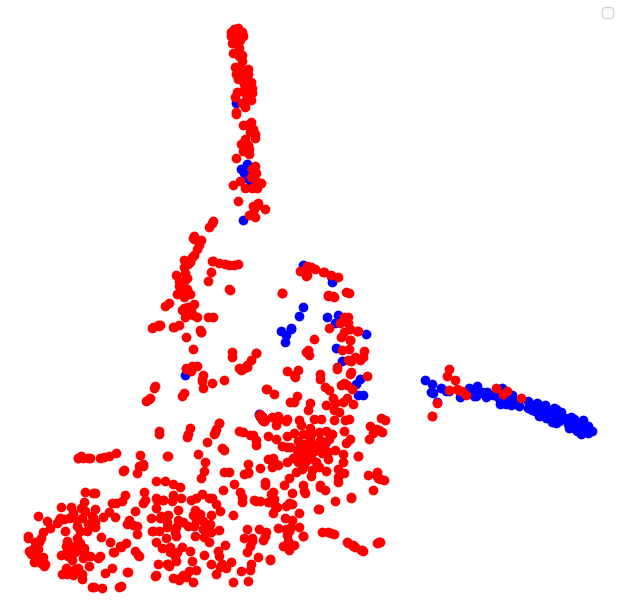} &
    \includegraphics[width=.12\textwidth]{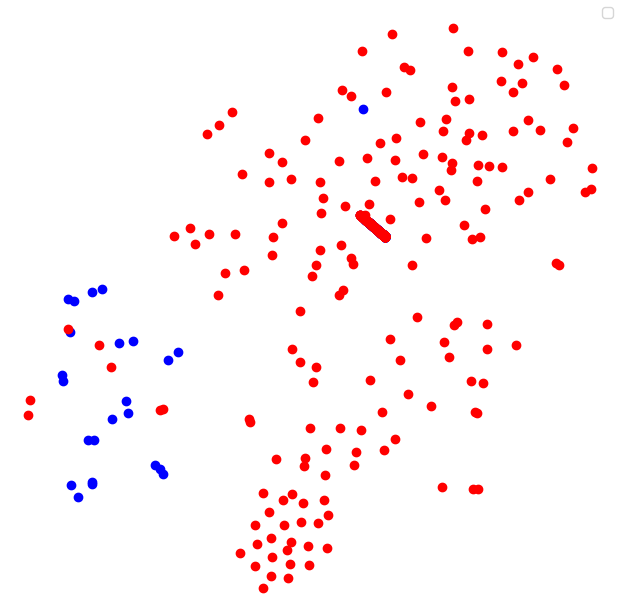} \\
     &
    \includegraphics[width=.12\textwidth]{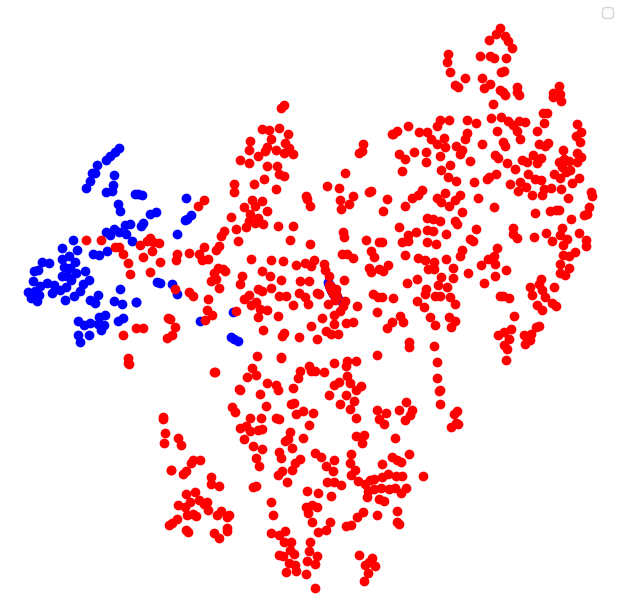} &
    \includegraphics[width=.12\textwidth]{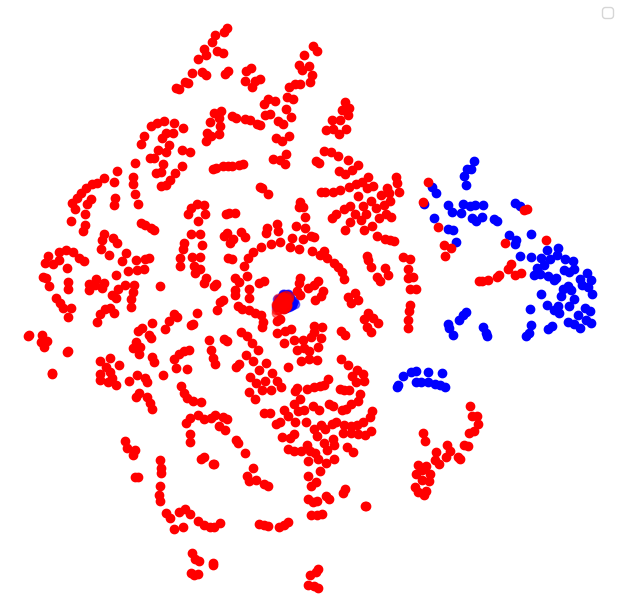} &
    \includegraphics[width=.12\textwidth]{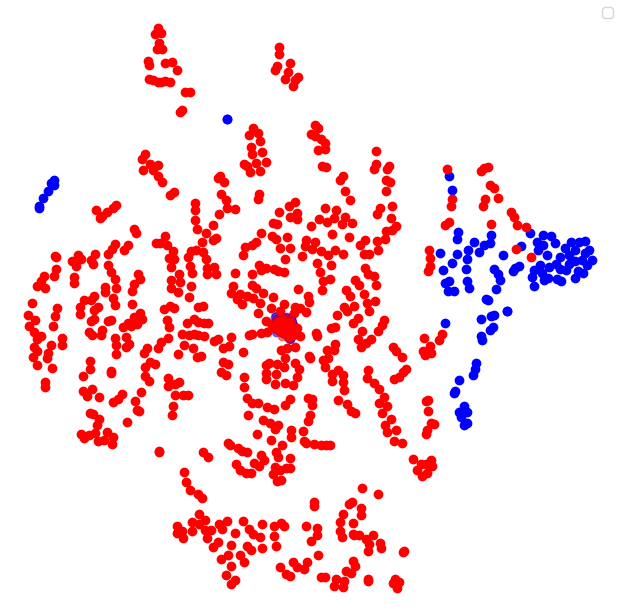} &
    \includegraphics[width=.12\textwidth]{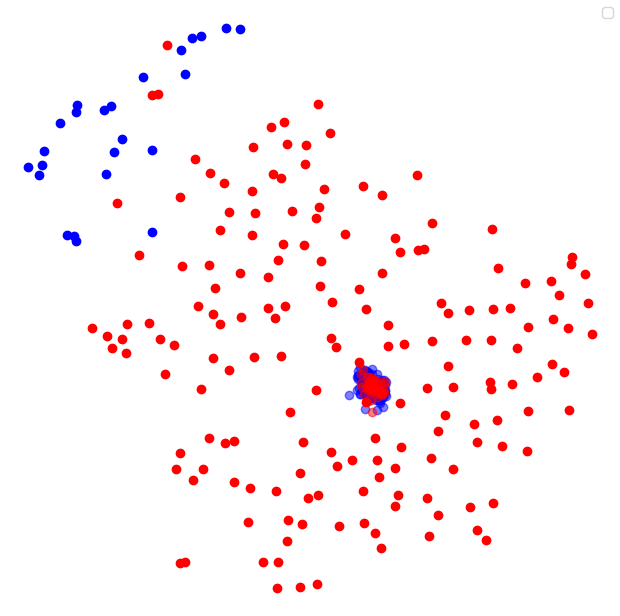} \\

     &
    \includegraphics[width=.12\textwidth]{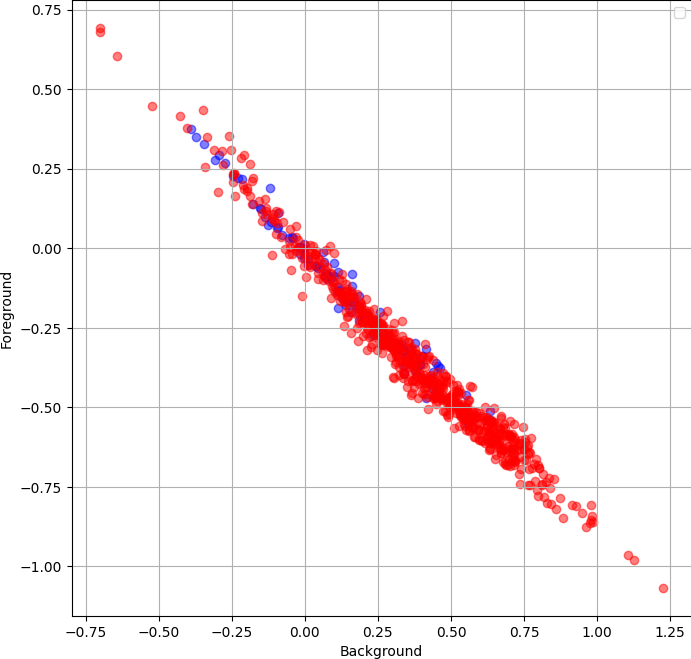} &
    \includegraphics[width=.12\textwidth]{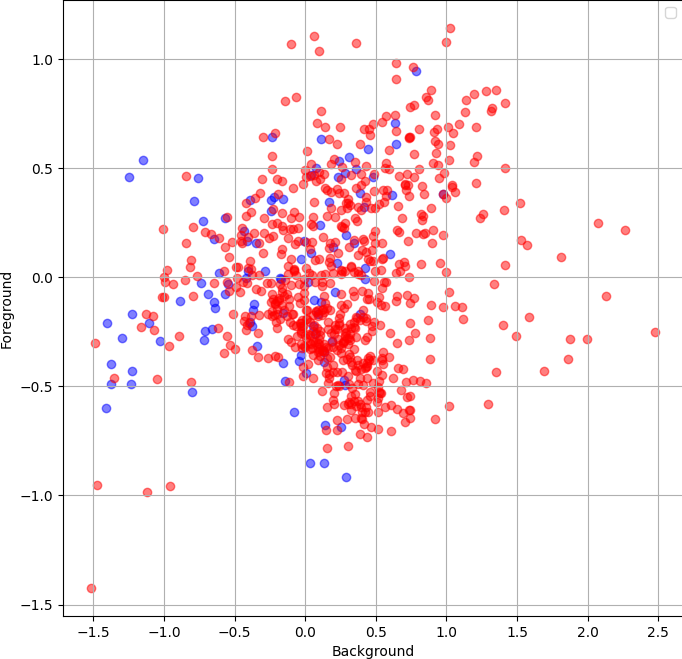} &
    \includegraphics[width=.12\textwidth]{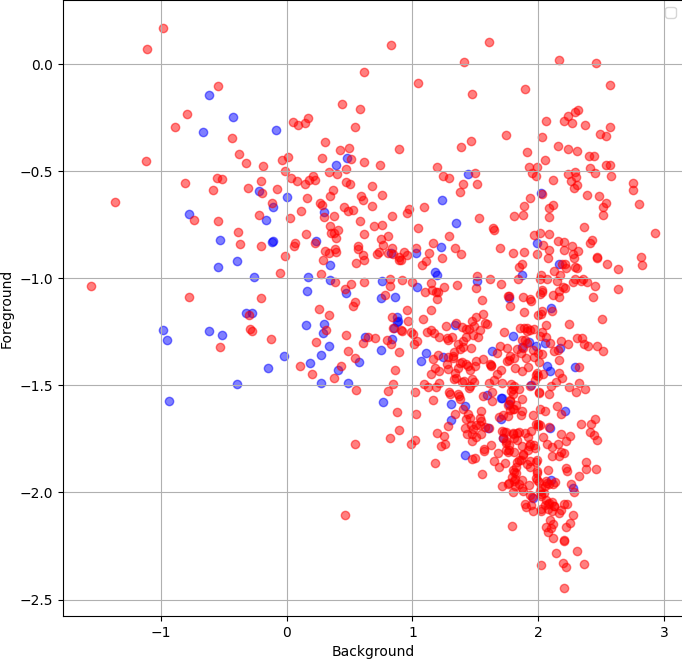} &
    \includegraphics[width=.12\textwidth]{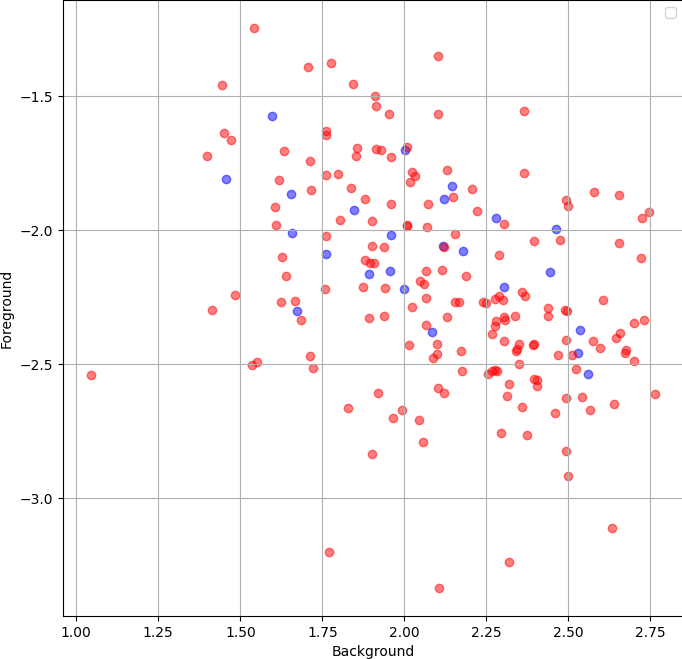} \\

     \includegraphics[width=.12\textwidth]{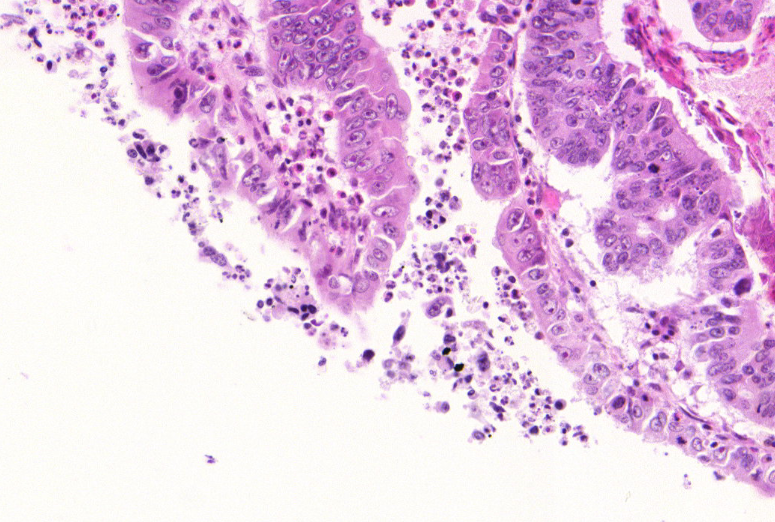} &
    \includegraphics[width=.12\textwidth]{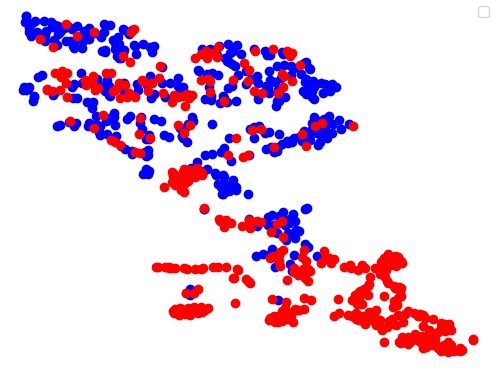} &
    \includegraphics[width=.12\textwidth]{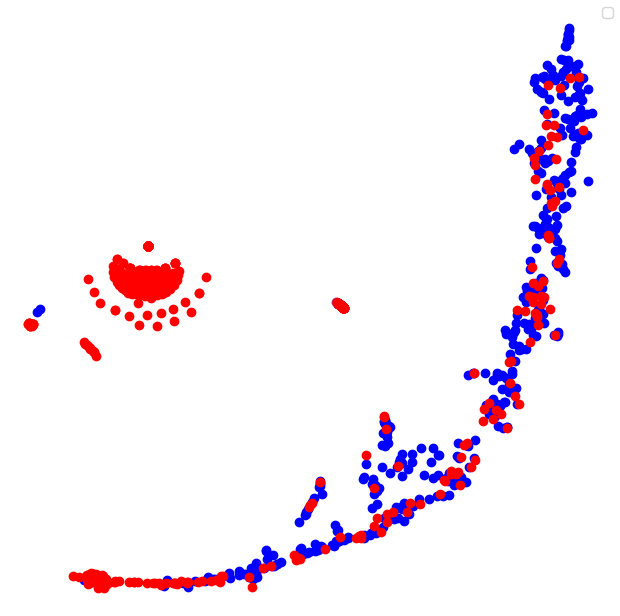} &
    \includegraphics[width=.12\textwidth]{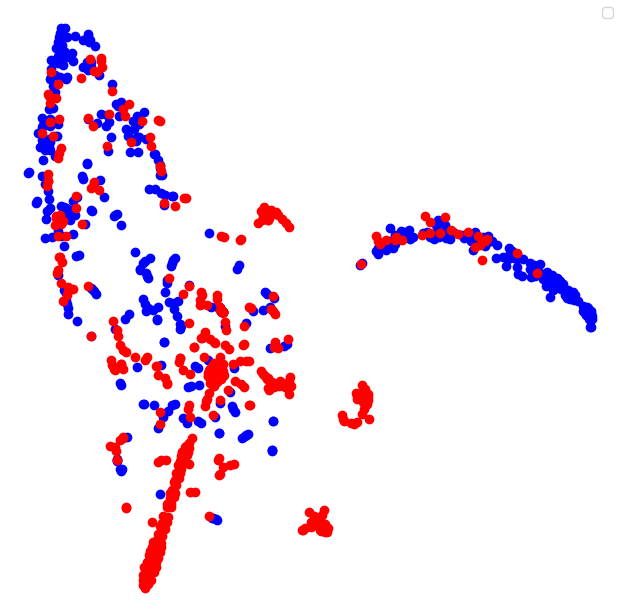} &
    \includegraphics[width=.12\textwidth]{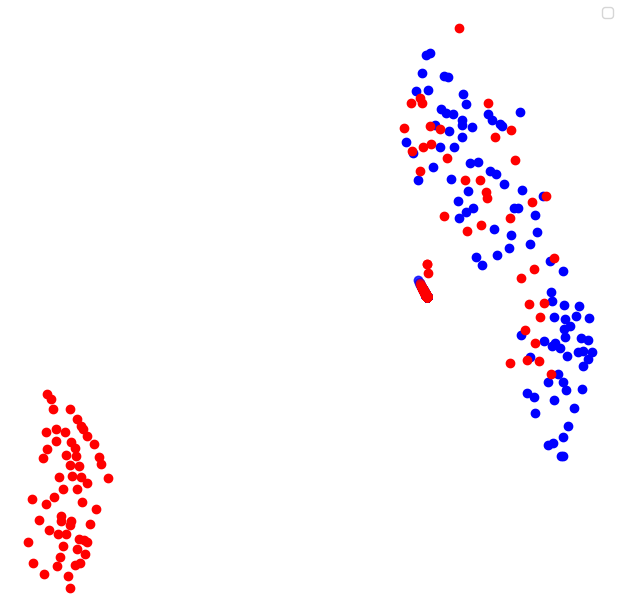} \\
     &
    \includegraphics[width=.12\textwidth]{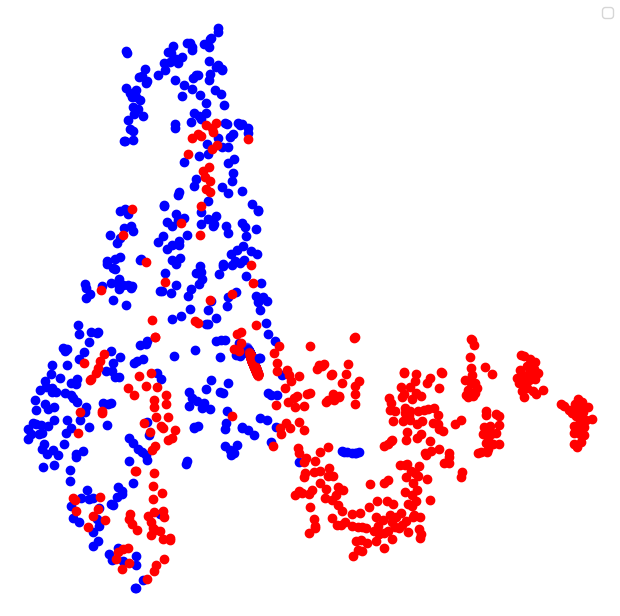} &
    \includegraphics[width=.12\textwidth]{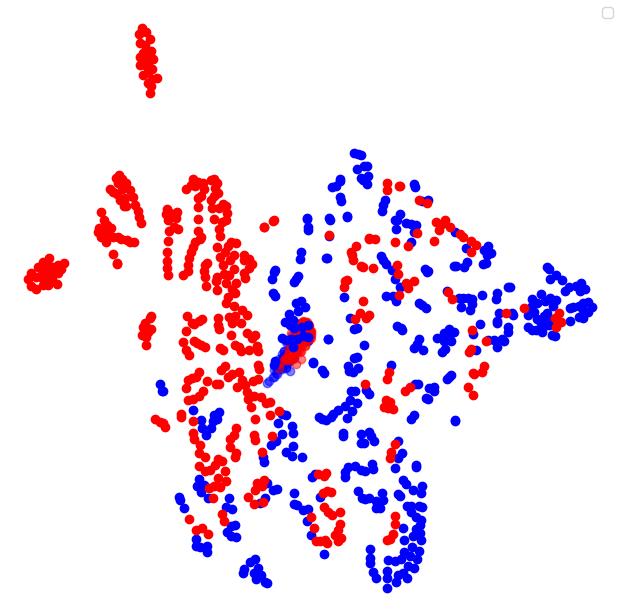} &
    \includegraphics[width=.12\textwidth]{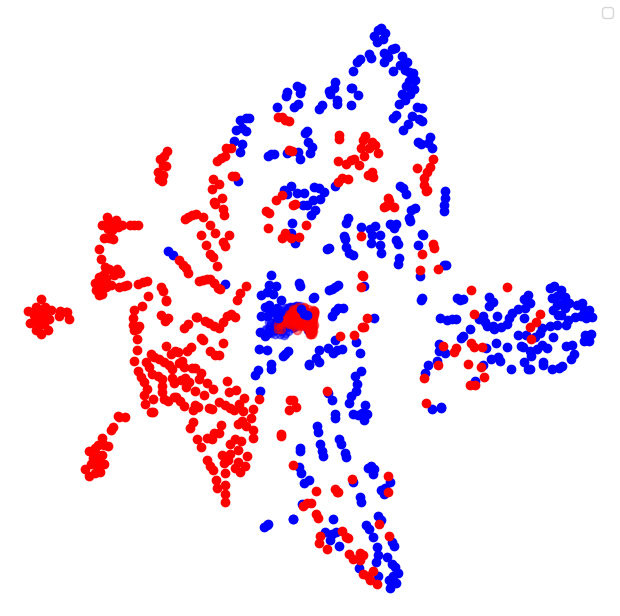} &
    \includegraphics[width=.12\textwidth]{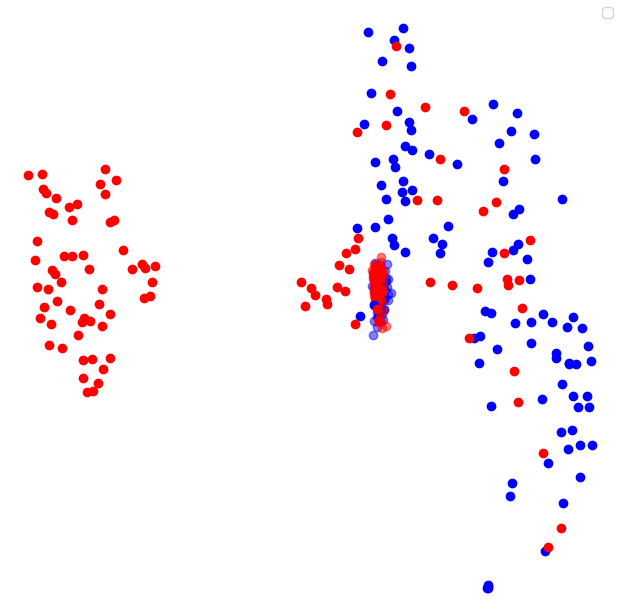} \\
     &
    \includegraphics[width=.12\textwidth]{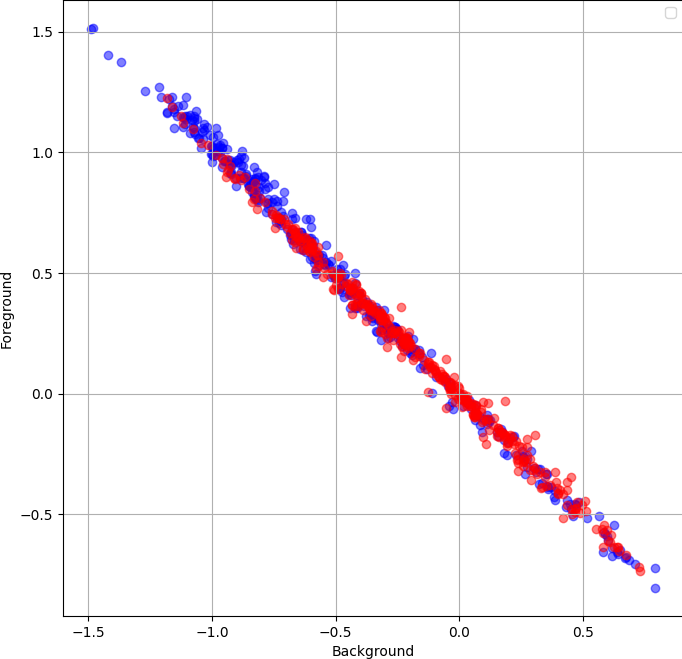} &
    \includegraphics[width=.12\textwidth]{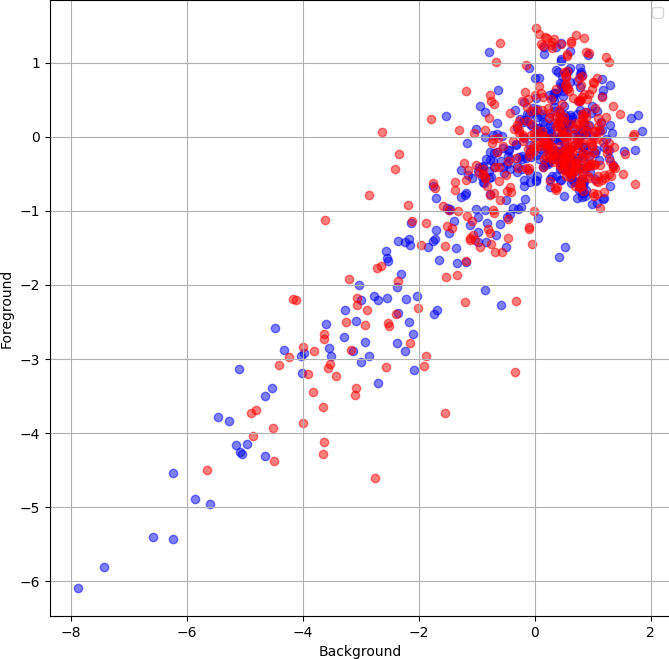} &
    \includegraphics[width=.12\textwidth]{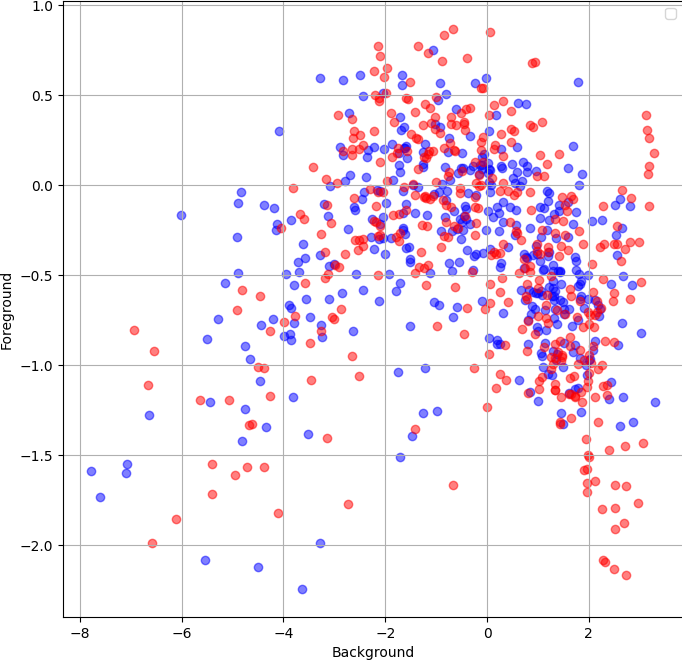} &
    \includegraphics[width=.12\textwidth]{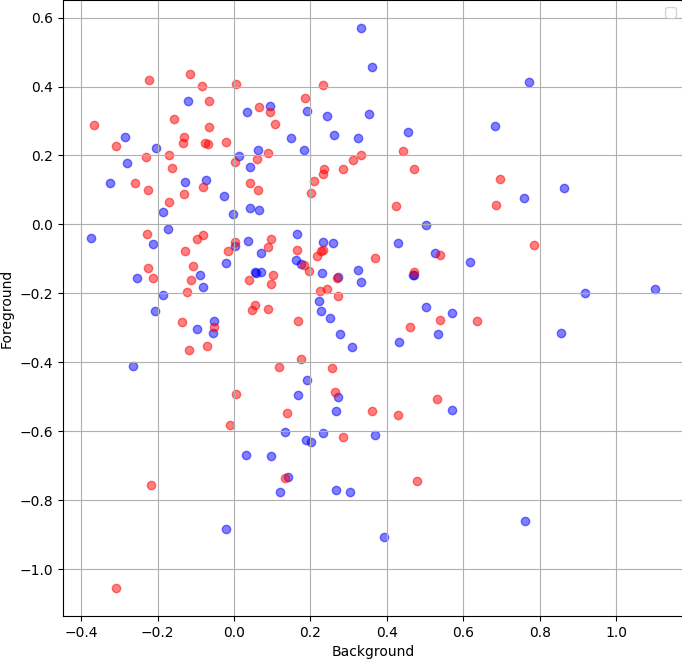} \\

    \includegraphics[width=.12\textwidth]{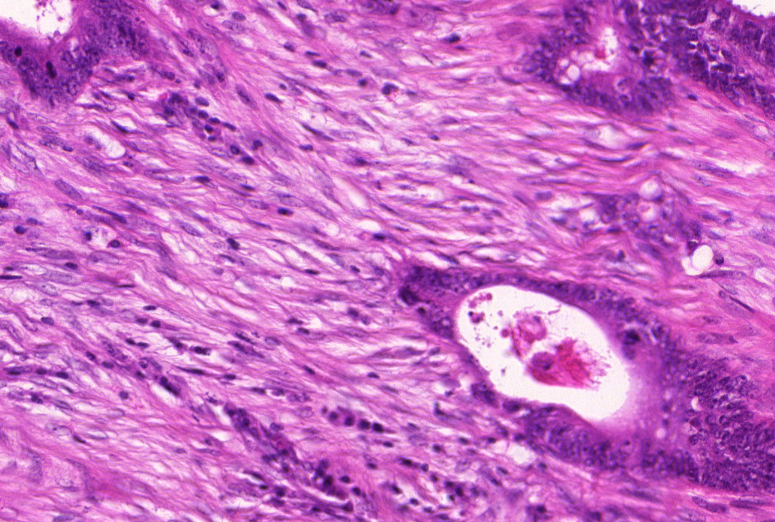} &
    \includegraphics[width=.12\textwidth]{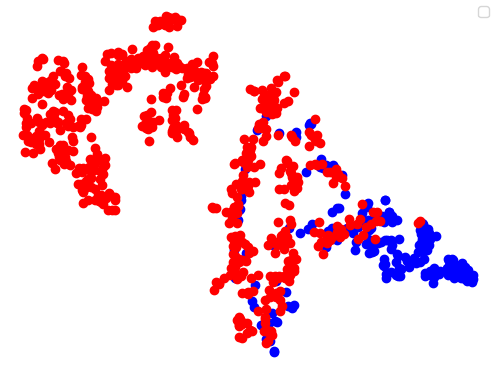} &
    \includegraphics[width=.12\textwidth]{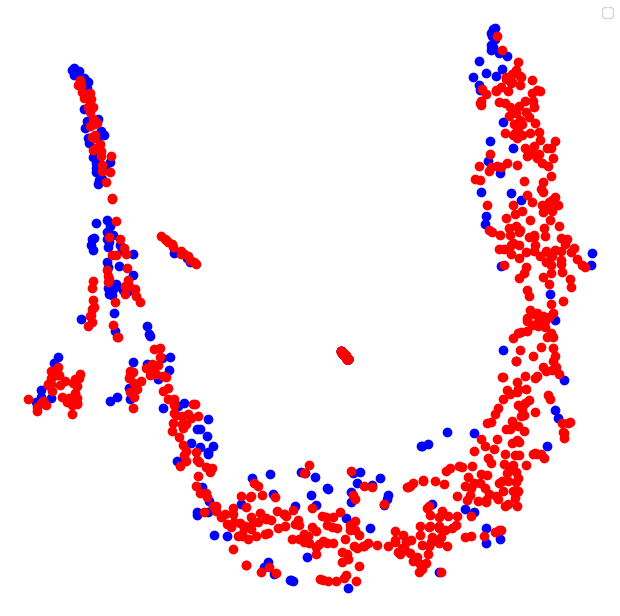} &
    \includegraphics[width=.12\textwidth]{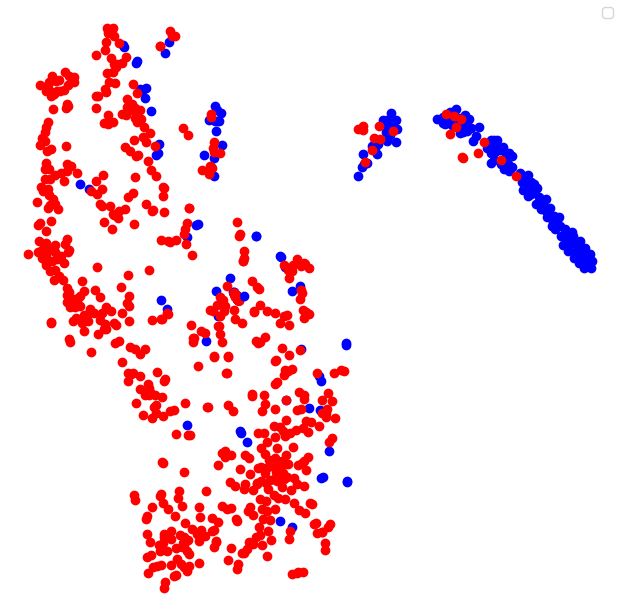} &
    \includegraphics[width=.12\textwidth]{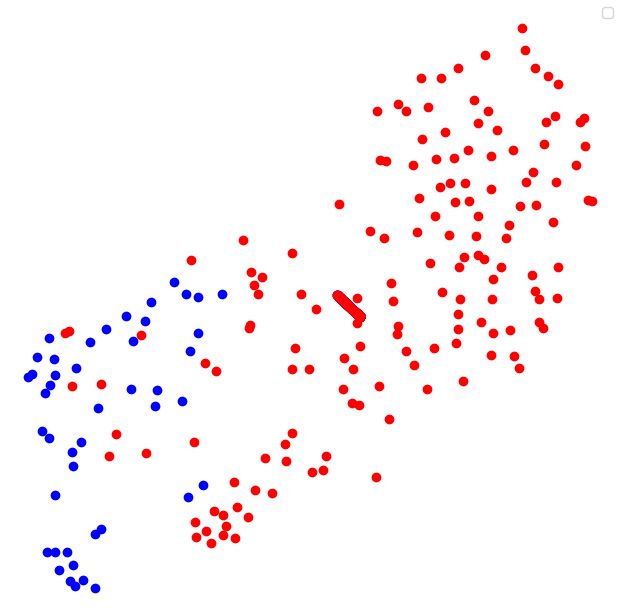} \\
    
    &
    \includegraphics[width=.12\textwidth]{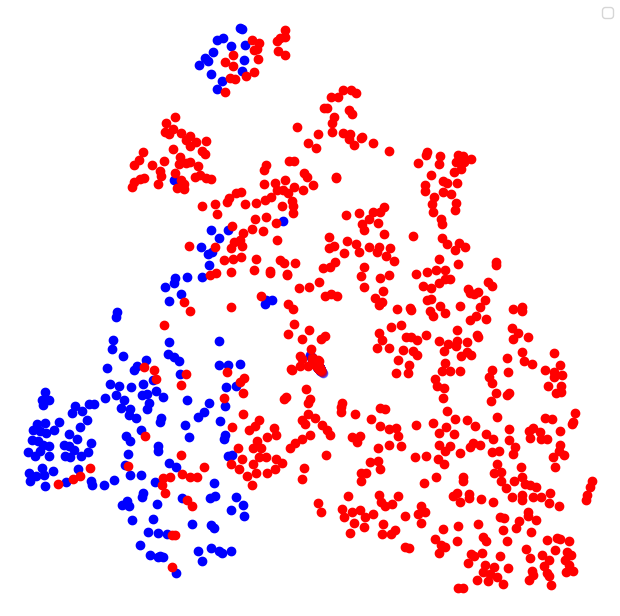} &
    \includegraphics[width=.12\textwidth]{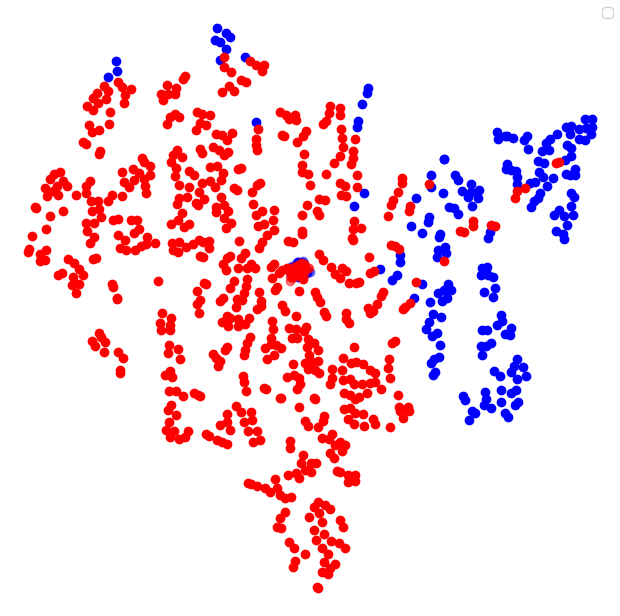} &
    \includegraphics[width=.12\textwidth]{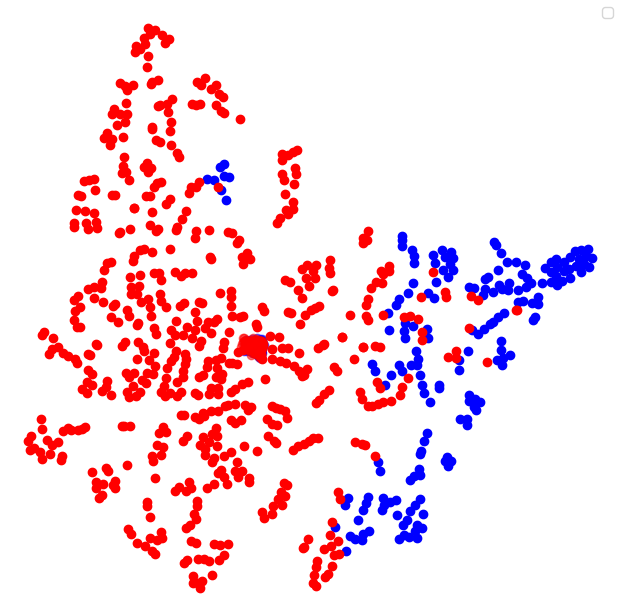} &
    \includegraphics[width=.12\textwidth]{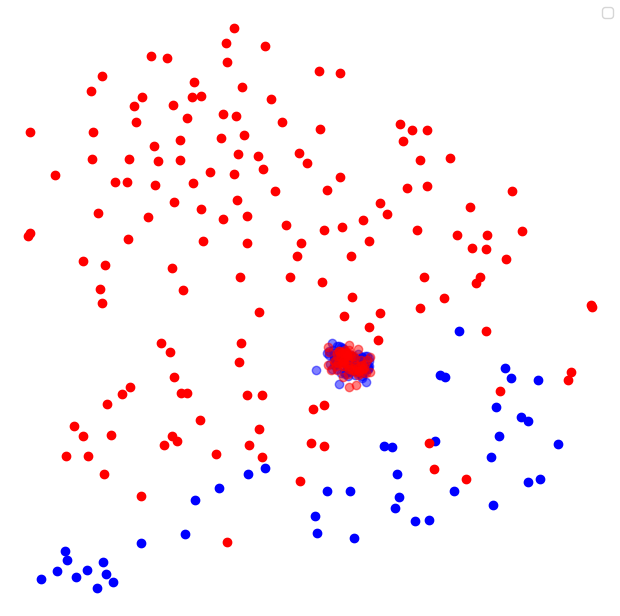} \\

     &
    \includegraphics[width=.12\textwidth]{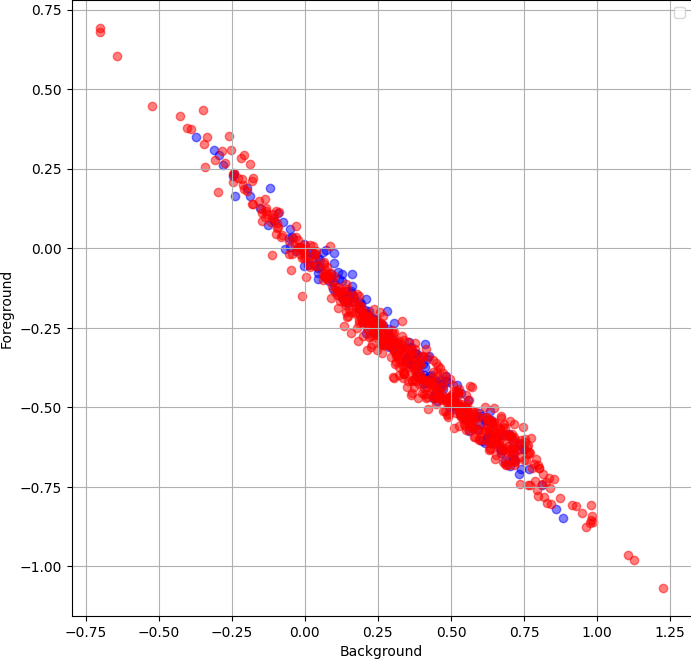} &
    \includegraphics[width=.12\textwidth]{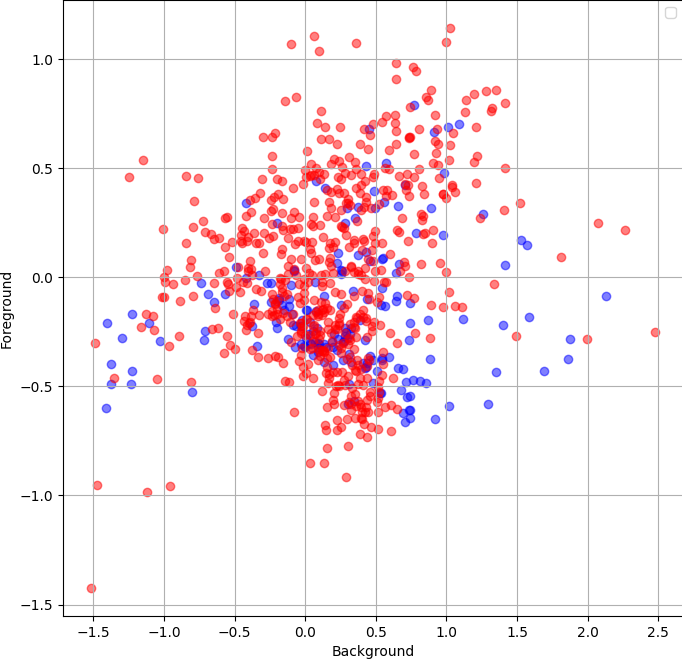} &
    \includegraphics[width=.12\textwidth]{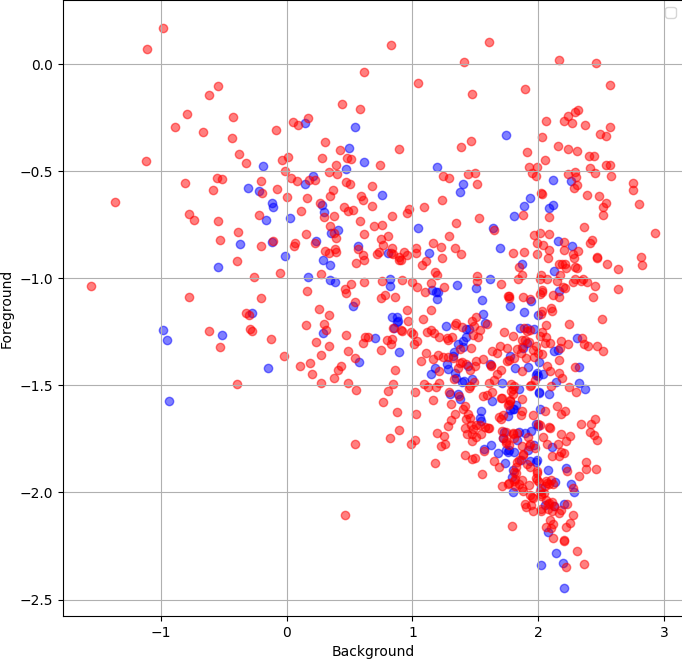} &
    \includegraphics[width=.12\textwidth]{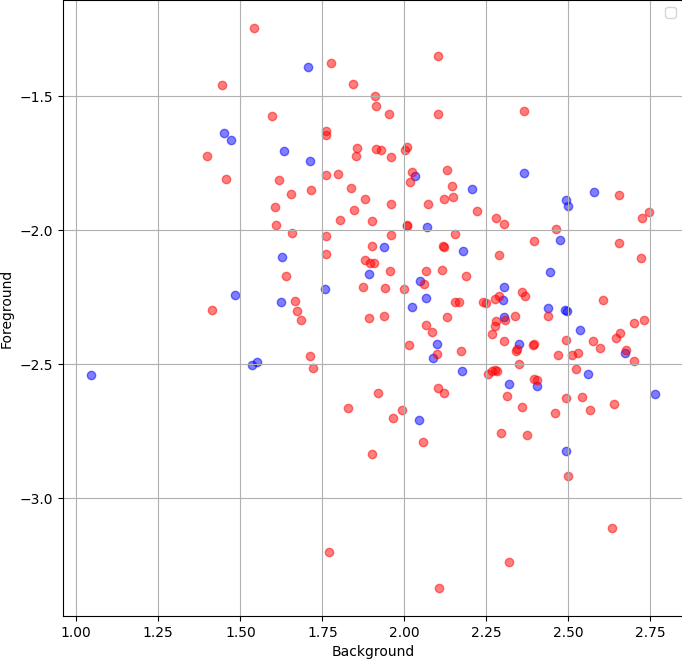} \\

  \end{tabular}
  \caption{
  OOD setup (\mbox{\camsixteen $\rightarrow$ \glas}): For the three \textbf{cancerous} images from \glas, we display the t-SNE projection of foreground and background pixel-features of WSOL baseline methods without (1st row) and with (2nd row) \pixelcam. The 3rd row presents the logits of the pixel classifier of \pixelcam.
  }
  \label{fig:feature-tsne-target-cancer-glas}
\end{figure*}

\begin{figure*}[!ht]
\centering
  \begin{tabular}{@{}c@{\hskip 10pt}c@{\hskip 10pt}c@{\hskip 10pt}c@{\hskip 10pt}c@{}}
    \makecell{Input} & \makecell{DeepMil} & \makecell{GradCAM++} & \makecell{LayerCAM} & \makecell{SAT} \\
   \includegraphics[width=.12\textwidth]{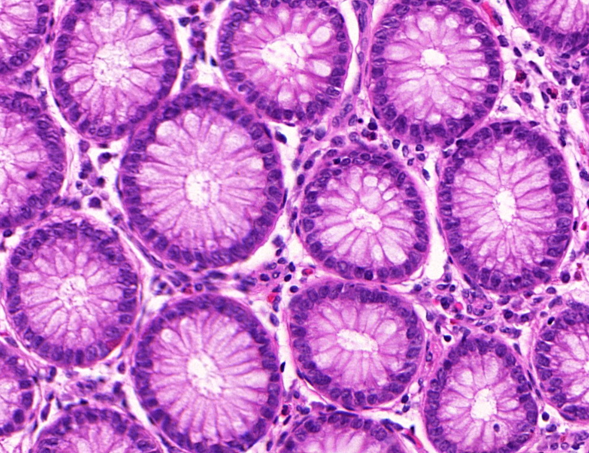} &
    \includegraphics[width=.12\textwidth]{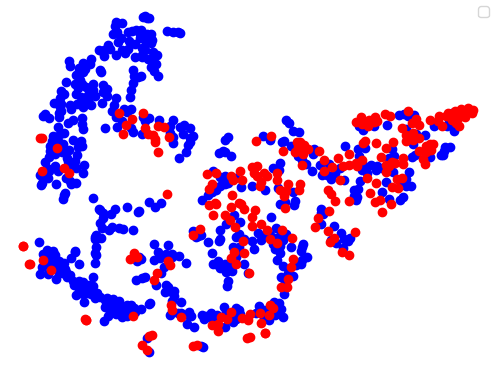} &
    \includegraphics[width=.12\textwidth]{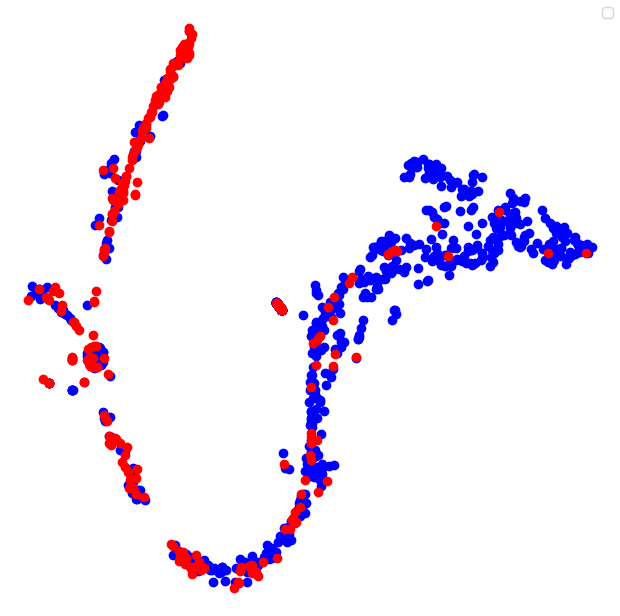} &
    \includegraphics[width=.12\textwidth]{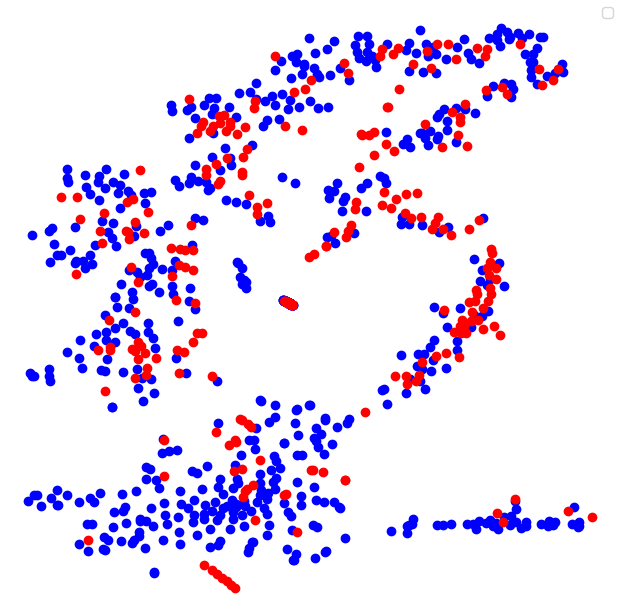} &
    \includegraphics[width=.12\textwidth]{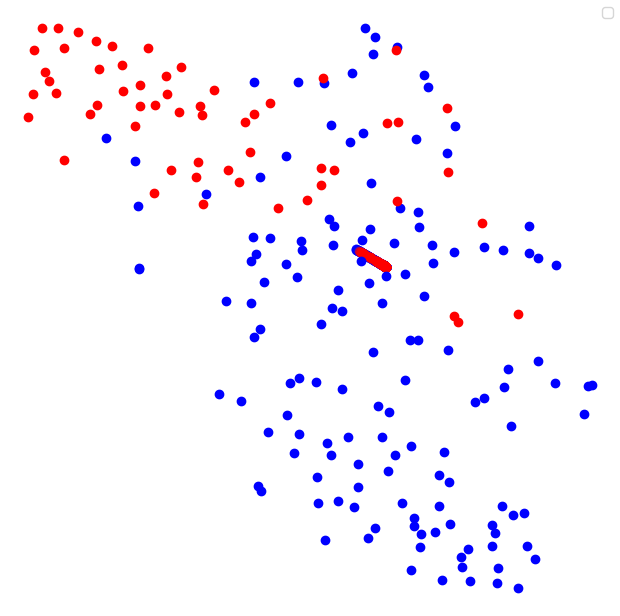} \\
     &
    \includegraphics[width=.12\textwidth]{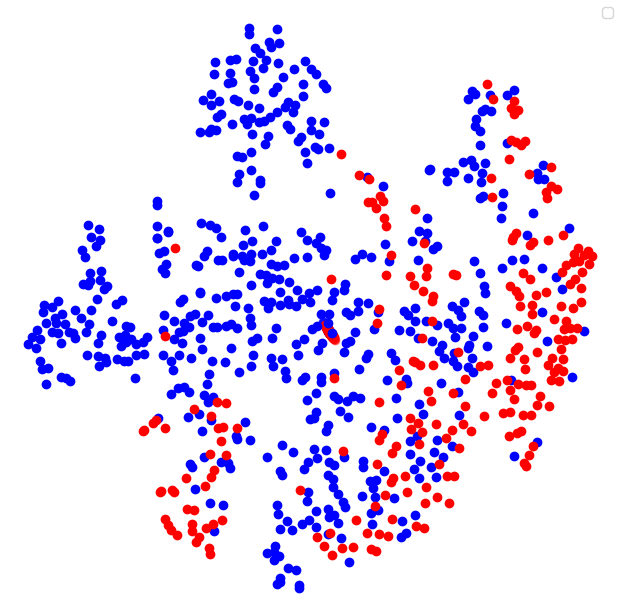} &
    \includegraphics[width=.12\textwidth]{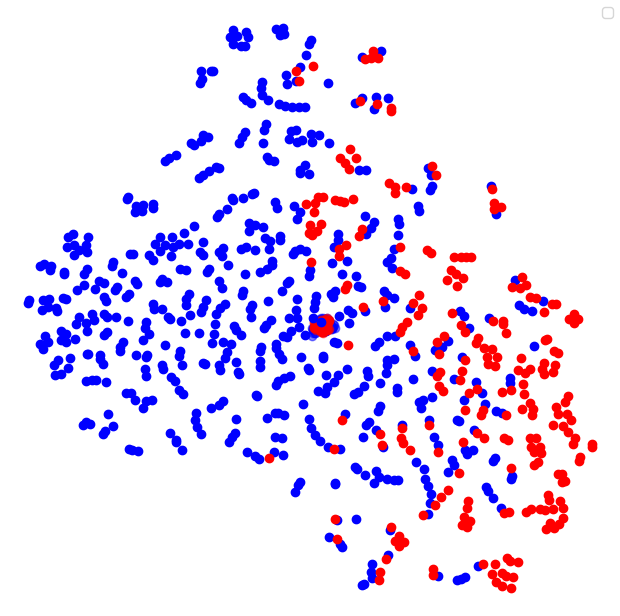} &
    \includegraphics[width=.12\textwidth]{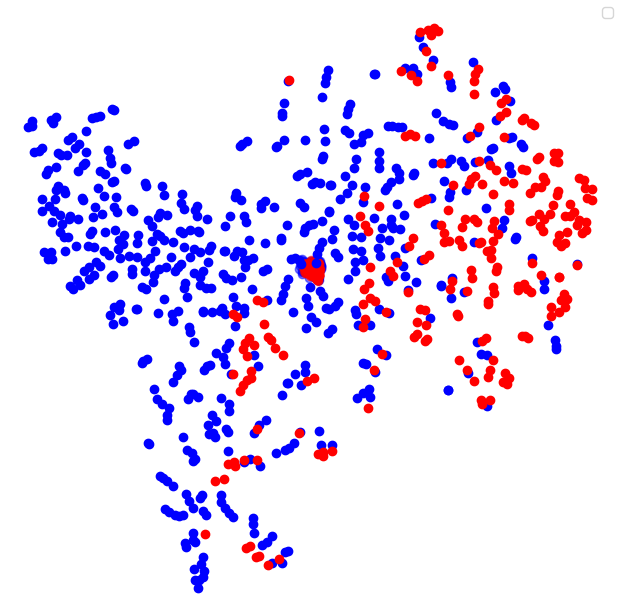} &
    \includegraphics[width=.12\textwidth]{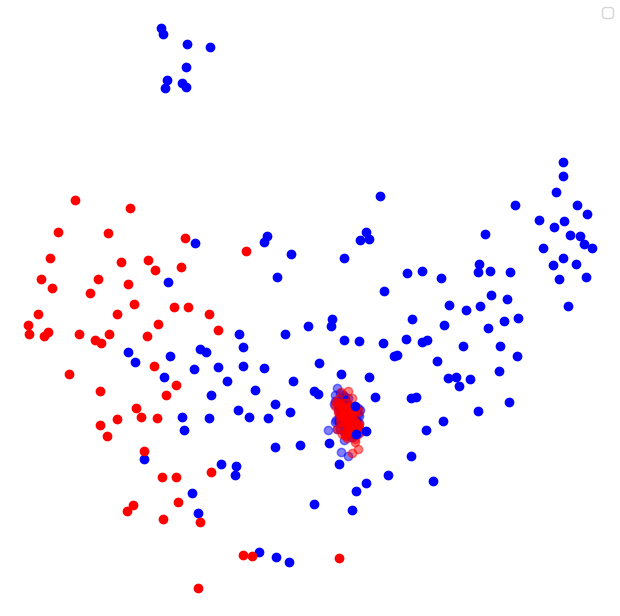} \\

     &
    \includegraphics[width=.12\textwidth]{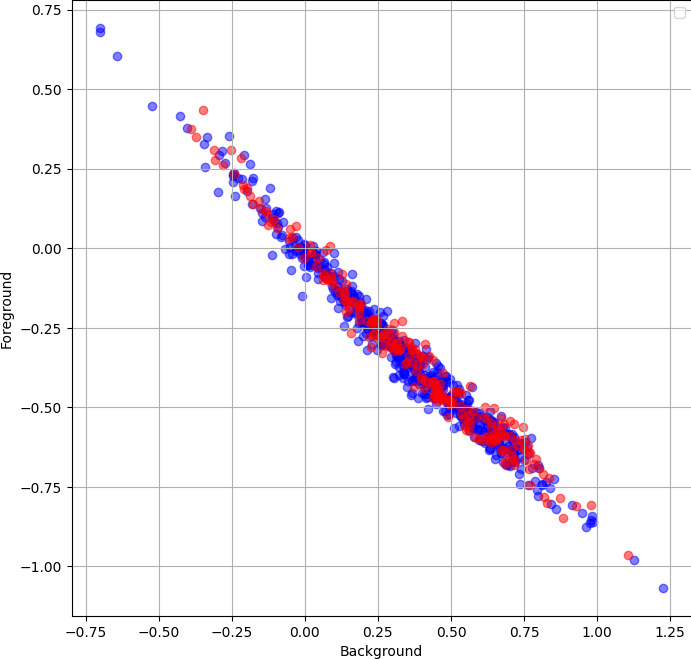} &
    \includegraphics[width=.12\textwidth]{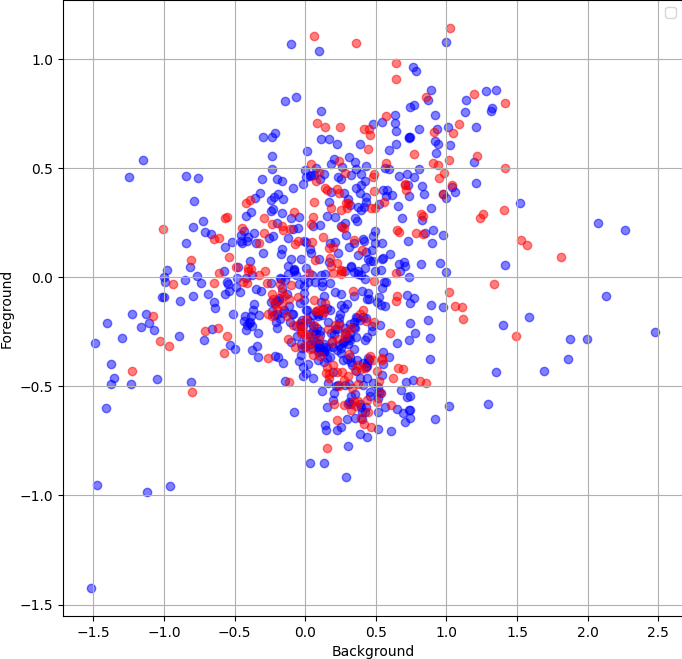} &
    \includegraphics[width=.12\textwidth]{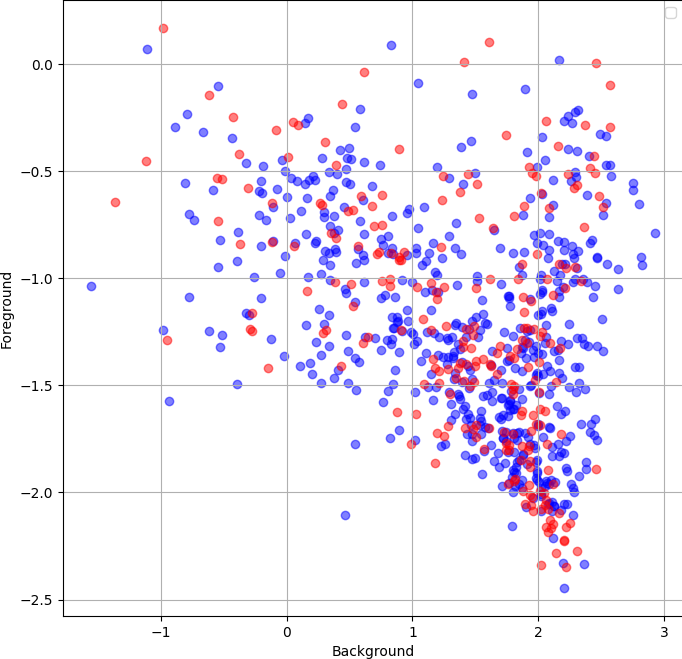} &
    \includegraphics[width=.12\textwidth]{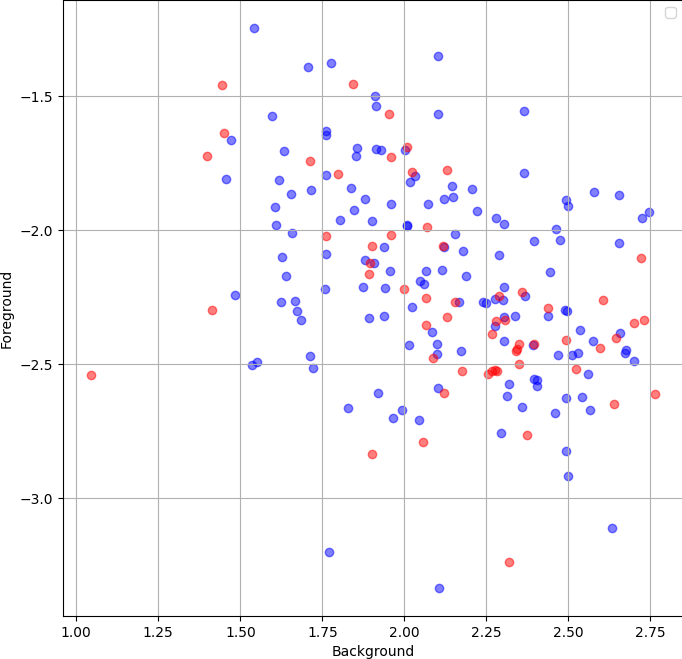} \\

     \includegraphics[width=.12\textwidth]{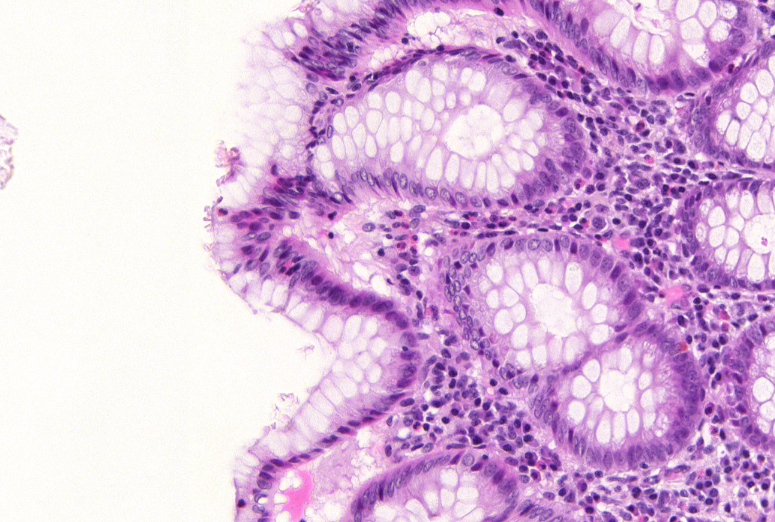} &
    \includegraphics[width=.12\textwidth]{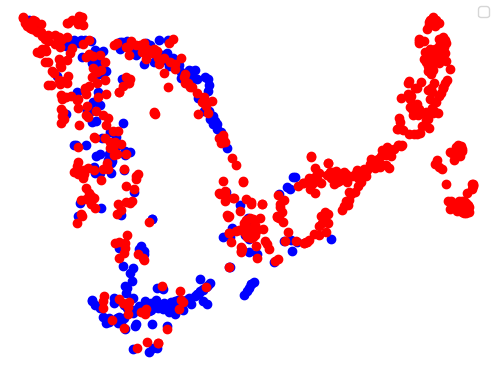} &
    \includegraphics[width=.12\textwidth]{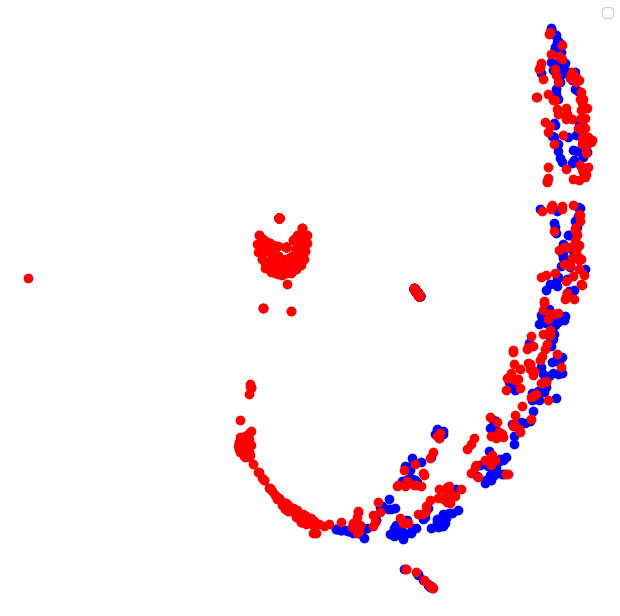} &
    \includegraphics[width=.12\textwidth]{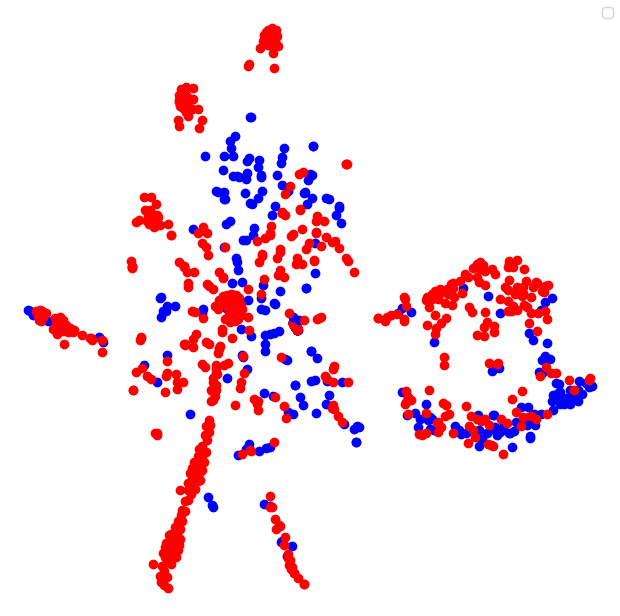} &
    \includegraphics[width=.12\textwidth]{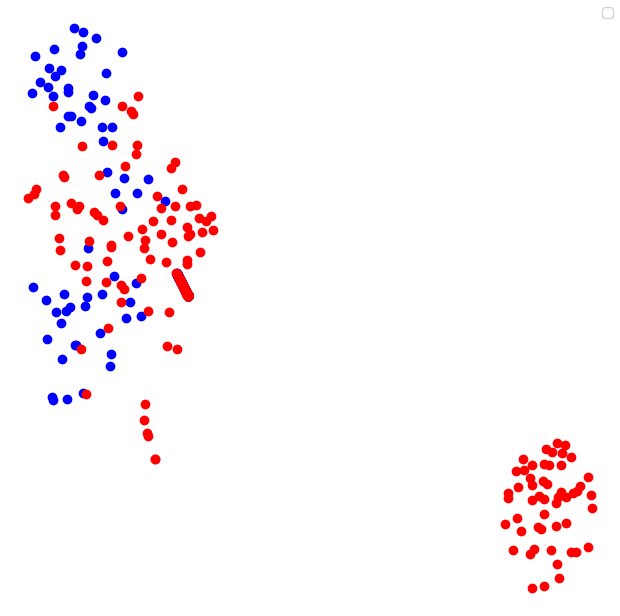} \\
     &
    \includegraphics[width=.12\textwidth]{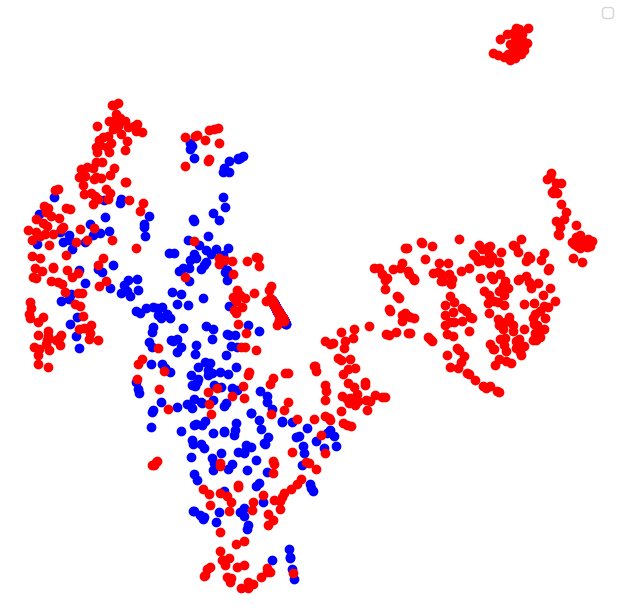} &
    \includegraphics[width=.12\textwidth]{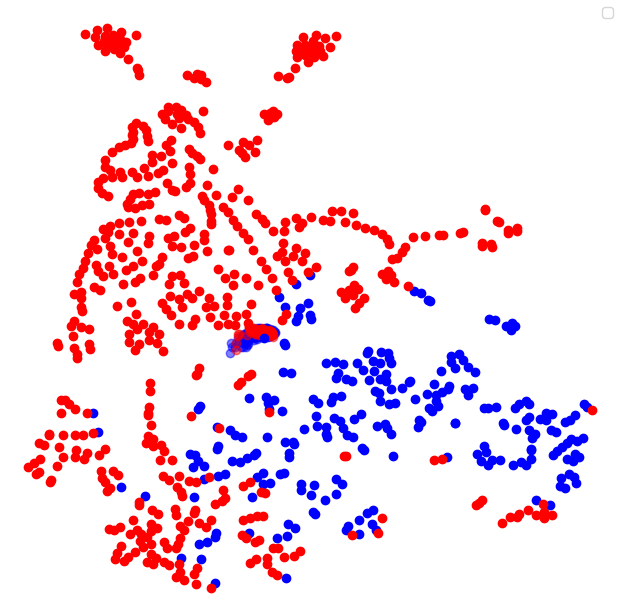} &
    \includegraphics[width=.12\textwidth]{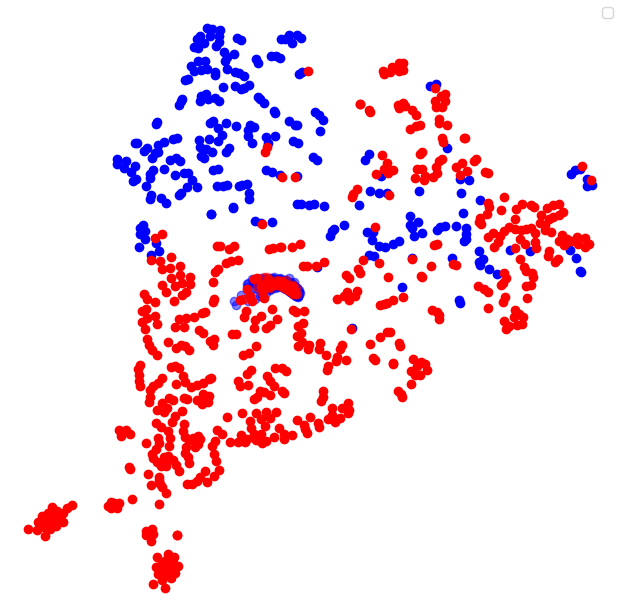} &
    \includegraphics[width=.12\textwidth]{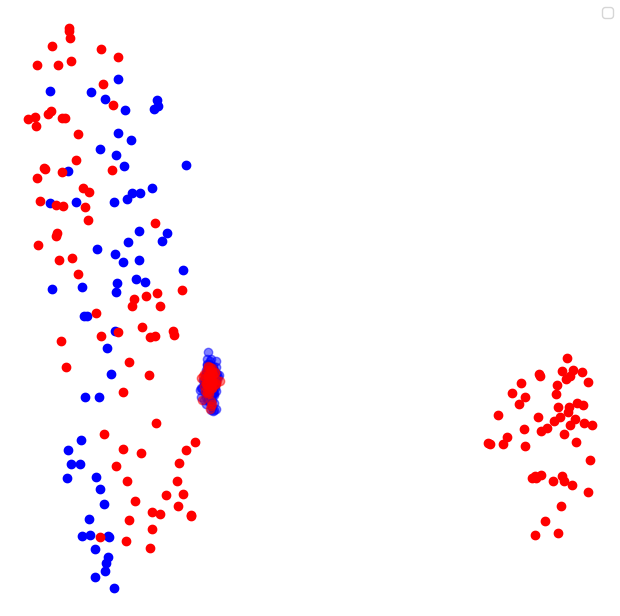} \\
     &
    \includegraphics[width=.12\textwidth]{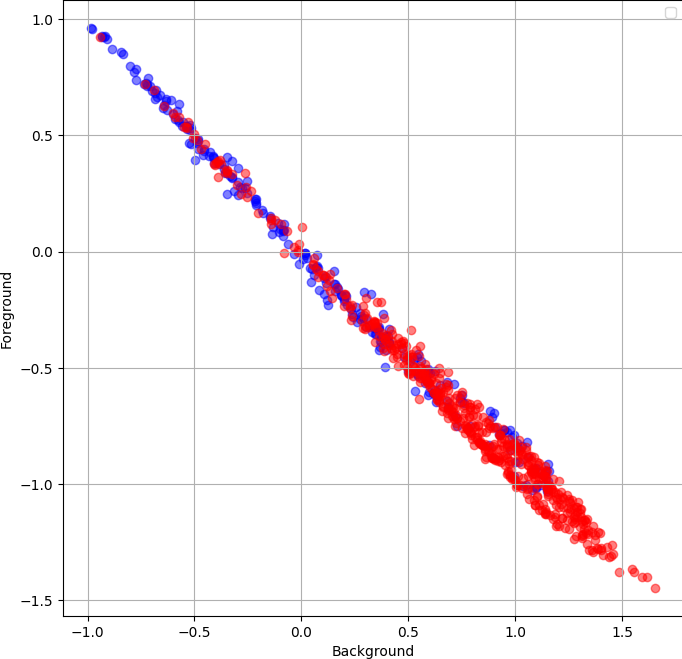} &
    \includegraphics[width=.12\textwidth]{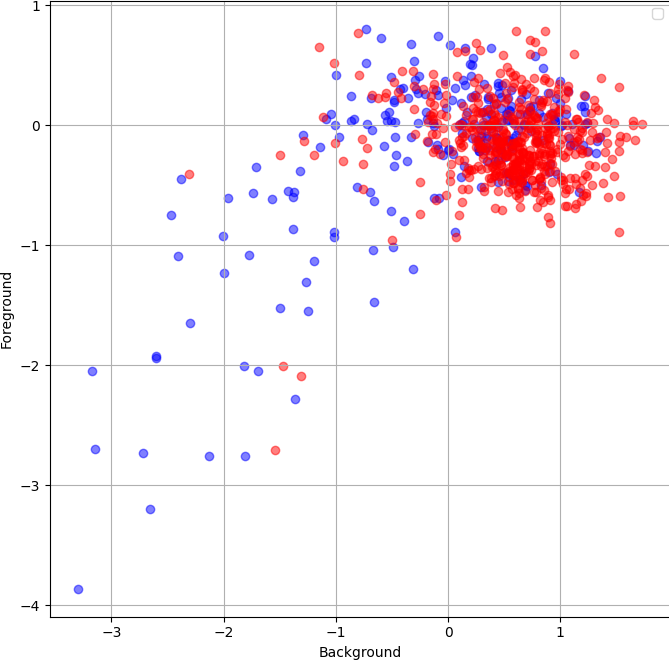} &
    \includegraphics[width=.12\textwidth]{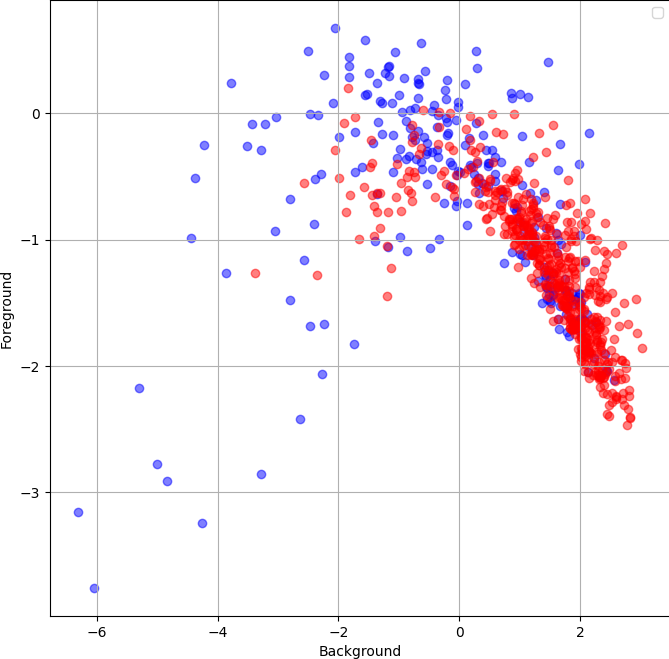} &
    \includegraphics[width=.12\textwidth]{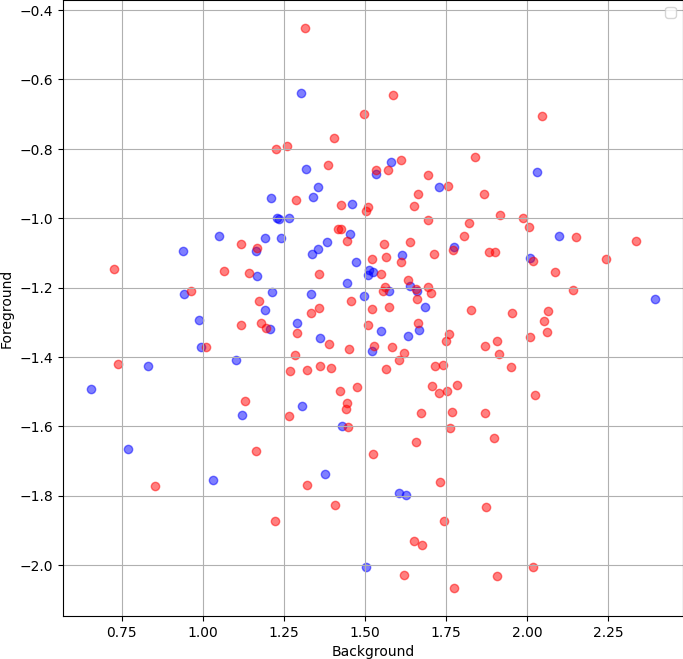} \\

    \includegraphics[width=.12\textwidth]{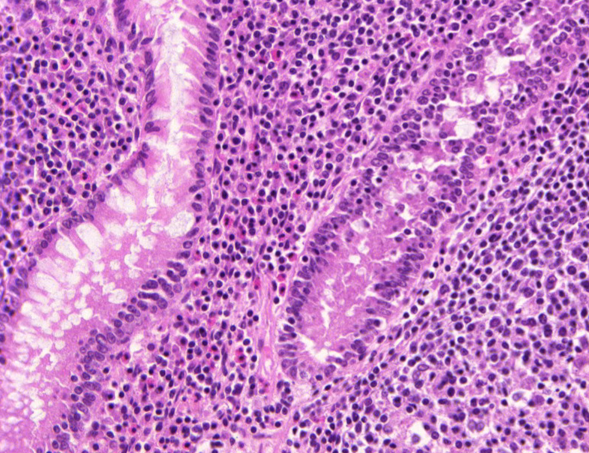} &
    \includegraphics[width=.12\textwidth]{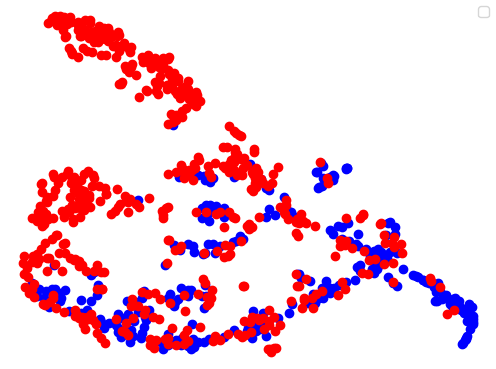} &
    \includegraphics[width=.12\textwidth]{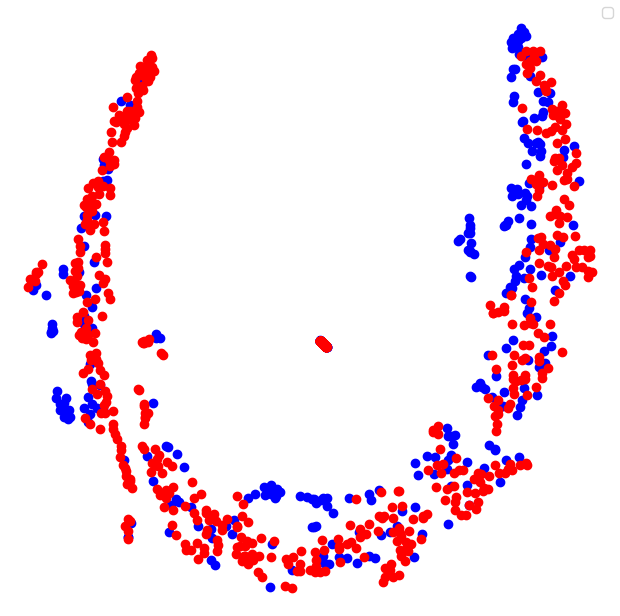} &
    \includegraphics[width=.12\textwidth]{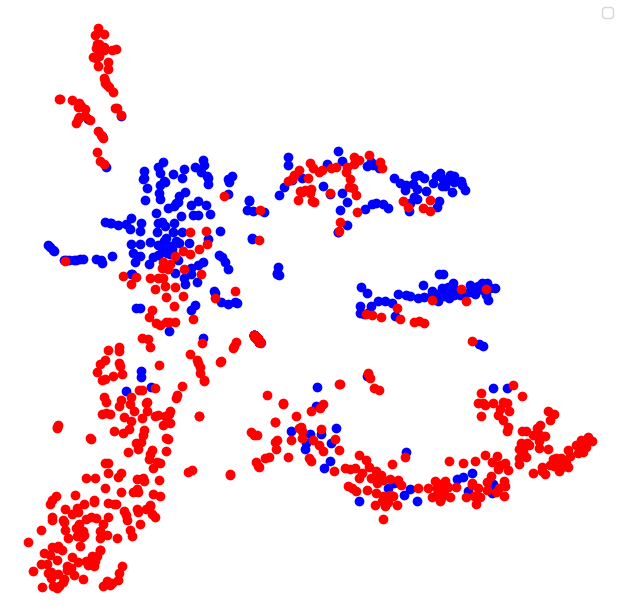} &
    \includegraphics[width=.12\textwidth]{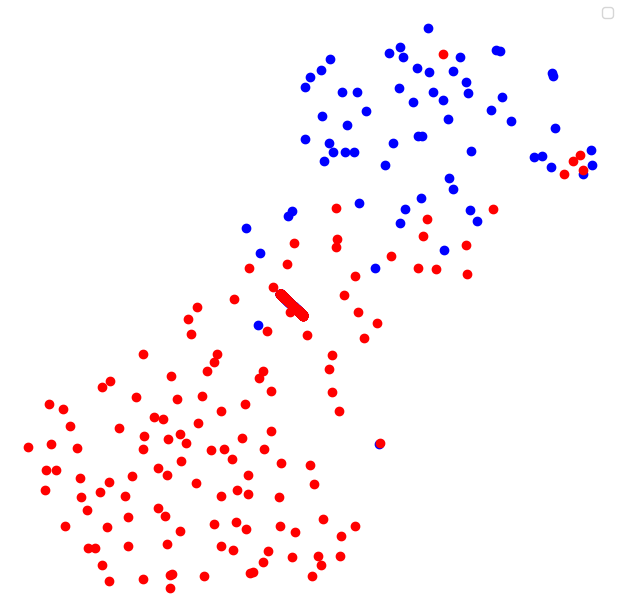} \\
    
    &
    \includegraphics[width=.12\textwidth]{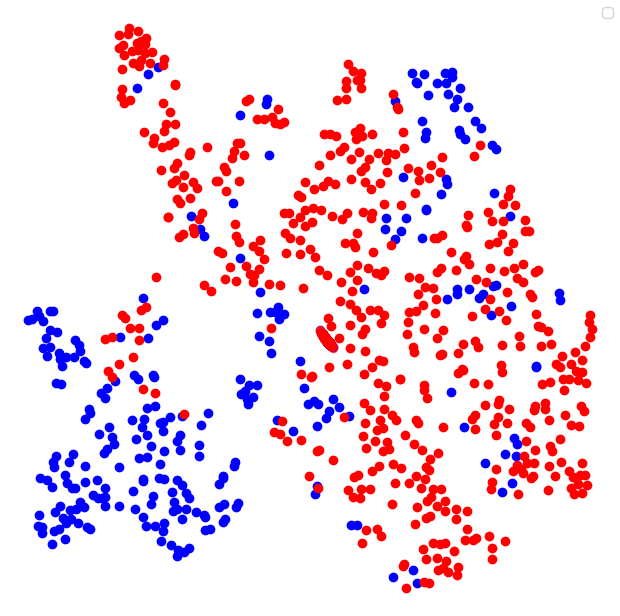} &
    \includegraphics[width=.12\textwidth]{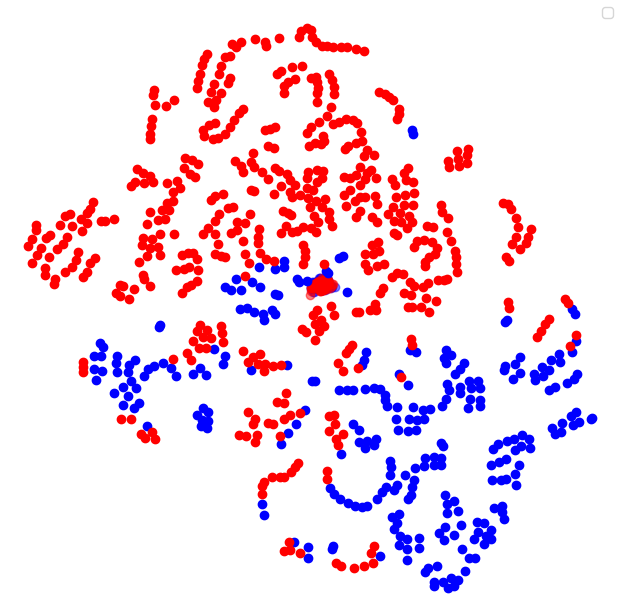} &
    \includegraphics[width=.12\textwidth]{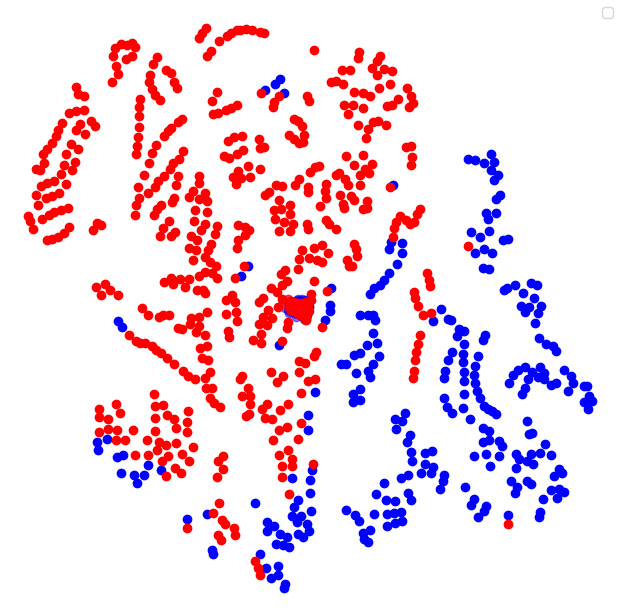} &
    \includegraphics[width=.12\textwidth]{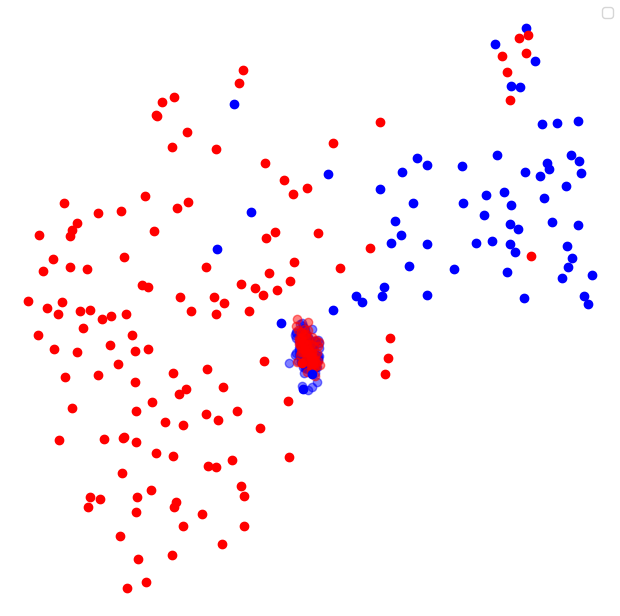} \\

     &
    \includegraphics[width=.12\textwidth]{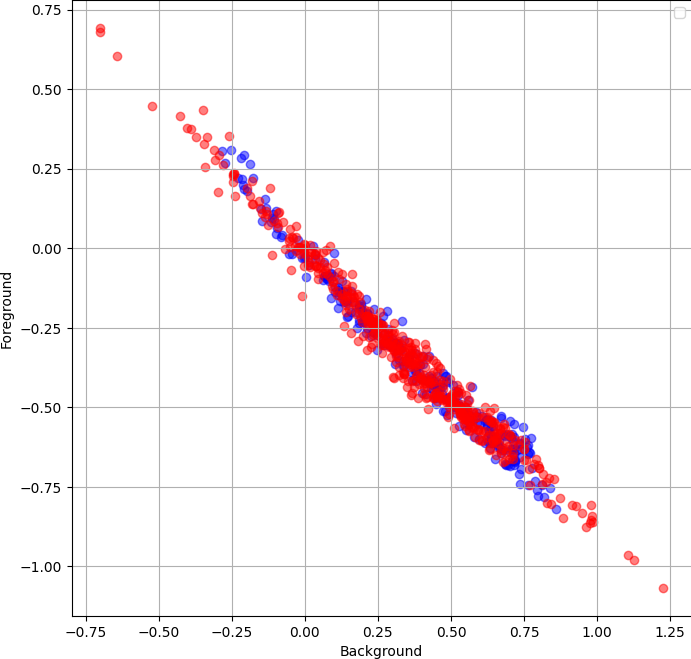} &
    \includegraphics[width=.12\textwidth]{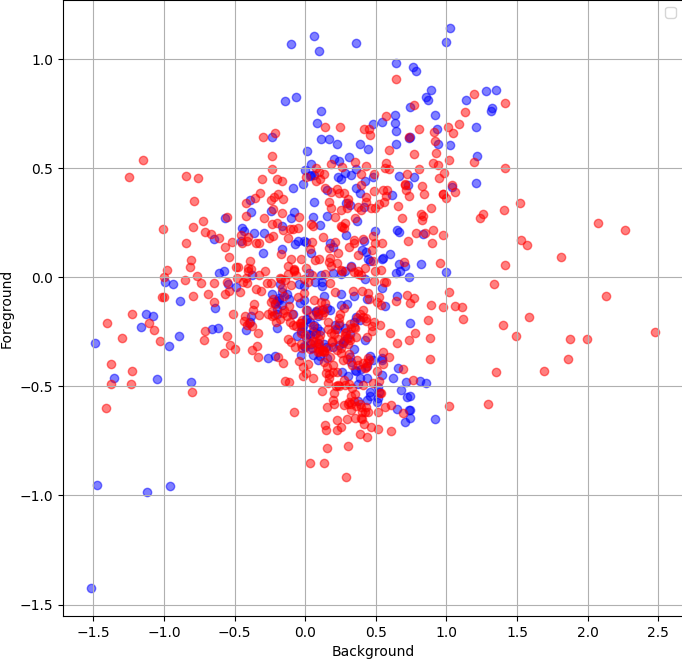} &
    \includegraphics[width=.12\textwidth]{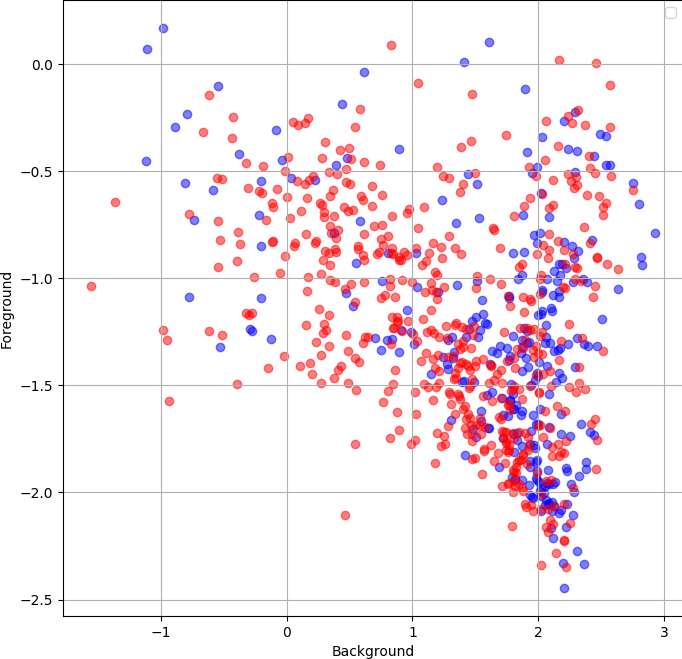} &
    \includegraphics[width=.12\textwidth]{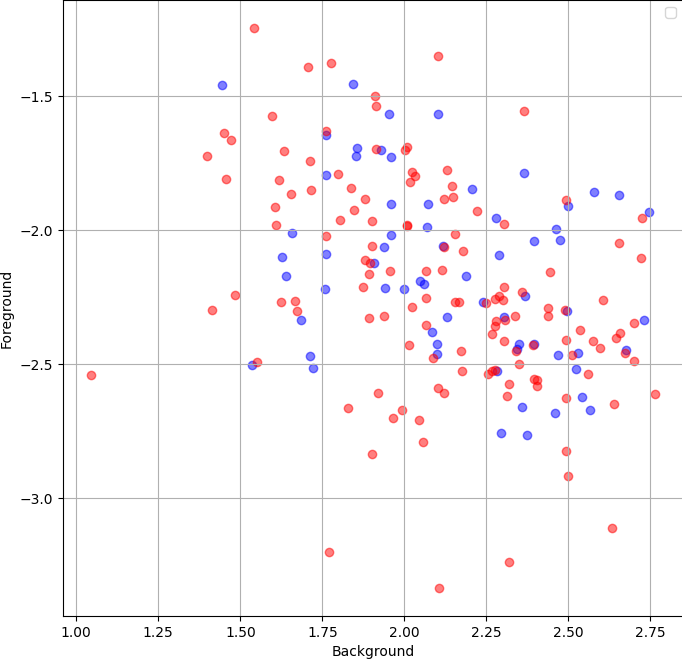} \\
       
  \end{tabular}
  \caption{
  OOD setup (\mbox{\camsixteen $\rightarrow$ \glas}): For the three \textbf{normal} images from \glas, we display the t-SNE projection of foreground and background pixel-features of WSOL baseline methods without (1st row) and with (2nd row) \pixelcam. The 3rd row presents the logits of the pixel classifier of \pixelcam.
  }
  \label{fig:feature-tsne-target-normal-glas}
\end{figure*}

\begin{figure*}[!ht]
\centering
  \begin{tabular}{@{}c@{\hskip 10pt}c@{\hskip 10pt}c@{\hskip 10pt}c@{\hskip 10pt}c@{}}
    \makecell{Input} & \makecell{DeepMil} & \makecell{GradCAM++} & \makecell{LayerCAM} & \makecell{SAT} \\
   \includegraphics[width=.12\textwidth]{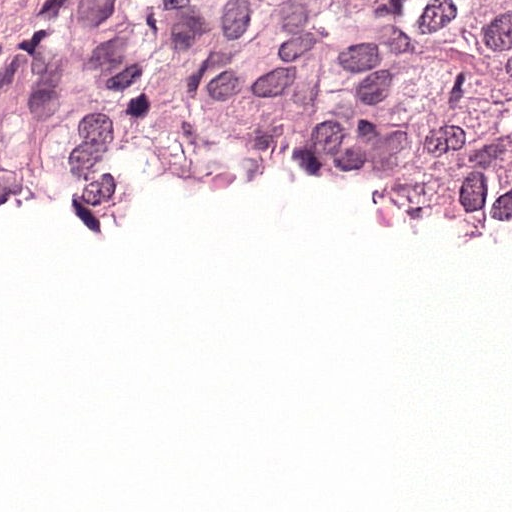} &
    \includegraphics[width=.12\textwidth]{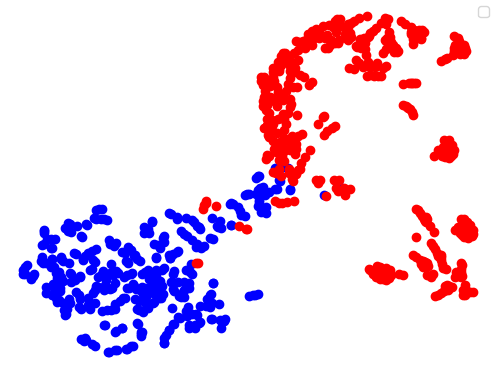} &
    \includegraphics[width=.12\textwidth]{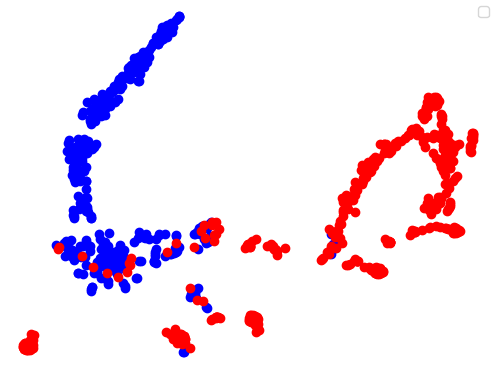} &
    \includegraphics[width=.12\textwidth]{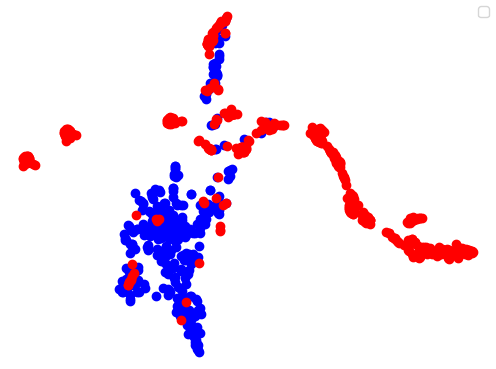} &
    \includegraphics[width=.12\textwidth]{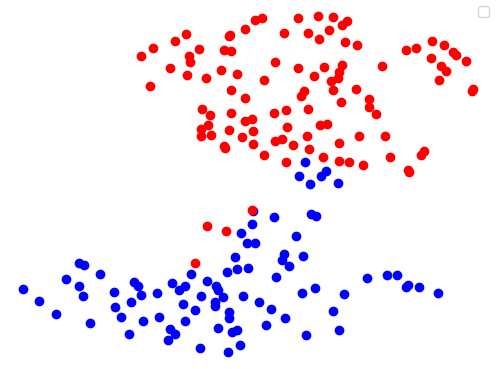} \\
     &
    \includegraphics[width=.12\textwidth]{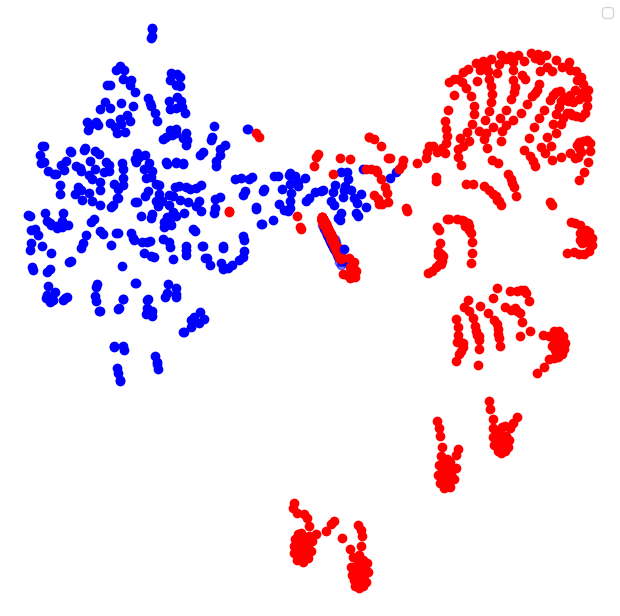} &
    \includegraphics[width=.12\textwidth]{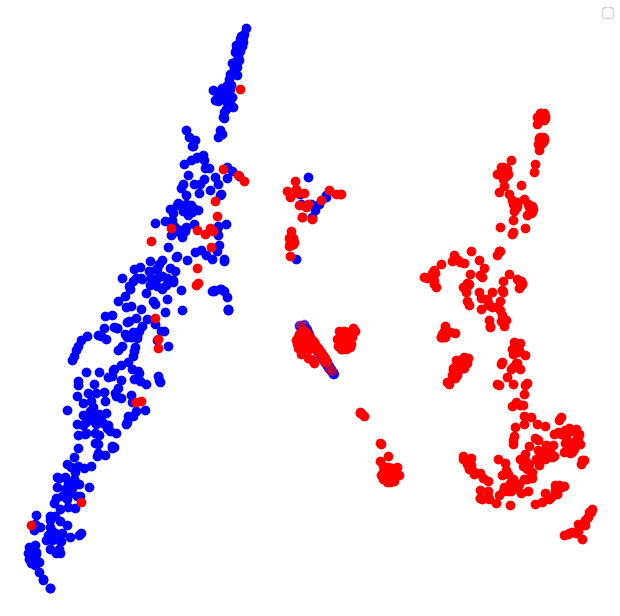} &
    \includegraphics[width=.12\textwidth]{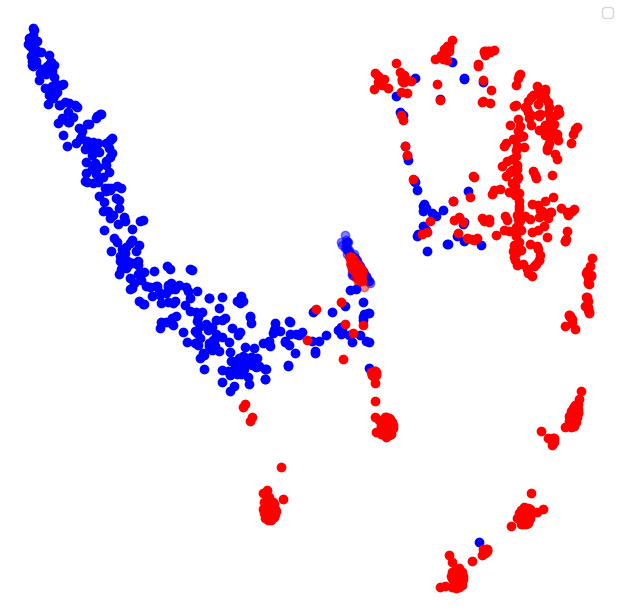} &
    \includegraphics[width=.12\textwidth]{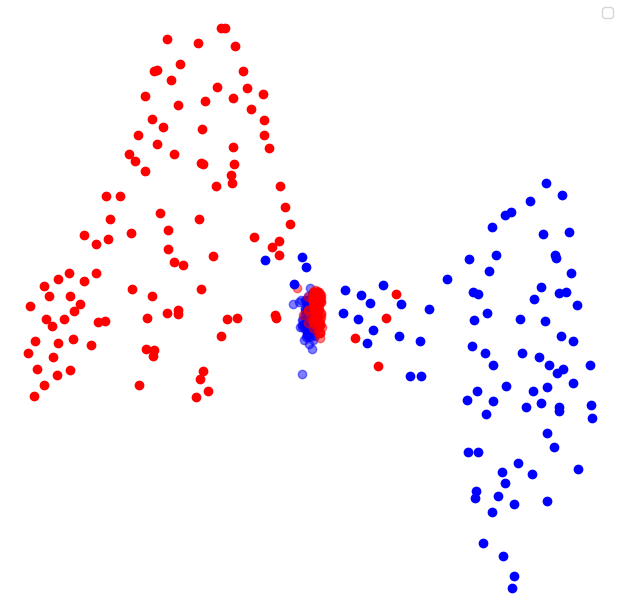} \\

     &
    \includegraphics[width=.12\textwidth]{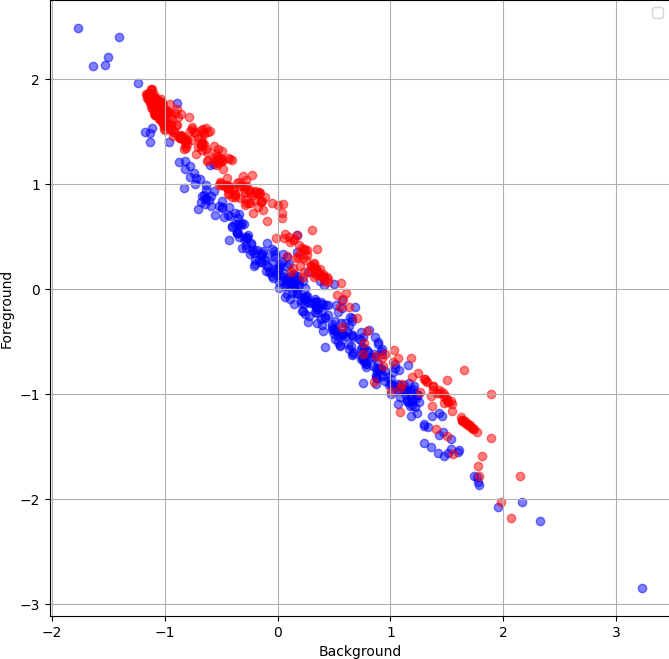} &
    \includegraphics[width=.12\textwidth]{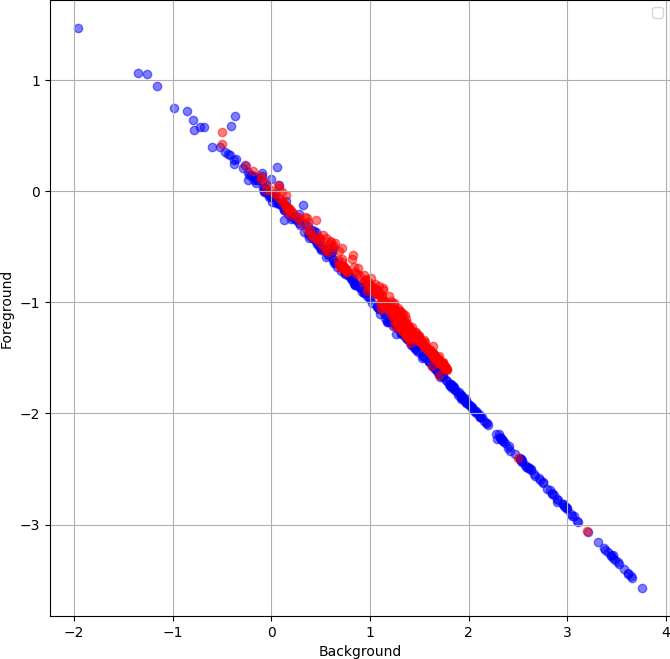} &
    \includegraphics[width=.12\textwidth]{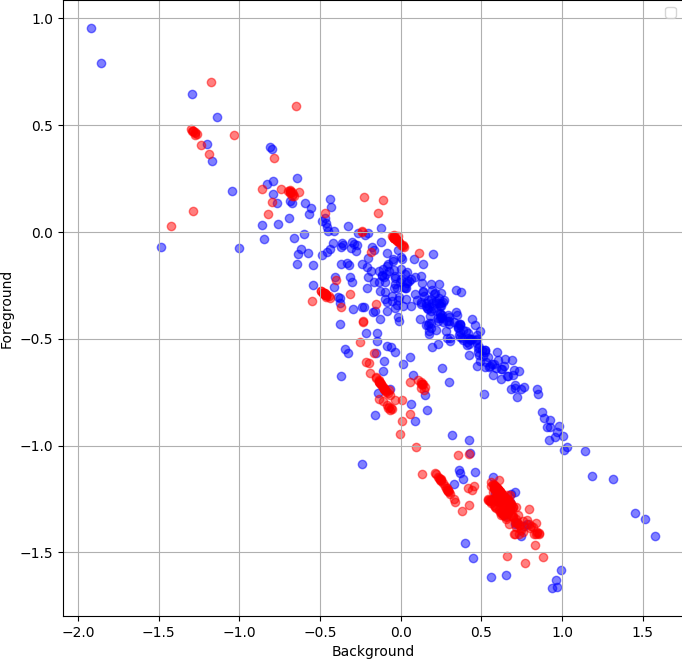} &
    \includegraphics[width=.12\textwidth]{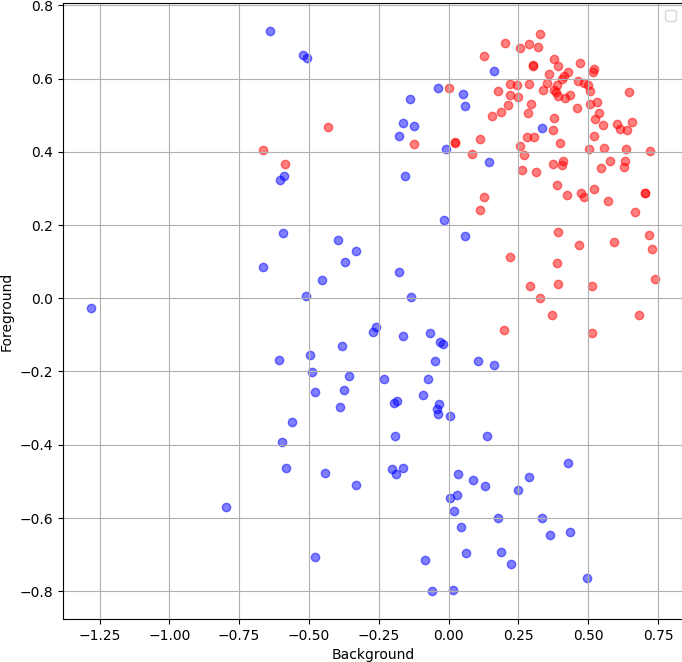} \\

     \includegraphics[width=.12\textwidth]{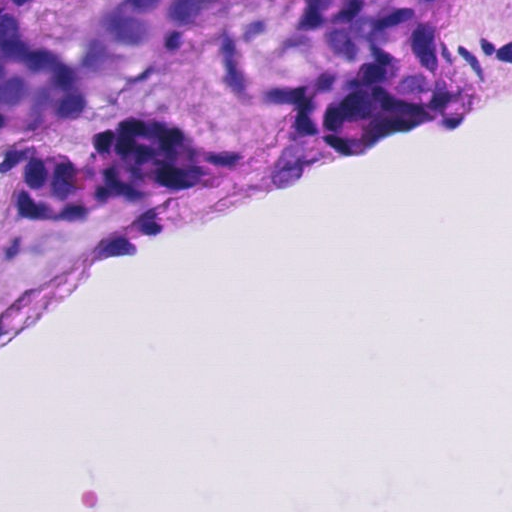} &
    \includegraphics[width=.12\textwidth]{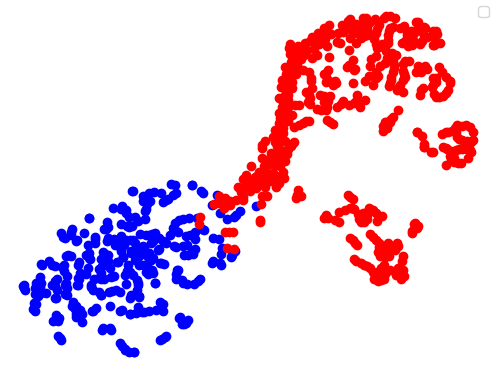} &
    \includegraphics[width=.12\textwidth]{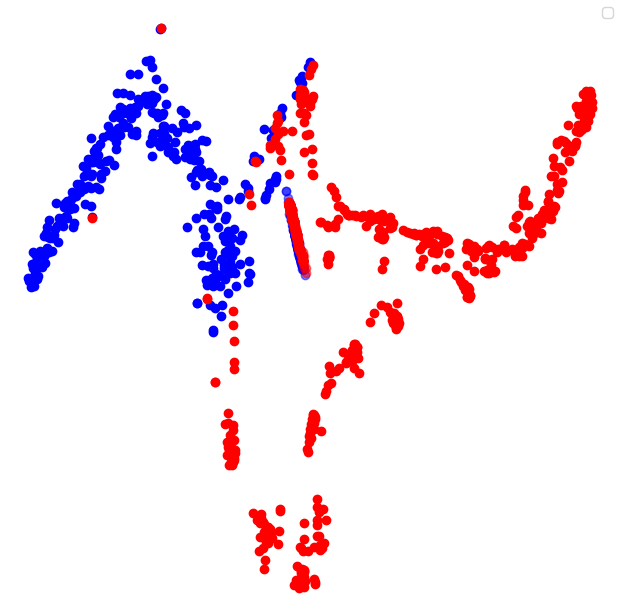} &
    \includegraphics[width=.12\textwidth]{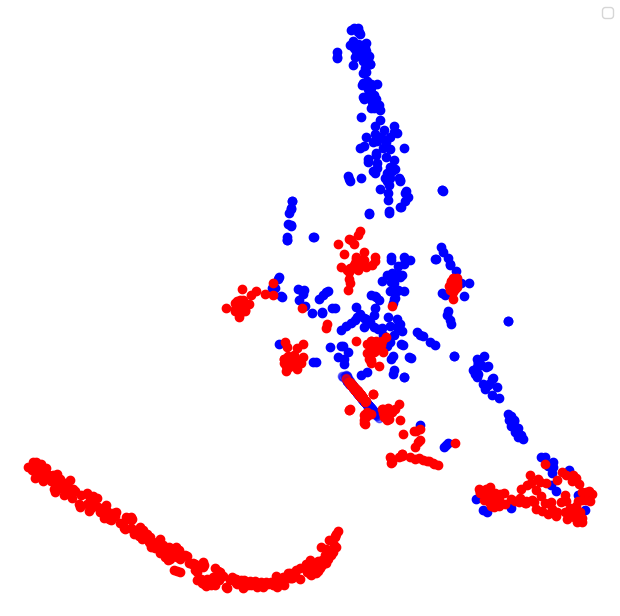} &
    \includegraphics[width=.12\textwidth]{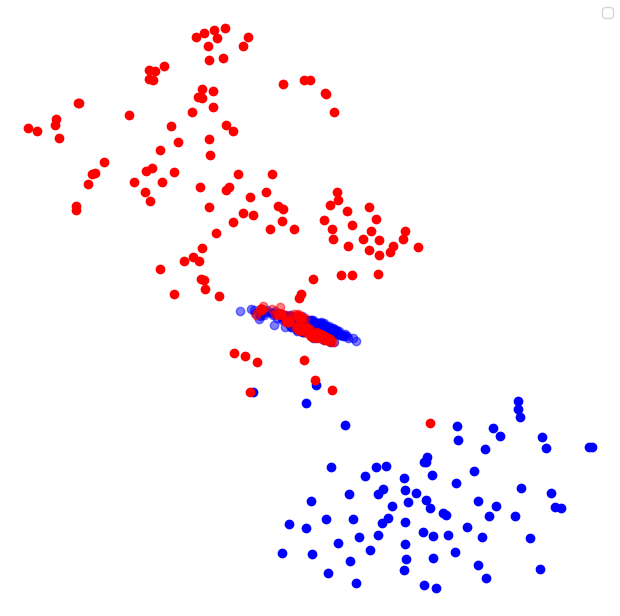} \\
     &
    \includegraphics[width=.12\textwidth]{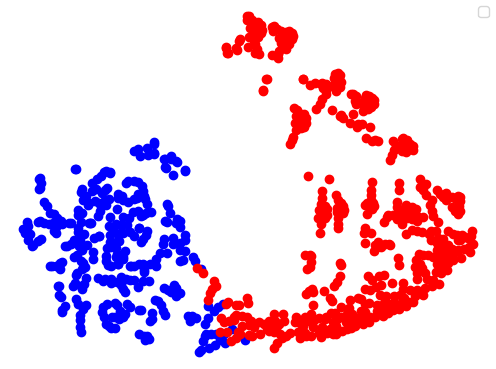} &
    \includegraphics[width=.12\textwidth]{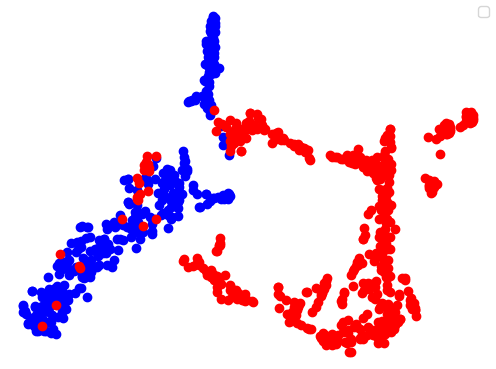} &
    \includegraphics[width=.12\textwidth]{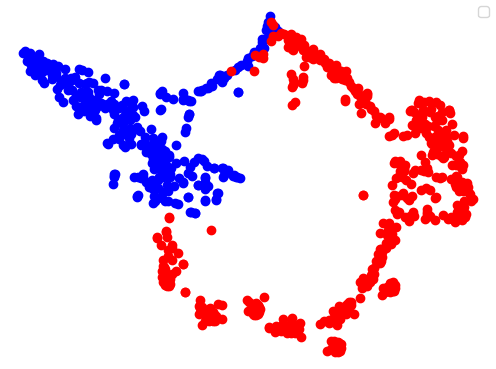} &
    \includegraphics[width=.12\textwidth]{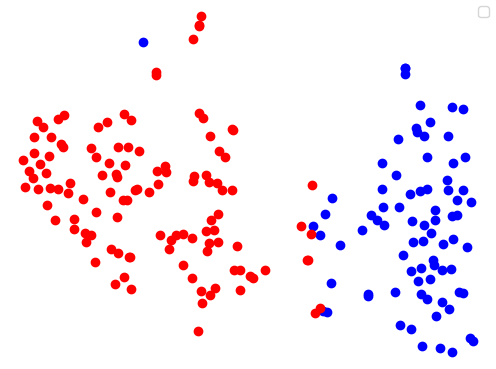} \\
     &
    \includegraphics[width=.12\textwidth]{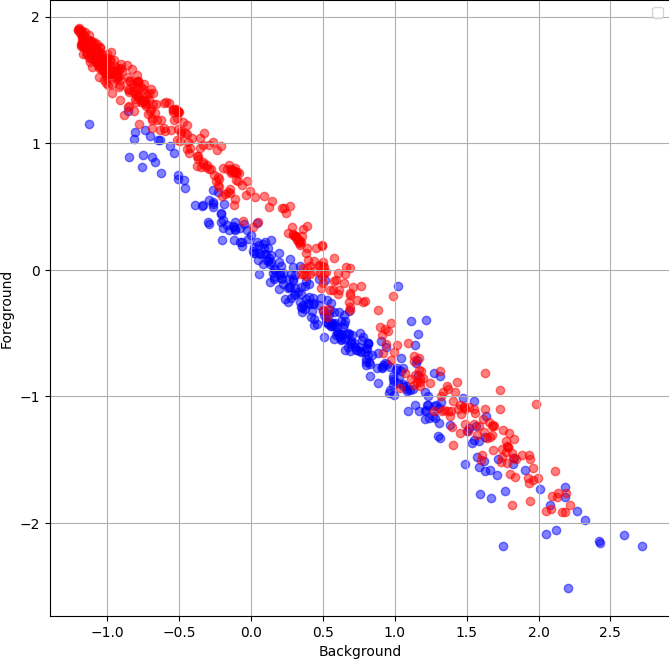} &
    \includegraphics[width=.12\textwidth]{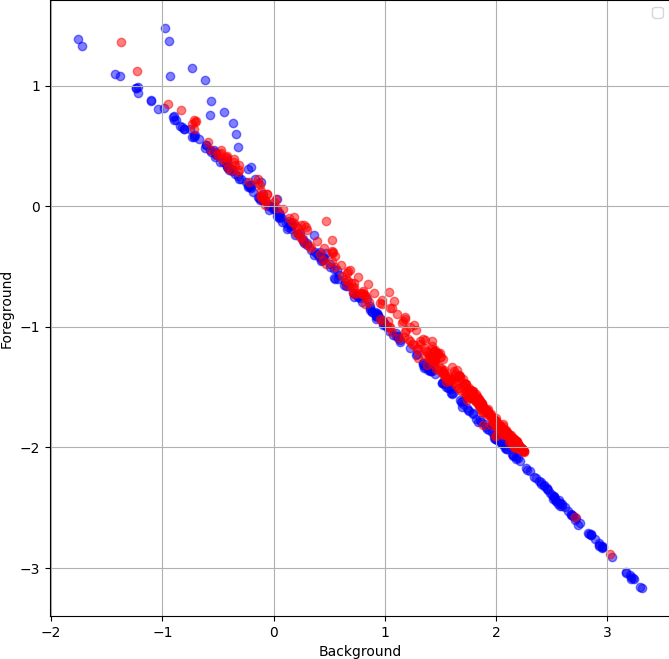} &
    \includegraphics[width=.12\textwidth]{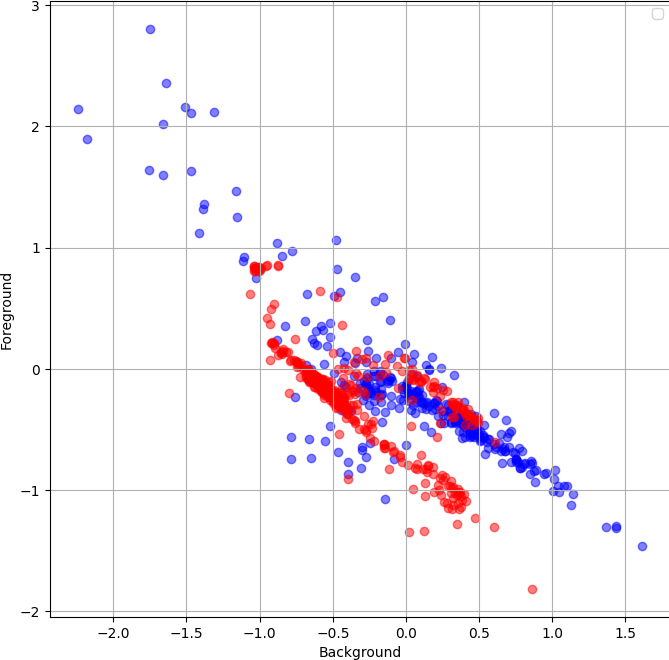} &
    \includegraphics[width=.12\textwidth]{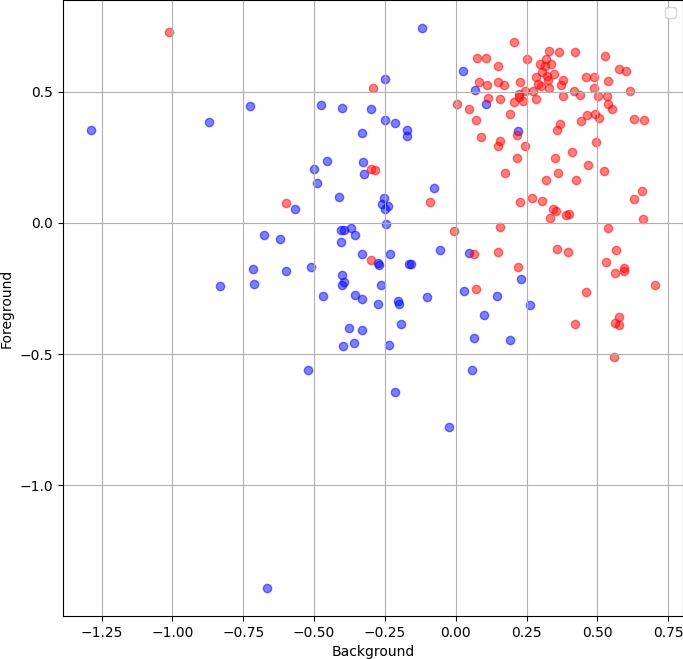} \\

    \includegraphics[width=.12\textwidth]{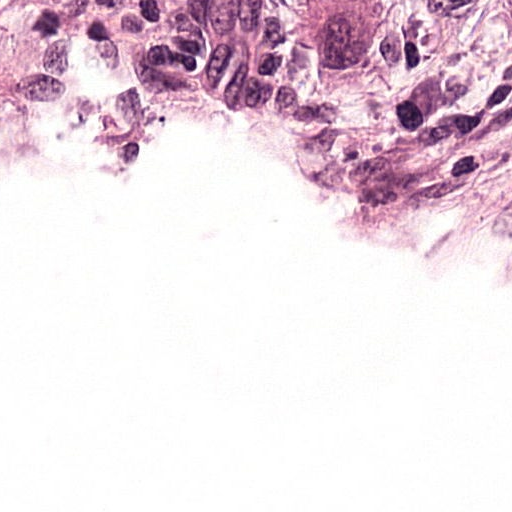} &
    \includegraphics[width=.12\textwidth]{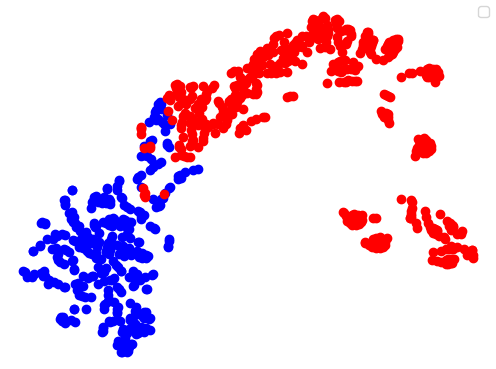} &
    \includegraphics[width=.12\textwidth]{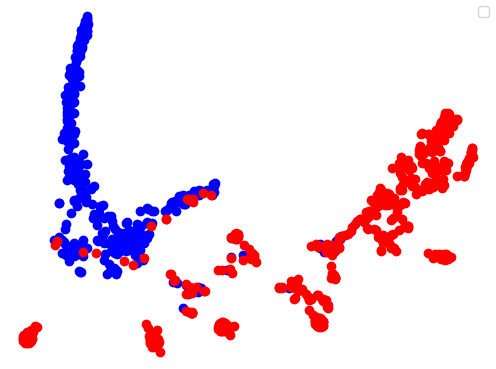} &
    \includegraphics[width=.12\textwidth]{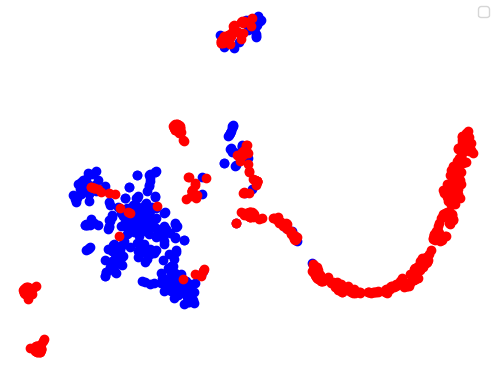} &
    \includegraphics[width=.12\textwidth]{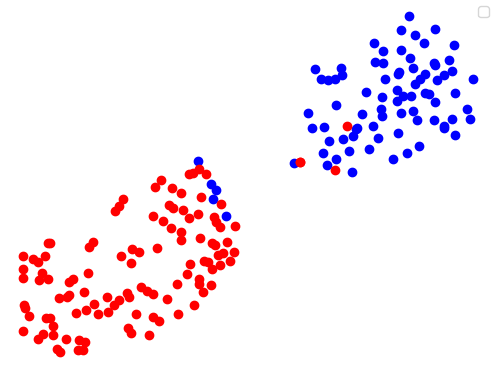} \\
    
    &
    \includegraphics[width=.12\textwidth]{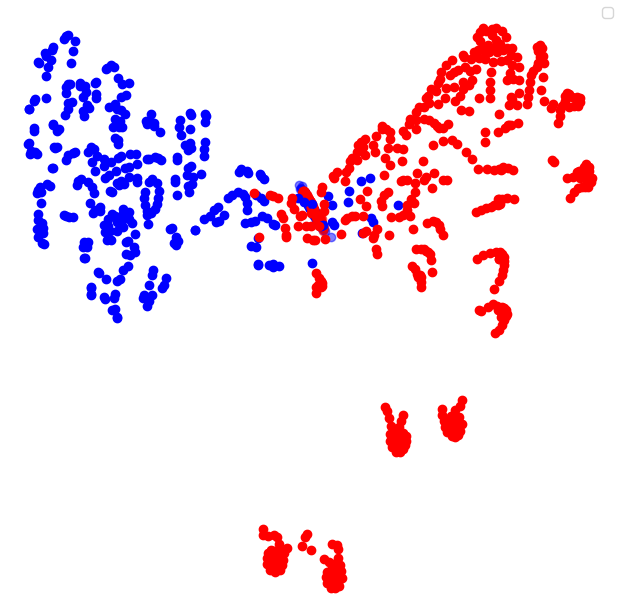} &
    \includegraphics[width=.12\textwidth]{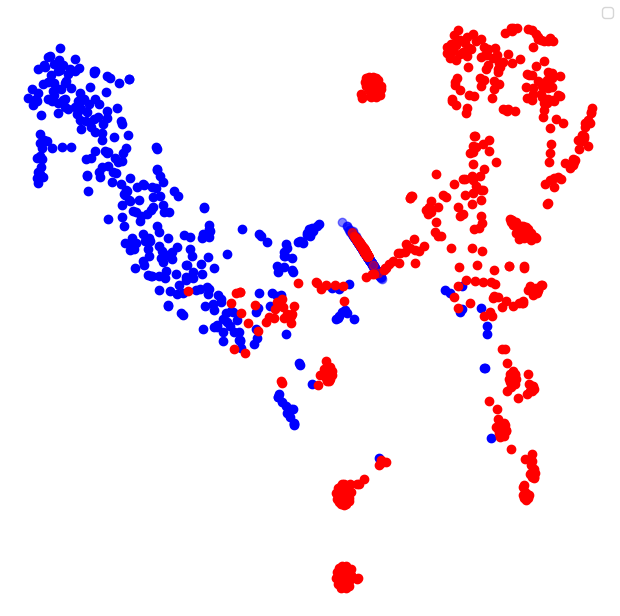} &
    \includegraphics[width=.12\textwidth]{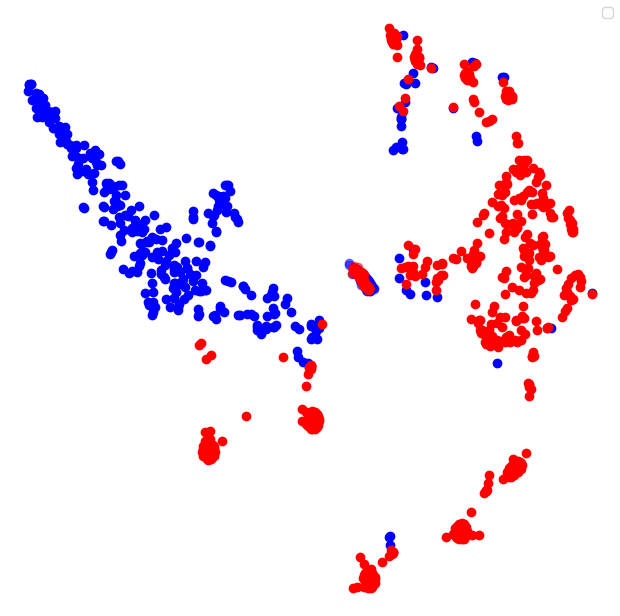} &
    \includegraphics[width=.12\textwidth]{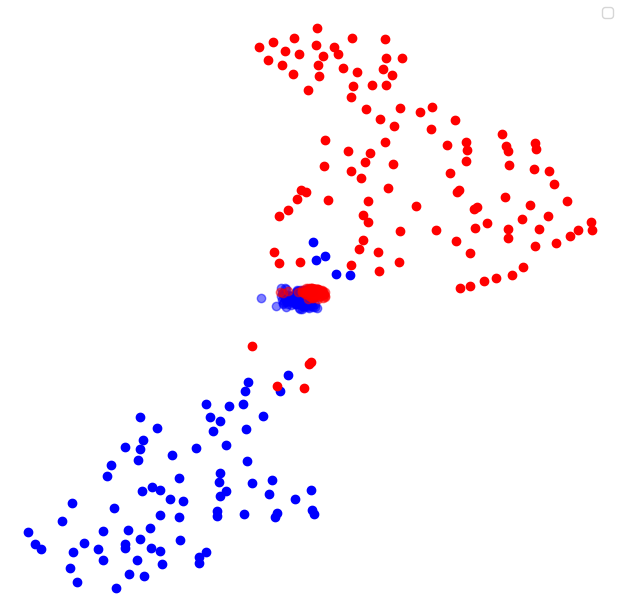} \\

     &
    \includegraphics[width=.12\textwidth]{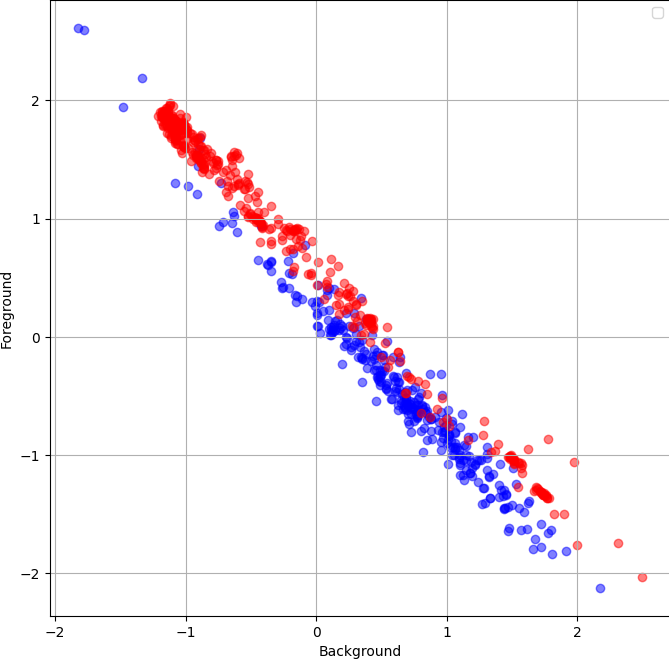} &
    \includegraphics[width=.12\textwidth]{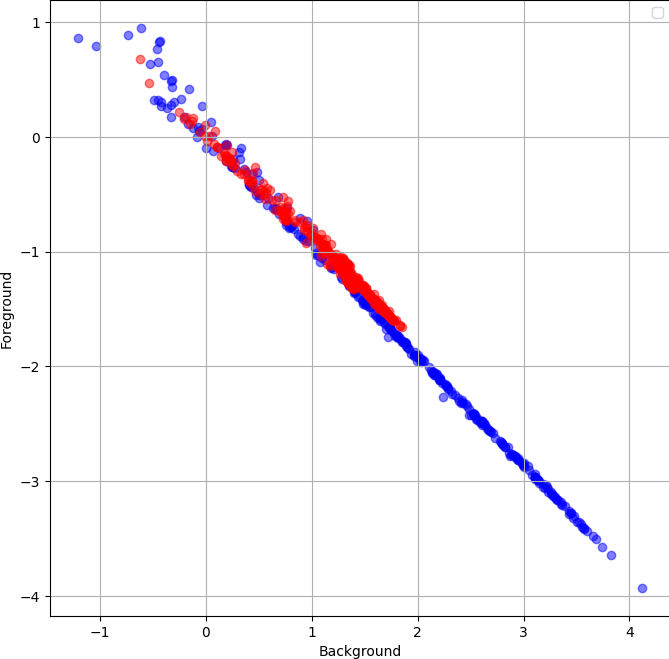} &
    \includegraphics[width=.12\textwidth]{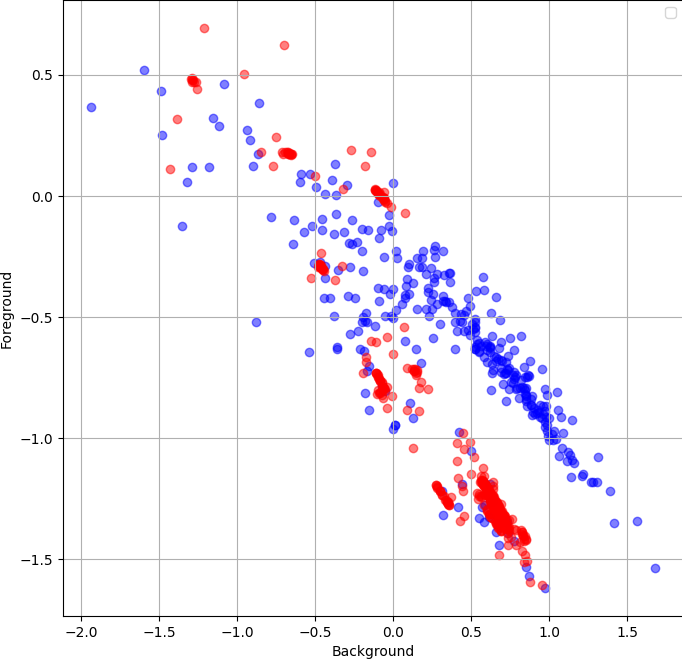} &
    \includegraphics[width=.12\textwidth]{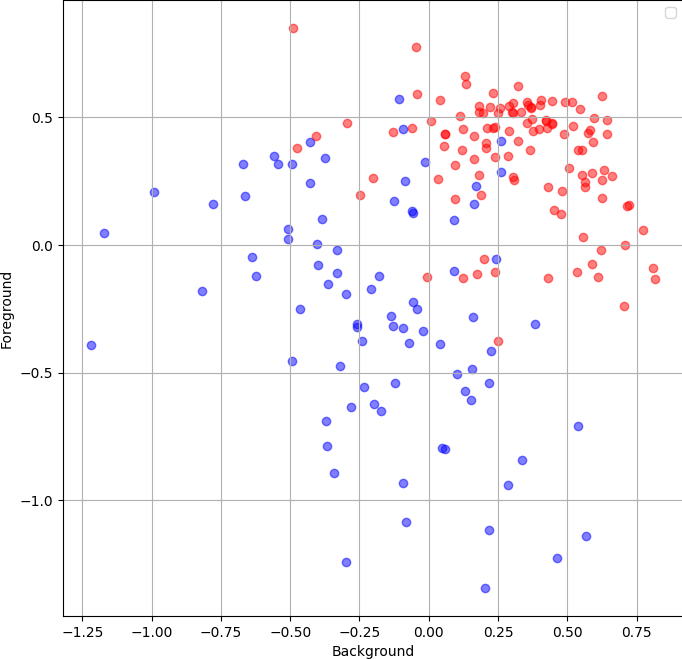} \\
       
  \end{tabular}
  \caption{
  OOD setup (\mbox{\glas $\rightarrow$ \camsixteen}): For the three \textbf{cancerous} images from \camsixteen, we display the t-SNE projection of foreground and background pixel-features of WSOL baseline methods without (1st row) and with (2nd row) \pixelcam. The 3rd row presents the logits of the pixel classifier of \pixelcam.
  }
  \label{fig:feature-tsne-target-cancer-cam16}
\end{figure*}
\clearpage
\balance
\bibliographystyle{abbrv}
\bibliography{main}

\end{document}